\documentclass[journal,12pt,draftclsnofoot,onecolumn]{IEEEtran}
\ifCLASSINFOpdf
\else
\fi

\usepackage{algorithm}
\usepackage{algorithmic}
\usepackage{multirow}
\usepackage{times}
\usepackage{epsfig}
\usepackage{graphicx}
\usepackage{amsmath}
\usepackage{amssymb}
\usepackage[font=small,labelfont=bf]{caption,subfig}
\usepackage{color}
\usepackage{nomencl}
\usepackage{relsize}
\usepackage{array}
\usepackage{url}
\usepackage{cite}
\usepackage{booktabs}
\usepackage{empheq}
\usepackage{bigints}

\newtheorem{theorem}{Theorem}[section]

\newtheorem{lemma}[theorem]{Lemma}
\newtheorem{proposition}[theorem]{Proposition}
\newtheorem{corollary}[theorem]{Corollary}

\newenvironment{proof}[1][Proof]{\begin{trivlist}
\item[\hskip \labelsep {\bfseries #1}]}{\end{trivlist}}
\newenvironment{definition}[1][Definition]{\begin{trivlist}
\item[\hskip \labelsep {\bfseries #1}]}{\end{trivlist}}

\newcommand{\qed}{\nobreak \ifvmode \relax \else
      \ifdim\lastskip<1.5em \hskip-\lastskip
      \hskip1.5em plus0em minus0.5em \fi \nobreak
      \vrule height0.75em width0.5em depth0.25em\fi}

\DeclareMathOperator*{\argmin}{argmin}
\DeclareMathOperator*{\argmax}{argmax}

\begin{document}
%
\title{Piecewise Linear Patch Reconstruction for Segmentation and Description of Non-smooth Image Structures}
%
%
%
\author{Junyan~ Wang*~\IEEEmembership{Student Member,~IEEE}~and~ Kap~Luk~Chan~\IEEEmembership{Member,~IEEE}
\thanks{Junyan Wang and Kap Luk Chan are with the
School of Electrical and Electronic Engineering, Nanyang Technological University, Singapore 639798 e-mail: \{wa0009an,eklcan\}@ ntu.edu.sg}}

%
%

\markboth{}%
{}
%



\maketitle

\begin{abstract}
In this paper, we propose a unified energy minimization model for the segmentation of non-smooth image structures. The energy of piecewise linear patch reconstruction is considered as an objective measure of the quality of the segmentation of non-smooth structures. The segmentation is achieved by minimizing the single energy without any separate process of feature extraction. We also prove that the error of segmentation is bounded by the proposed energy functional, meaning that minimizing the proposed energy leads to reducing the error of segmentation. As a by-product, our method produces a dictionary of optimized orthonormal descriptors for each segmented region. The unique feature of our method is that it achieves the simultaneous segmentation and description for non-smooth image structures under the same optimization framework. The experiments validate our theoretical claims and show the clear superior performance of our methods over other related methods for segmentation of various image textures. We show that our model can be coupled with the piecewise smooth model to handle both smooth and non-smooth structures, and we demonstrate that the proposed model is capable of coping with multiple different regions through the one-against-all strategy.
\end{abstract}



%
\IEEEpeerreviewmaketitle

\section{Introduction}




Computer vision problems are often addressed by using mathematical models. The quality of the solutions to the problems are measured objectively in the mathematical models. With the valid mathematical models, we can elucidate the phenomenon of the natural computations, e.g. by human vision, that try to accomplish the tasks, and we can reproduce the result of the natural computations by computerized simulations.
%

An objective measure of piecewise smooth image segmentation is the Mumford-Shah functional energy \cite{MumfordShah89Functional}. By minimizing the energy, we expect to achieve high quality of segmentation for piecewise smooth images. The problem of minimization of the Mumford-Shah functional energy is the mathematical abstraction, i.e. the model, of piecewise smooth image segmentation. The model is now known as the Mumford-Shah model. This methodology is different from that of those in \cite{Martin_Malik01FTM} \cite{Unnikrishnan2007ObjEva} which target at the objective evaluation of the segmentation based on the ground-truth results from normal subjects.

\begin{figure}
\centering
  \subfloat[]{\label{SUBFIG:exp1}\includegraphics[height=0.23\columnwidth,width=0.2\columnwidth]{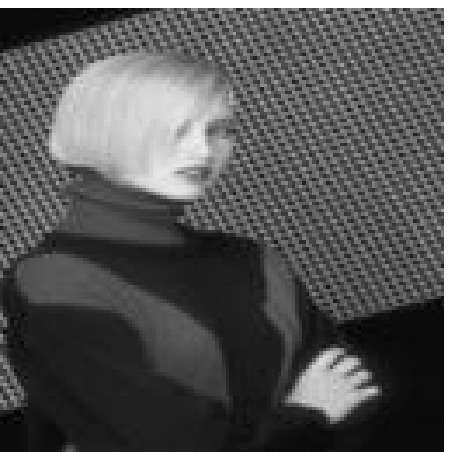}\includegraphics[height=0.23\columnwidth,width=0.23\columnwidth]{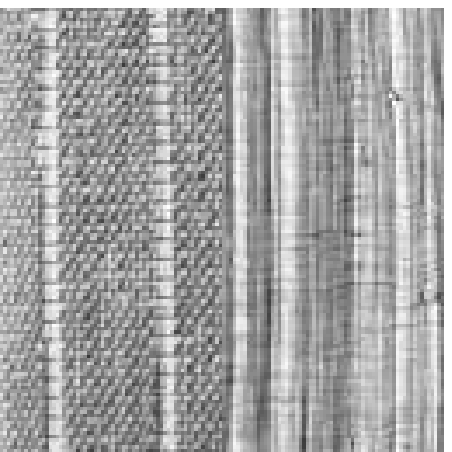}}~  \subfloat[]{\label{SUBFIG:exp2}\includegraphics[height=0.23\columnwidth,width=0.55\columnwidth]{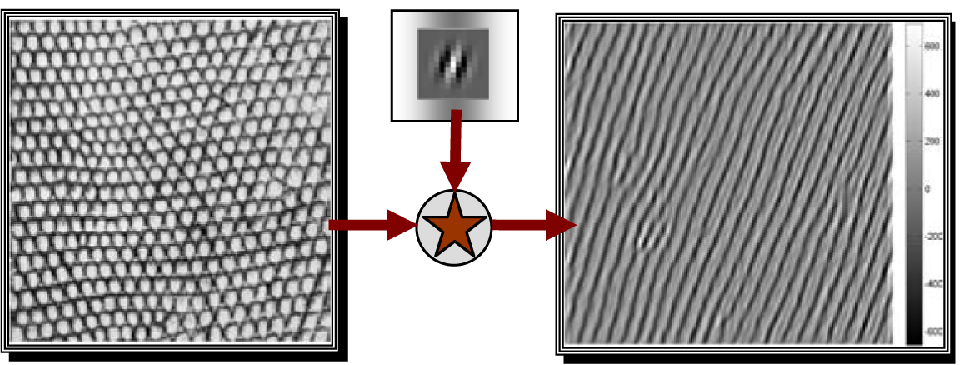}}
\caption{Illustration of the problem. The left two images are composed of different image structures that the conventional Mumford-Shah model would fail to segment; The diagram on the right shows that the Gabor feature image of a texture is nonsmooth.}\label{FIG:example}
\end{figure}

In Mumford-Shah model, each image region is modeled as a smooth or constant function with parameters. By minimizing the energy of the Mumford-Shah model, we can restore the smooth image in each region, and we can also obtain the partition boundaries of the segmentation located at the discontinuities in the restored image. However, the image values to be restored may not be piecewise smooth or flat as a whole. The images may contain regions of non-smooth structures. Such as the images in Fig. \ref{FIG:example}(a). Imposing the smoothness on these images leads to the destructive averaging of the image content. This poses problem to the segmentation with the conventional Mumford-Shah model. For example, it is possible that non-smooth visual patterns different in structure may have similar average image values (See Fig. \ref{FIG:example}(a)). Consequently, the Mumford-Shah model cannot separate such patterns in the image space.

To cope with non-smooth image data, the segmentation has been considered as a framework of two independent processes, i.e. the feature extraction and the segmentation of the feature image. There exist combinations of the Mumford-Shah model or other active contour model with predefined features \cite{Rousson03DiffuAC} \cite{Houhou08FastDiffuAC} \cite{NiChan09CVHist}. The predefined features does not necessarily match the underlying non-smooth image structures. For example, in Fig. \ref{FIG:example}(b) the feature values obtained by image filtering with the Gabor filters that well match the texture pattern, or the local averages of the feature values, can have significant spatial variations. Unsupervised feature selection has also been used \cite{Sandberg2002level} \cite{Sagiv06Texture} \cite{Kokkinos09GenTextureCE}.
The major problem concerns us is that these frameworks, formed by separate feature selection and segmentation, do not provide a unified optimization model for segmentation. These concerns motivate us to explore a unified and valid mathematical model for segmentation of non-smooth structures.

In this paper, we propose a novel unified energy minimization model of the segmentation of general non-smooth image structures, motivated by the original Mumford-Shah model. We formulate the segmentation of general non-smooth image structures as a single energy minimization problem of piecewise linear patch reconstruction. The formulation is not heuristic since we prove that the error of segmentation is bounded by the energy of piecewise linear patch reconstruction for a fixed number of bases. Thus, minimizing the energy of piecewise linear patch reconstruction can reduce the error of segmentation. Regarding the energy minimization, we prove that the eigen-patches constitute the global optimal solution to the minimization of the error of linear patch reconstruction. We also explore a more efficient alternative to the eigen-decomposition of matrix for computing eigenvectors. We prove that the linear patch reconstruction based on gradient descent converges to the eigen-patches under some mild assumption on the initializations. The assumption can often be met, and the convergence is linear. We therefore propose the gradient descent algorithm for solving the linear patch reconstruction problem even though the problem is not convex. The segmentation algorithm is based on alternating piecewise linear patch reconstruction and curve evolution. The piecewise linear patch reconstruction can naturally be coupled with the conventional Mumford-Shah model for coping with both smooth and non-smooth structures. A unique feature of the method is that it produces both the segmentation and a dictionary of optimized (orthonormal) descriptors for each segmented region upon completion in the same optimization framework.

The rest of the paper is organized as follows. We reviewed the related works in section \ref{SEC:BAK}. We study the PC/PS model and propose the piecewise linear patch reconstruction model in sections \ref{SEC:SEG_RULE}-\ref{SEC:FS_Linear_patch_SEG}. We show our error bound of the segmentation in terms of reconstruction error in section \ref{SEC:SEG_ERR} and we present our proved theoretical claim of the global optimality of the gradient descent for the nonconvex linear patch reconstruction problem in section \ref{SEC:GD_opt_rec}. The experimental results are presented in section \ref{SEC:Exp}. We conclude the paper in section \ref{SEC:Con}.

\section{Background and related works}\label{SEC:BAK}
\subsection{The Mumford-Shah model}
%

%
The segmentation of piecewise smooth images has been formulated as an energy minimization problem in the Mumford-Shah model. For an image defined over the image domain $\{[x,y]^T\in\Omega\}$, the two-phase Mumford-Shah model can be formulated as follows.
\begin{equation}\label{EQ:PSMS}
\begin{split}
\min_{\phi} E&(\phi,g1,g2),\\
E(\phi,g1,g2)&=\int_{\Omega} \mathcal{E}_1(x,y,g_1) H(\phi) dxdy\\
&~~+\int_{\Omega} \mathcal{E}_2(x,y,g_2)(1-H(\phi)) dxdy\\
&~~+\nu \int_{\Omega} \delta(\phi(x,y))\|\nabla\phi\| dxdy\\
\end{split}
\end{equation}
where $\delta(\cdot)$ is a Dirac delta function, $H(\cdot)$ is a Heaviside (step) function, $\nu$ is a penalty coefficient. $g_1$ and $g_2$ are the reconstruction estimates, $\phi$ is the signed distance function
that partitions the image into two regions, i.e. $\phi>0$ for one region, $\phi\leq0$ for the other. $\mathcal{E}_1$ and $\mathcal{E}_2$ are the reconstruction errors of the two region models. $H$ assigns either of the two models to all the pixels, and the last term tries to minimize the complexity of the labeling. For piecewise smooth (PS) model, we have $\mathcal{E}_1(x,y) = (I-g_1)^2+\lambda\|\nabla g_1\|^2$, $\mathcal{E}_2(x,y) = (I-g_2)^2+\lambda\|\nabla g_2\|^2$, where $I(x,y)$ is the image value at $[x,y]^T$, $\lambda$ is a constant penalty coefficient. Note that, $I$, $g_1$ and $g_2$ are all functions defined on $[x,y]^T$. We will omit $(x,y)$ behind these functions henceforth if there is no risk of confusion.
The fundamental optimization technique for the reconstruction is the Green's functions solution to the Euler-lagrange equation obtained by Calculus of Variations, which was presented in \cite{MumfrodShah85CVPR}. The numerical schemes for solving the Euler-lagrange equation can also be found in \cite{Tsai01PSMumfordShah} \cite{Brox2009InterpMumfordShah}. The smoothness regularization guarantees the global optimality of the solution obtained by any of the optimization methods for a given partition.

If we assume the reconstruction functions $g_1$, $g_2$ to be constants, i.e. for all $(x,y)$ in $\Omega$, $g_1(x,y)=c_1$ and $g_2(x,y)=c_2$, then the gradient terms in the PS model vanish, and the functional becomes the piecewise constant (PC) Mumford-Shah model, which is the prototype of the region competition \cite{Zhusongchun96RegComp} and the Chan-Vese model \cite{ChanVese01ActiveCon}.

For minimizing the Mumford-Shah functional, the algorithm is often the alternating (or simultaneous) implementation of segmentation and reconstruction. The reconstruction is achieved by image smoothing within each region of segmentation. The level set method is often used in \cite{ChanVese01ActiveCon} for the segmentation. In recent years, the linear relaxation \cite{Grady09GraphMumfordShah} and the convex relaxation \cite{Brown10ConvexChanVese} of the Mumford-Shah model have been proposed, which lead to alternatives to the level set method.


\subsection{Region-based active contours for unsupervised texture segmentation}
The original the Mumford-Shah model was formulated for segmentation of piecewise smooth images. Later the model was extended for image texture segmentation. In \cite{Lee92MumfordShahTexture} \cite{Sandberg2002level} \cite{Sagiv06Texture}, the Gabor filtering was incorporated in the Mumford-Shah model.
In \cite{Lee92MumfordShahTexture}, Lee et al. embedded the manually selected 24 Gabor filters into the PS Mumford-Shah model. In \cite{Sandberg2002level}, Sandberg et al. adopted the Chan-Vese model as well as the maximum difference of feature means as the criterion for Gabor filter selection. Sagiv \emph{et al.} \cite{Sagiv06Texture} adopted the framework in \cite{Sandberg2002level} with manually selected filters. These frameworks require the textures to be piecewise smooth or flat in the feature space, but the filter selection for unsupervised segmentation cannot ensure this. Although the ad-hoc postprocessing of high dimensional anisotropic diffusion may be applied to ensure the piecewise smoothness, the reasons behind this remain obscure. Kokkinos \emph{et al.} \cite{Kokkinos09GenTextureCE} proposed to apply the Region Competition model to a type of modulation features. This framework assumes that the textures are globally oscillating, as on zebras and tigers, and it still requires filter selection by dominant component analysis (DCA) for parametric texture modeling. However, it is actually unknown what kind of feature or the principle of feature selection can help the segmentation by Mumford-Shah model without supervision. Therefore, the existing combinations of the feature extraction and the Mumford-Shah model is heuristic and they do not provide a unified mathematical model for the segmentation.

There exist other frameworks of texture segmentation based on other similar active contour models with fixed texture feature, e.g. \cite{NiChan09CVHist} \cite{Rousson03DiffuAC} and \cite{Houhou08FastDiffuAC}. For example, in \cite{Rousson03DiffuAC} and \cite{Houhou08FastDiffuAC} the structure tensor has been chosen as the texture feature, and in \cite{NiChan09CVHist} the histogram of image values on overlapping patches is chosen as the feature. These frameworks are more likely to form a unified theory of the segmentation. However, the choice of the structure tensor feature or local histogram has not been validated for general textures.

\section{The principle of segmentation behind the Mumford-Shah model}\label{SEC:SEG_RULE}
In this section, we study the segmentation by Mumford-Shah model to understand the rationale of this model for segmentation. We restricted ourselves to the two-phase model. The generalization of the two-phase framework to multi-phase may follow \cite{Vese02MultiPhase}, which is out of the scope here.

Let us consider the two-phase Mumford-Shah model in a simplified form as follows.
\begin{equation}\label{EQ:REC_SEG0}
\begin{split}
&H^* = \argmin\limits_{H} \int\limits_\Omega \mathcal{E}_1H dxdy+\int\limits_\Omega \mathcal{E}_2 (1-H)dxdy
\end{split}
\end{equation}
where $\mathcal{E}_1$ and $\mathcal{E}_2$ are the reconstruction errors corresponding to the subregions $\Omega_1$ and $\Omega_2$, such that $\Omega=\Omega_1\cup\Omega_2$. Since we focus on the region model in this work, we omit discussing about the prior term of arclength for imposing smoothness. Note that, the smoothness term is important for dealing with noisy data. For more detailed discussions on the smoothness term, we refer the readers to \cite{MumfordShah89Functional} \cite{Zhusongchun96RegComp}.

In the PC model, $\mathcal{E}_1(x,y) = \|I(x,y)-c_1\|^2$, $\mathcal{E}_2(x,y) = \|I(x,y)-c_2\|^2$, where $I(x,y)$ is the image value, $c_1,c_2$ are the regional means of the image value. In PS model, the $\mathcal{E}_1(x,y) = (I-g_1)^2+\lambda\|\nabla g_1\|^2$, $\mathcal{E}_2(x,y) = (I-g_2)^2+\lambda\|\nabla g_2\|^2$, where $g_1,g_2$ are the smooth restoration of the two image regions (due to the smoothness constraint). The smoothness regularization terms, $\lambda\|\nabla g_1\|^2$ and $\lambda\|\nabla g_2\|^2$ are often omitted if $g_1$ and $g_2$ are solved by the normalized Gaussian convolution, such as in \cite{Jerome07SolveMumfordShah}, which was interpreted as nonparametric regression \cite{Brox2009InterpMumfordShah}.
Regarding the minimization of this simplified functional (\ref{EQ:REC_SEG0}), we have the following interesting fact upon fixing the error functions $\mathcal{E}_1(x,y)$ and $\mathcal{E}_2(x,y)$.
\begin{proposition}\label{PROP:Opt_H}
\begin{equation}\label{EQ:Simp_MS}
\begin{split}
&\min_{H} \int\limits_\Omega \mathcal{E}_1H+\mathcal{E}_2(1-H)dxdy\\
&\hspace{0.8in}=  \int\limits_\Omega \min_{H} \left\{(\mathcal{E}_1-\mathcal{E}_2) H\right\} dxdy+C
\end{split}
\end{equation}
where $H(x,y)=\{0,1\}$ and $C$ is a constant independent of $H$.
\end{proposition}
This fact tells that optimizing the global assignment of $H$ is equivalent to optimizing the assignment locally. The assignment rule for determining the optimal $H$ at each pixel location is therefore the following.
\begin{equation}\label{EQ:SEG_RULE}
H(x,y) = \left\{\begin{array}{lr}
                  1,& \mathcal{E}_1(x,y)\leq\mathcal{E}_2(x,y) \\
                  0,& \mathcal{E}_1(x,y)\geq\mathcal{E}_2(x,y)
                \end{array}
\right.
\end{equation}
The proof is included in the appendix in a separate report. As in the interpretation in \cite{Zhusongchun96RegComp}, Mumford-Shah model tries to achieve clustering based on the measure of the membership defined by $\mathcal{E}_l,~l=1,2$. Hence, the proper choice of the measure is essential to the segmentation by Mumford-Shah functional.

\section{Piecewise linear patch reconstruction}\label{SEC:FS_Linear_patch_SEG}
In what follows, we establish the mathematical model and the associated solution for the segmentation of non-smooth image structures.

\subsection{The model of piecewise linear patch reconstruction}
In the original Mumford-Shah model, each region is modeled as a piecewise smooth surface $g_l$ that minimizes the functional energy as follows.
\begin{equation}
g_l^* = \argmin_{g_l}\int_{\Omega_l} (I-g_l)^2 + \lambda\|\nabla g_l\|^2 dxdy
\end{equation}
where $l=1,2$, $\lambda$ is a penalty coefficient. A major reason for this formulation is that the energy can be minimized by solving standard partial differential equations, i.e. the diffusion equation. Besides, the energy is convex. Hence, if the local optimal solution exists it is also global optimal. The problem is that this formulation imposes the smoothness.

To cope with non-smooth image structure, we consider the image patches as linear combination of patch bases, such as in many image appearance models, as follows.
\begin{equation}\label{EQ:patch_basis}
\begin{split}
\mathbf{p}(x,y) &= \sum_{k=1}^{K}\alpha_k(x,y)\mathbf{v}_k=\sum_{k=1}^{K}\big\langle\mathbf{p},\mathbf{v}_k\big\rangle_{\square}\mathbf{v}_k
\end{split}
\end{equation}
with the varying weights $\alpha_k=\big\langle\mathbf{p},\mathbf{v}_k\big\rangle_{\square}$ for different patches over the image, where $\{\mathbf{v}_k,i=1,2,...,K\}$ is a set of $K$ orthonormal bases that fully reconstruct the patch. $\mathbf{p}(x,y)=[p_{uv}(x,y)|[u,v]^T\in\square_{xy}]$ where ${p}_{uv}(x,y) = I(x-u,y-v)$ and $\square_{xy}$ is a square region centered at $[x,y]^T$.

It is obvious that the exact reconstruction can happen regardless of smoothness of the image patch. To find the bases, we may formulate the reconstruction as energy minimization as follows.
\begin{equation}
\{\mathbf{v}^l_k\}^* = \argmin_{\{\mathbf{v}^l_k\}}\bigintsss_{\Omega_l} \left\|\mathbf{p}-\sum_{k=1}^{K}\big\langle\mathbf{p},\mathbf{v}^l_k\big\rangle_{\square}\mathbf{v}^l_k\right\|_{\square}^2dxdy
\end{equation}
Replacing the image reconstruction error in simplified two-phase Mumford-Shah functional (\ref{EQ:Simp_MS}) with the patch reconstruction error formulated above, we obtain the formulation of piecewise linear patch reconstruction as follows.
\begin{equation}
\min_{\left\{H,~\{\mathbf{v}^1_k\},\{\mathbf{v}^2_k\}\right\}} \int_\Omega \mathcal{E}^{\square}_1 H+ \int_\Omega \mathcal{E}^{\square}_2 (1-H) dxdy
\end{equation}
where
\begin{equation}\label{EQ:Err_expand}
\begin{split}
\mathcal{E}^{\square}_l &= {1\over\|\square\|}\left\|\mathbf{p}-\sum_{k=1}^{K}\big\langle\mathbf{p},\mathbf{v}^l_k\big\rangle_{\square}\mathbf{v}^l_k\right\|_{\square}^2\\
&={1\over\|\square\|}\left[\int_\square I(x+u,y+v)^2 dudv\right.\\
&\hspace{10pt}\left.- \sum_{k=1}^{K} \left(\int_\square I(x+u,y+v)\mathbf{v}^l_k(u,v)dudv\right)^2\right]
\end{split}
\end{equation}
for $l=1, 2$, where $\|\square\|$ is the size of the patch. From the above, we can note that the error can be computed by convolutions.

From the assignment rule in (\ref{EQ:SEG_RULE}), we know that the patches will be assigned to a region if the set of bases of this region produces the patch reconstruction error smaller than that of the other regions.


\subsection{The error of segmentation by piecewise linear patch reconstruction}\label{SEC:SEG_ERR}
In the above, we have shown that the piecewise linear patch reconstruction is capable of determining the assignment of the patches according to the patch reconstruction error, but we did not answer whether the assignment is correct. In the following, we establish the theoretical foundation of the proposed formulation. Specifically, we show that the error of the assignment, i.e. the segmentation error, is bounded by the patch reconstruction error for a fixed number of, say $K$, bases. Thus, by minimizing the error of reconstruction, we may reduce the error of segmentation.

The assignment rule of (\ref{EQ:SEG_RULE}) enables us to assess the probabilistic analysis of the correctness of the segmentation. We ask whether the segmentation by (\ref{EQ:SEG_RULE}) is consistent with the truth. Specifically, we wish to know if $\mathbf{p}'=\mathbf{p}(x',y')\in\big\{\mathbf{p}\big|[x,y]^T\in\Omega_1\big\}=\mathbf{P}_1$, whether $\mathcal{E}_1(x',y')<\mathcal{E}_2(x',y')$; and if $\mathbf{p}'=\mathbf{p}(x',y')\in\big\{\mathbf{p}\big|[x,y]^T\in\Omega_2\big\}=\mathbf{P}_2$, whether $\mathcal{E}_2(x',y')<\mathcal{E}_1(x',y')$, where $\Omega=\Omega_1\cup\Omega_2$ is the true partition. $\big\{\mathbf{p}\big|[x,y]^T\in\Omega_l\big\}$ means the set of patches that have their centers in $\Omega_l$, where $l=1,2$. This requires us to analyze the following segmentation error rate.
\begin{equation}
\begin{split}
&\varepsilon_{seg} = {1\over|\Omega_1|}\sum\limits_{[x',y']^T\in\Omega_1} \mathbf{1}_{[\mathcal{E}_1(x',y')>\mathcal{E}_2(x',y')]}\\
&\hspace{20pt}+{1\over|\Omega_2|}\sum\limits_{[x',y']^T\in\Omega_2} \mathbf{1}_{[\mathcal{E}_2(x',y')>\mathcal{E}_1(x',y')]}
\end{split}
\end{equation}
where $|\Omega_1|$, $|\Omega_2|$ are the sizes of the sets $\Omega_1$ and $\Omega_2$.

The segmentation error rate can be represented by probability, i.e. $\varepsilon_{seg}~=~P_{_{\Omega_1}}\big[\mathcal{E}_1(x',y')>\mathcal{E}_2(x',y')\big]+P_{_{\Omega_2}}\big[\mathcal{E}_2(x',y')>\mathcal{E}_1(x',y')\big]$ for sufficiently large population of $\Omega_1$ and $\Omega_2$, where $P_{_{\Omega_1}}$ is the probability of certain events due to $\mathbf{P}_1$, $P_{_{\Omega_2}}$ is the probability of the events due to $\mathbf{P}_2$. We hope the error rate to be small. For our patch reconstruction model, we can deduce the following error bound for segmentation.
\begin{proposition}\label{Prop:Seg_Patch_err}
If the subregion $\Omega_1$, $\Omega_2$ are sufficiently large, such that $E_{_{\mathbf{P}_1}}[\mathcal{E}^{\square}_1]={1\over|\Omega_1|}\int_\Omega \mathcal{E}^{\square}_1 H dxdy ={1\over|\Omega_1|}\sum\limits_{\Omega_1} \mathcal{E}^{\square}_1$, $E_{_{\mathbf{P}_2}}[\mathcal{E}^{\square}_2]={1\over|\Omega_2|}\int_\Omega \mathcal{E}^{\square}_2 (1-H) dxdy ={1\over|\Omega_2|}\sum\limits_{\Omega_2} \mathcal{E}^{\square}_2$, then the following holds.
\begin{equation}\label{BD:SEG_ERR}
\varepsilon_{seg}\leq \sum_{l=1,2}{\|\square\|\int_\Omega \mathcal{E}^{\square}_l(x,y) H_l dxdy\over|\Omega_l|(\big\|\mathbf{p}\big\|_{_{\square}}-R_{3-l}^2K^{2-2{\textstyle q}_{3-l}})}\\
\end{equation}
where $\|\|_{_\square}$ is the patch norm, $R_1$, $R_2$, $q_1$, $q_2$ are constants, and $0\leq q_1<\infty$, $0\leq q_2<\infty$. Besides, the denominator in the RHS of the bound is positive.
\end{proposition}
The proof is included in the appendix in a separate report. The above error bound guarantees in theory that the segmentation error rate due to the assignment rule (\ref{EQ:SEG_RULE}) can be decreased by minimizing the reconstruction error with respect to a fixed number of bases. The number of bases will definitely smaller than the number of dimensions of the the patch, since the denominator in the RHS of the bound is positive.

To conclude, we showed that the segmentation error rate is upper bounded by the total patch reconstruction error for a fixed $K$. The minimization of the patch reconstruction error can therefore minimize the segmentation error.

\subsection{Global optimal linear patch reconstruction}\label{SEC:GD_opt_rec}
In what follows, we address the optimal linear patch reconstruction.
\begin{equation}\label{EQ:Rec_Error_K}
\begin{split}
&\{\mathbf{v}_k^{l*}\}=\argmin\limits_{\{\mathbf{v}_k^{l}\}}
\int_\Omega\mathcal{E}_l(x,y,\{\mathbf{v}_i^{l}\})H_ldxdy\\
&\hbox{s.t.~~} \forall i\neq j,\Big\langle \mathbf{v}_i^l, \mathbf{v}_j^l\Big\rangle_{\square} = 0, \forall k, \Big\|\mathbf{v}_k^l\Big\|_{\square}=1
\end{split}
\end{equation}
where $l=\{1,2\}$, $\mathcal{E}_l(x,y,\{\mathbf{v}_k^{l}\})= \|\mathbf{p}-\sum\limits_{k=l}^{K}\langle\mathbf{p},\mathbf{v}_k^l\rangle_{\square} \mathbf{v}_k^l\|^2_\square$, $\langle \mathbf{p},\mathbf{v}_k^l\rangle_{_\square}=\iint_{-m/2}^{m/2}\mathbf{p}\cdot\mathbf{v}_k^ldudv$
and $\|\mathbf{f}\|_{_\square}^2=\iint_{-m/2}^{m/2}\mathbf{f}^2(u,v)dudv$, where $m$ is the width of a patch, and we consider square patch in this paper.

A useful identity regarding this formulation is the following.
\begin{equation}\label{EQ:Proj_K}
\begin{split}
&\argmin\limits_{\{\mathbf{v}_k^{l}\}}\int\limits_\Omega\mathcal{E}_l(x,y,\{\mathbf{v}_k^{l}\})H_ldxdy
=\argmax\limits_{\{\mathbf{v}_k^{l}\}}~~ U\Big(\{\mathbf{v}_k^{l}\}\Big)
\end{split}
\end{equation}
where
\begin{equation*}
\begin{split}
&U(\{\mathbf{v}_k^{l}\})=\int\limits_\Omega \sum\limits_{k=1}^{K}\langle\mathbf{p}^l,\mathbf{v}_k^l\rangle_{\square}^2 H_l dxdy\\
&= \sum_{k=1}^{K} \int\limits_\Omega \int_\square |I(x+u,y+v)\mathbf{v}^l_k(u,v)|^2dudy H_l dxdy
\end{split}
\end{equation*}
 according to Eq. (\ref{EQ:Err_expand}), and
$l=\{1,2\}$, $\{\mathbf{v}_k^l\}$ are orthonormal. This identity can be verified by expanding the squared error.

Before we try to solve the reconstruction problem (\ref{EQ:Rec_Error_K}) (or (\ref{EQ:Proj_K}) equivalently), we also note that \emph{the optimization problem defined in Eq. (\ref{EQ:Rec_Error_K}) is a problem of minimizing constrained concave function}. The formal statements with their proofs are included in the appendix in a separate report. Such a problem is known to have local optimal solutions \cite{Rosen83MinConcave}. However, we are able to show that the global optimal solution to the linear patch reconstruction problem is attainable. The key result regarding the optimality is the following theorem.
\begin{theorem}\label{THM:BOUND}
Given a set of functions $\{\mathbf{w}_k,k=1,2,...,K\}$ defined as follows,
\begin{equation}\label{EQ:W_rep}
\left\{\begin{split}
&\mathbf{w}_k = \sum_{h=1}^N\alpha_{kh}\mathbf{e}_h, \\
&\forall i,j, \langle\mathbf{w}_i,\mathbf{w}_j\rangle_{_\square}=0, \hbox{and~}\forall i,~\|\mathbf{w}_i\|_{_\square}=1
\end{split}\right.
\end{equation}
where $\{\mathbf{e}_h,h=1,2,...,N\}$ are all the eigenvectors of $\Lambda_{Hl}$,  $\Lambda_{Hl}$ is defined by the following.
\begin{equation}
\begin{split}
&\Lambda_{Hl}(u,v,u',v')\\
&= \int I_{Hl}(x+u,y+v) I_{Hl}(x+u',y+v') dxdy
\end{split}
\end{equation}
where $I_{Hl} = I\cdot H_l$, then the following bound is true.
\begin{equation}
U(\{\mathbf{w}_k\})\leq\sum_{k=1}^{K} \lambda_{k}=U(\{\mathbf{e}_k\})
\end{equation}
where $U(\{\mathbf{w}_k\})$ is defined in Eq. (\ref{EQ:Proj_K}), $\{\lambda_{k}\big|1\leq k\leq K\}$ are the first $K$ eigenvalues of $\Lambda_{Hl}$.
\end{theorem}
The proof is included in the appendix in a separate report. The above suggests the matrix eigen-decomposition as our solution to the global optimal reconstruction. However, the eigen-decomposition typically requires converting the image to patches, computing the covariance matrix followed by Singular Value Decomposition (SVD). In our segmentation framework, we require the optimal reconstruction iteratively during the segmentation. Thus, the eigen-decomposition of the matrix obtained from the extracted and labeled patches can be time-storage consuming. Alternatively, we suggest the gradient descent as the solution. The gradient descent equation to minimize the error (\ref{EQ:Rec_Error_K}) is the following.
\begin{equation}\label{EQ:GD_min_err}
{\partial \mathbf{v}^l_n(u,v,t)\over\partial t} = \int\limits_\Omega \Big\langle I_H,\mathbf{v}^l_n\Big\rangle_{_\square} I_H(x+u, y+v) dxdy
\end{equation}

The reasons for this choice are the following.
\begin{theorem}\label{THM:EigenProc}
Given the following eigenvalue problem.
\begin{equation}
\big[\Lambda_{Hl}\big]\mathbf{e}_k = \lambda_k \mathbf{e}_k
\end{equation}
where $\lambda_k$ is the eigenvalue, $\mathbf{e}_k$ is the eigenvector and
$\big[\Lambda_{Hl}\big]\mathbf{x}=\int_{\{u,v\}} \Lambda_{Hl} \mathbf{x} dudv$, and if there are finitely many, say $N$, eigenvectors of $[\Lambda_{Hl}\big]$, the following \emph{\textbf{gradient descent procedure}} for solving (\ref{EQ:Rec_Error_K}) (or (\ref{EQ:Proj_K}) equivalently)
converges to the global optimal solution of (\ref{EQ:Rec_Error_K}), with the initial bases $\mathbf{v}_0=\sum_{k=1}^Ka_k\mathbf{e}_k$ for any $\{a_k|a_k>0,~1\leq k\leq K\}$.
\begin{equation}\label{EQ:EigenProc}
\begin{split}
&\hbox{\textbf{while}~} 1\leq n\leq K~\hbox{\textbf{do}}\\
&~~\hbox{\small\textsc{Step} $n$:}\left\{\begin{array}{l}\hbox{Solve } \mathbf{v}^l_n \hbox{ by (\ref{EQ:GD_min_err})}\\
\hbox{s.t.~}\mathbf{v}_{n}^l = \mathbf{v}_{n}^l - \sum\limits_{k=1}^{n-1}\langle \mathbf{v}_{n}^l,\mathbf{v}_{k}^{l*}\rangle_{_\square} \mathbf{v}_{k}^{l*},\\ ~~~~~\|\mathbf{v}_{n}^l\|_{_\square}=1\end{array}\right.\\
&\hbox{\textbf{end while}}
\end{split}
\end{equation}
where $\mathbf{v}_{k}^{l*}$ for $k=1,2,...,n-1$ are the solved patch bases.
\end{theorem}
The proof is included in the appendix in a separate report.
Moreover, the convergence of the gradient descent is linear as claimed in the following corollary.
\begin{corollary}\label{COR:LinCvg}
If there are $N<\infty$ eigenvectors of $\Lambda_{Hl}$, and the eigenvalues are ordered such that $\lambda_1>\lambda_1>...>\lambda_N$, the rate of convergence for the gradient descent iteration in each step of the greedy procedure defined by Eq. (\ref{EQ:EigenProc}) does not exceed $1$. In other words, the procedure converges linearly.
\end{corollary}
The proof is included in the appendix in a separate report. Besides, the gradient descent does not require computing the matrix $\Lambda_{Hl}$ or the matrix eigen-decomposition but only the convolutions.
%

\subsection{Coupling with piecewise smooth model}
It is easy to find two patches, which can be reconstructed equally well (yielding the same residue) by the same set of bases, while having significant difference in intensity. Hence, the proposed linear patch reconstruction model cannot be used for differentiating the image patches that are different only in their illumination and color but similar in their local structure. To cope with both the cases via the same model, we propose to combine our model of linear patch reconstruction with the original piecewise-smooth Mumford-Shah model to form an integrated functional model of segmentation as follows.
\begin{equation}\label{EQ:Jmodel}
\begin{split}
\min_{H} &\int\limits_\Omega \mathcal{E}^{\star}_1H+\mathcal{E}^{\star}_2(1-H)dxdy+\nu\int_\Omega \|\nabla H\|dxdy
\end{split}
\end{equation}
where $\mathcal{E}^{\star}_l = \alpha(I-g_l)^2+(1-\alpha){\mathcal{E}^{\square}_l}$, for $l=1,2$ and $0\leq\alpha\leq1$ is a predefined weight. The last term penalizes the complexity of the segmentation.

The numerical solution of $H$ to the energy minimization model can be derived from the level set method, in which a signed distance function $\phi$ is used to generate $H$ through Heaviside function, i.e. $H = H(\phi)$.
\begin{equation}\label{EQ:CE_Joint}
{\partial \phi\over \partial t}=-(\mathcal{E}^{\star}_1-\mathcal{E}_2^{\star})\delta(\phi)+\nu \mathrm{div}\left({\nabla\phi\over\|\nabla\phi\|}\right)
\end{equation}

By this equation, it is implied that the reconstruction errors $\mathcal{E}^{\star}_1$ and $\mathcal{E}^{\star}_2$ are fixed when implementing curve evolution. The energy in (\ref{EQ:Jmodel}) is minimized by alternatively updating the image partition by solving $H$ and updating the reconstruction error in each region by solving $g_l$ and $\{\mathbf{v}_k^l\}$. This energy minimization procedure is in fact the gradient descent method for the separate variables. Hence, the convergence of the procedure is guaranteed. 

\section{Experiments}\label{SEC:Exp}

\subsection{Data preparation and implementation details}
\begin{figure}
\centering
  \subfloat[]{\includegraphics[width=0.4\columnwidth]{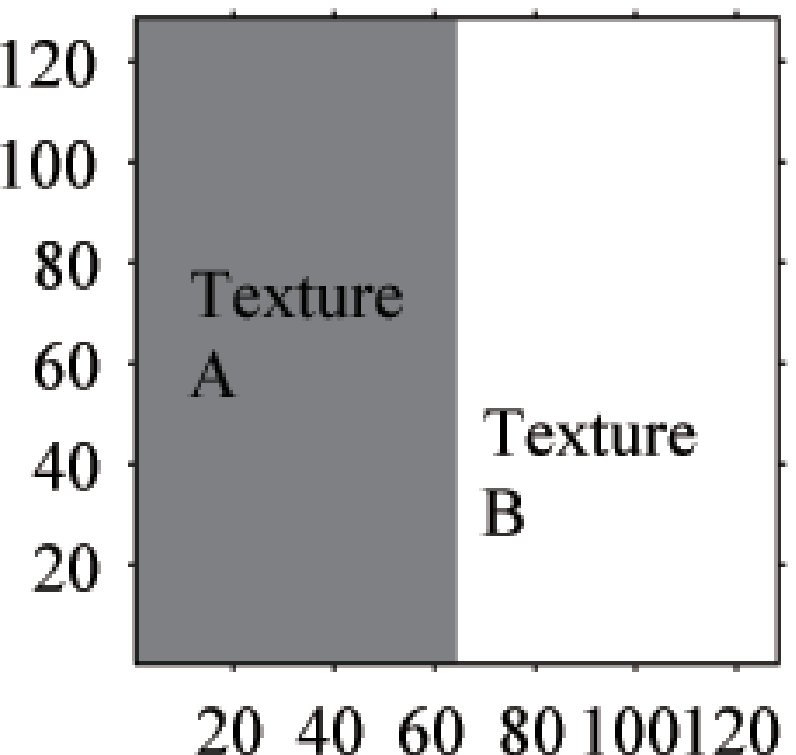}}~
  \subfloat[]{\includegraphics[width=0.4\columnwidth]{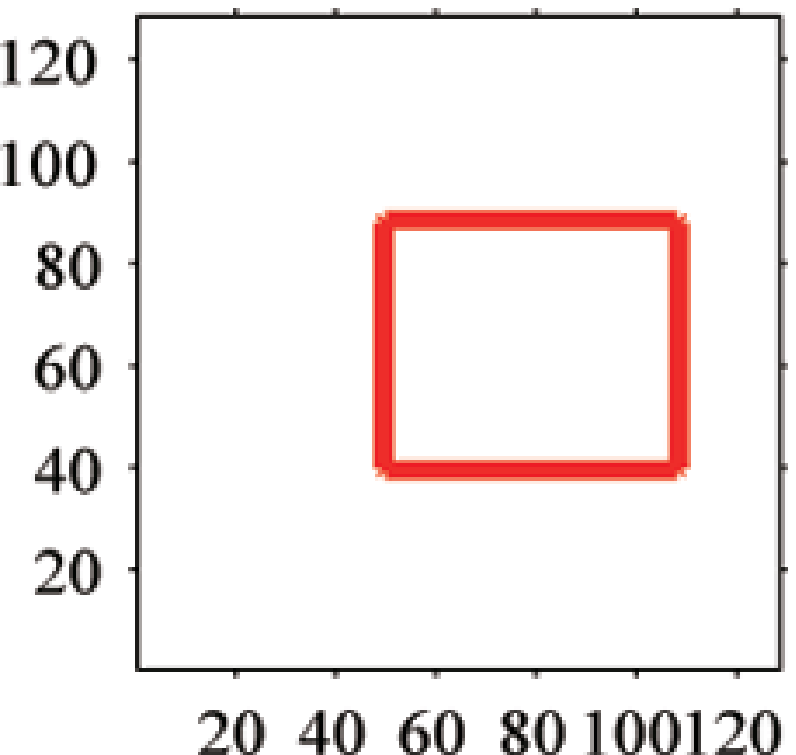}}\\
  \caption{(a) is the template for texture mosaicing. (b) is the initial contour for curve evolutions.}\label{FIG:GTIni}
\end{figure}

Natural textures are examples of the non-smooth structured visual patterns. Evidences show that the texture patches can be reconstructed (modeled or represented) well by patch subspaces, i.e. the eigenfilters, as reported in \cite{Randen99FilterTextureReview} where the texture classification was investigated more carefully. Therefore, we mainly evaluate our method of segmentation by piecewise linear patch reconstruction on a subset of Brodatz textures \cite{BrodatzTexture}. We empirically choose the textures that appear relatively \emph{spatially regular} with \emph{similar size} of texture stimuli (textons) for evaluation. Therefore, we obtain a collection of textures: $\{$D3, D5-6, D15-22, D24, D34-36, D49, D52-53, D55, D57, D65, D68, D76-77, D79, D81-85, D101-106$\}$, which we call the set $\mathbf{S}$. Afterwards, we generate two sets of mosaic texture images by paring all different the textures from $\mathbf{S}$ for evaluation our segmentation method. Each set contains 1260 images. One of the sets is made by the original textures, the other is by the textures with the mean intensity subtracted. The template for paring the textures is shown in Figure \ref{FIG:GTIni}(a). We also use this template as the ground truth for evaluating the segmentation. Both the sets are challenging for segmentation, especially the second one.

Our segmentation algorithm for minimizing the functional energy (\ref{EQ:Jmodel}) can be implemented by alternating an algorithm of image partitioning and the patch reconstruction via either the SVD or the gradient descent procedure in (\ref{EQ:EigenProc}) . We adopt the curve evolution governed by Eq. (\ref{EQ:CE_Joint}) as the image partitioning algorithm in the implementation. All the methods in our experiments are based on the curve evolution for fair comparison. For piecewise smooth image we may choose a small penalty coefficient for the contour length, e.g. $1$ in our implementation, for non-smooth image we require a large penalty of the contour length for coping with the randomness, e.g. $100$ in our implementation. We use the maximum number of iterations to detect the convergence of the curve evolution algorithms. The maximum iteration number is set to be 600, since it is observed that the curve evolutions in the experiment converge before this iteration number is reached. The convergence of the gradient descent method for patch reconstruction is fast. We set the maximum iteration number to be 5.

\subsection{Evaluation of the gradient descent patch reconstruction}
In this subsection, we evaluate the proposed gradient descent method for computing the optimal bases. The principle is to compare the error of reconstruction by gradient descent with the error of reconstruction by SVD. We evaluate the gradient descent method for linear patch reconstruction on the Set $\mathbf{S}$ and Yale face database. We compare the reconstruction errors according to the first 1-20 optimal bases produced by the gradient descent method with the errors according to the bases produced by SVD. The reconstruction errors can be computed by evaluating (\ref{EQ:Rec_Error_K}) for the entire image domain. We also present the comparison of the reconstructions by orthogonalized Gabor filter and the SVD solution. We apply the principle of maximum filtering response to select the first $20$ Gabor filters from a filter bank of $40$ filters. We then orthogonalize the selected filters to compute the energy. We compare the averaged errors but not the total error. The total error is the sum of the reconstruction errors for all the patches in the image. The averaged error is the total error divided by the number of patches.

\begin{figure}
\centering
  \subfloat{\includegraphics[width=0.25\columnwidth]{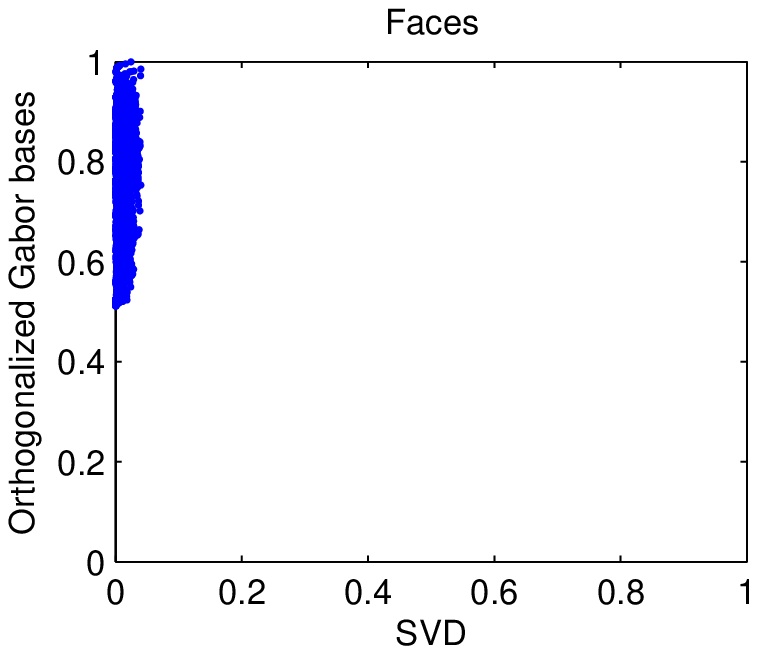}}
  \subfloat{\includegraphics[width=0.25\columnwidth]{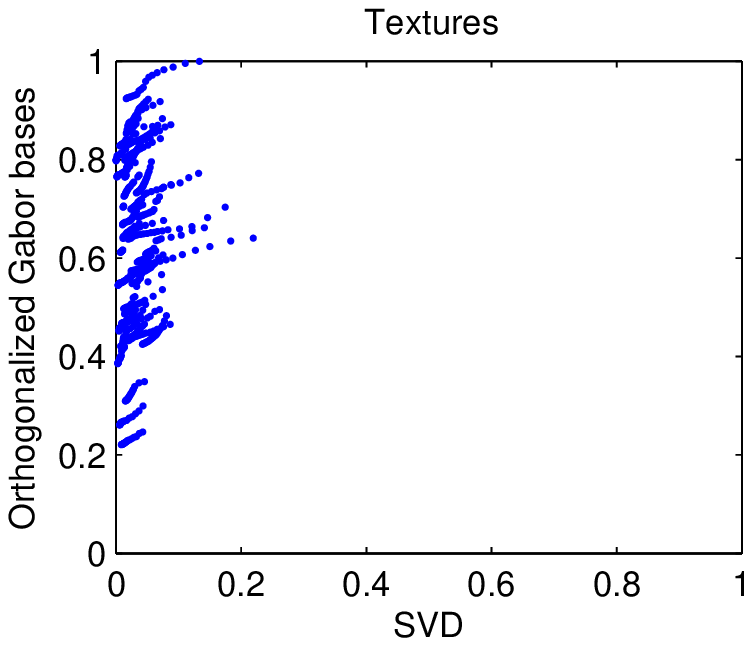}}
  \subfloat{\includegraphics[width=0.25\columnwidth]{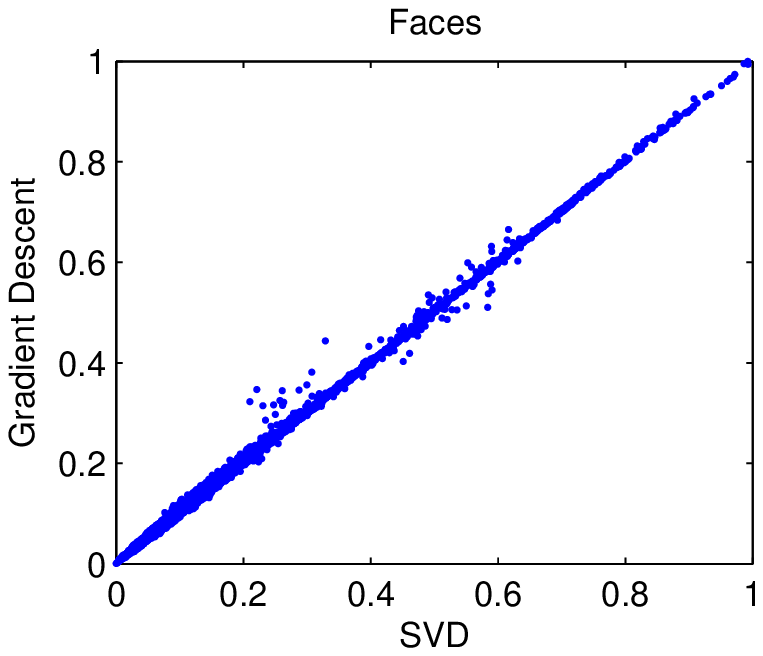}}
  \subfloat{\includegraphics[width=0.25\columnwidth]{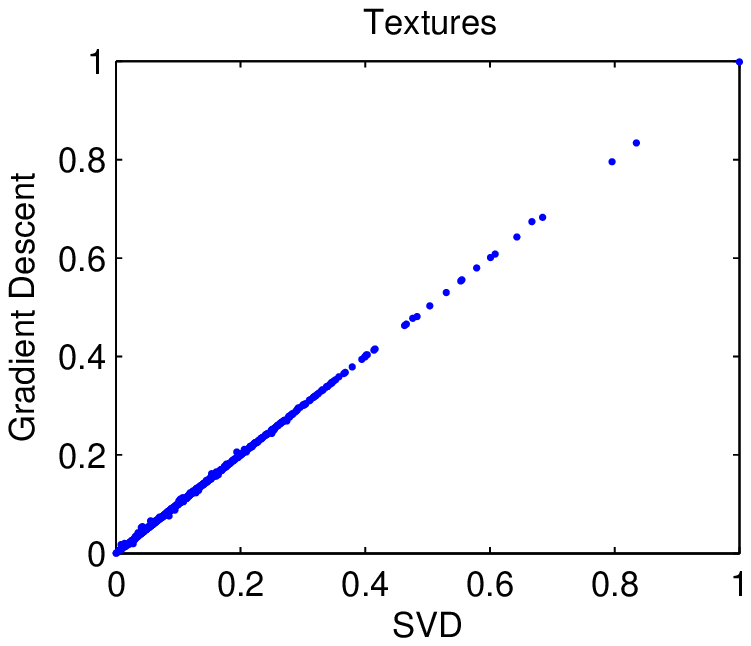}}  \caption{The scatter plots of the \emph{\textbf{normalized}} reconstruction errors for each face or texture image by the first 1-20 bases. The value 1 corresponds to the worst reconstruction, and 0 corresponds to the perfect reconstruction. The left two plots are the orthogonalized Gabor bases (vertical) vs. SVD (horizontal); The other two are the Gradient descent (vertical) vs. SVD (horizontal). Results by gradient descent is almost as the same as SVD for different numbers of bases.}\label{FIG:Grad_vs_SVD}
  \vspace{-5pt}
\end{figure}

The results are shown in Figure \ref{FIG:Grad_vs_SVD}. When visualizing the comparison of the errors, we divide the errors by the value of the maximum averaged errors to form a \emph{normalized error}. The results by gradient descent are very close to that of the SVD, while the results of Gabor filters do not match the SVD. The initial bases for the gradient descent method are randomly generated.

\subsection{Evaluation of the segmentation by linear patch reconstruction}

We mainly evaluate the segmentation of our methods, i.e. the SVD and GD based methods, on the two subsets of mosaic textures introduced previously. We compare our methods with the PS Mumford-Shah model \cite{Jerome07SolveMumfordShah} \cite{Brox2009InterpMumfordShah}, the Region-Scalable Fitting (RSF), a.k.a. the Local Binary Fitting (LBF) \cite{Li08LBF}, the Gabor filtering based method \cite{Sandberg2002level} which is also adopted in \cite{Sagiv06Texture} and the local histogram based Chan-Vese model \cite{NiChan09CVHist}, which we denote as HistPC henceforth. The Gabor filtering based method could be viewed as a baseline approach for texture segmentation, and the local histogram-based method is the state-of-the-art approach. We use 8 bases for each region in SVD and GD. We choose the 8 Gabor filters from a bank of 24 filters for each region according to the criterion of maximum filtering response. This filter selection criterion appears like model fitting \cite{Roth09FoE}. The patch size is $13\times 13$. The methods are all based on curve evolution. We used a common initial curve for the curve evolutions. The initial curve in the image domain is shown in Figure \ref{FIG:GTIni}(b). Note that the initial contour crosses the true boundary of the two regions, and the converged contour is expected to outline the region of texture B on the right.

\begin{figure}
\centering
  \subfloat{\includegraphics[width=0.49\columnwidth]{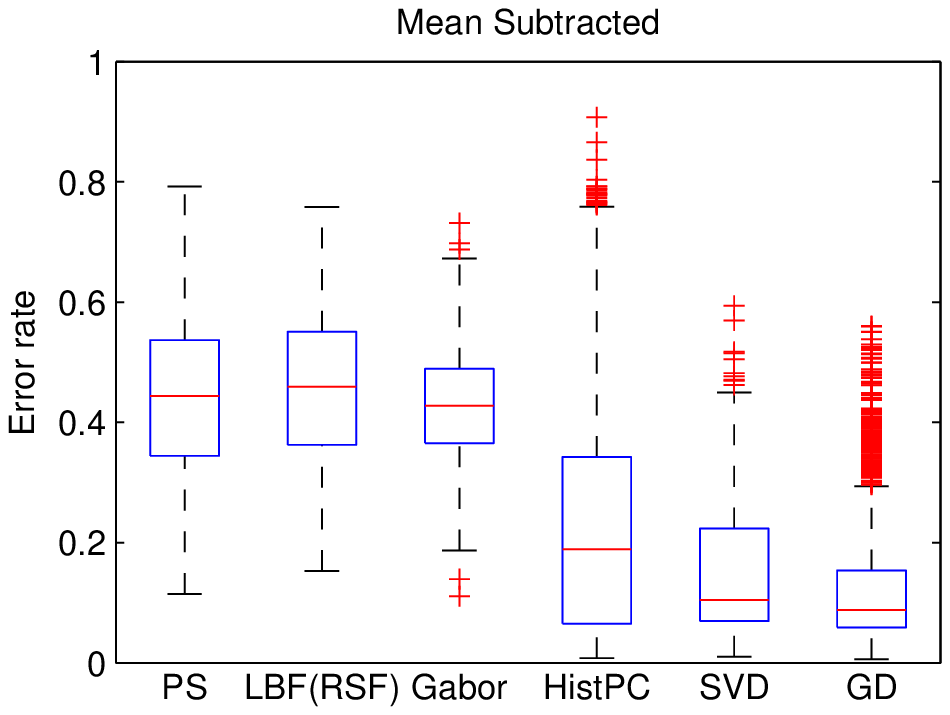}}
  \subfloat{\includegraphics[width=0.49\columnwidth]{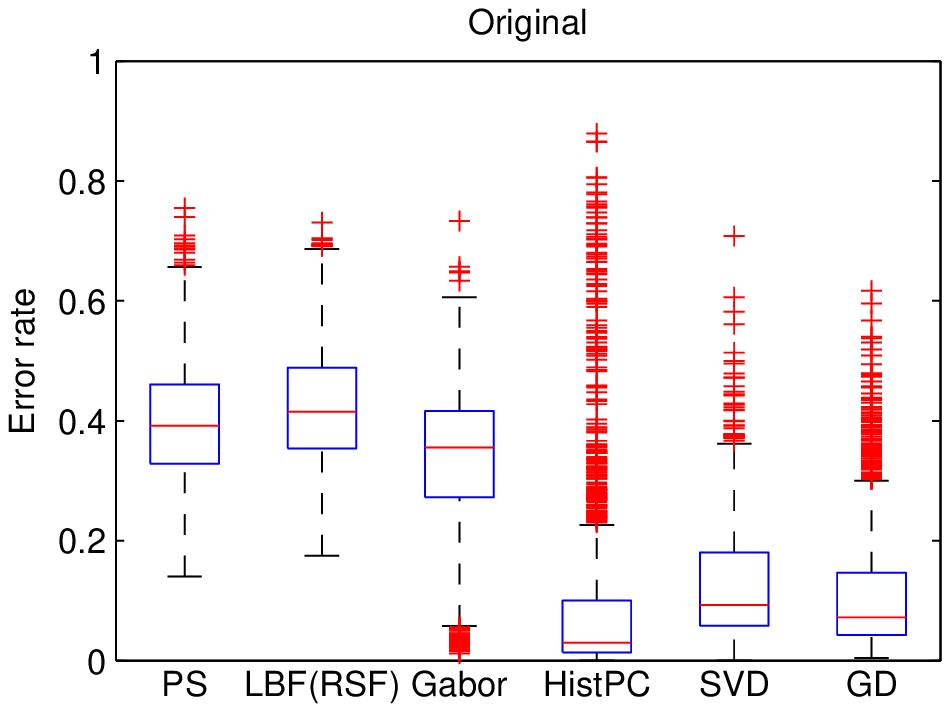}}\\
  \caption{Comparison of segmentation error rates on the images with the mean intensity subtracted (left), and on the images of original textures (right).}\label{FIG:Error_rates}
\end{figure}

\begin{figure}
\centering
  \subfloat{\includegraphics[height=47pt]{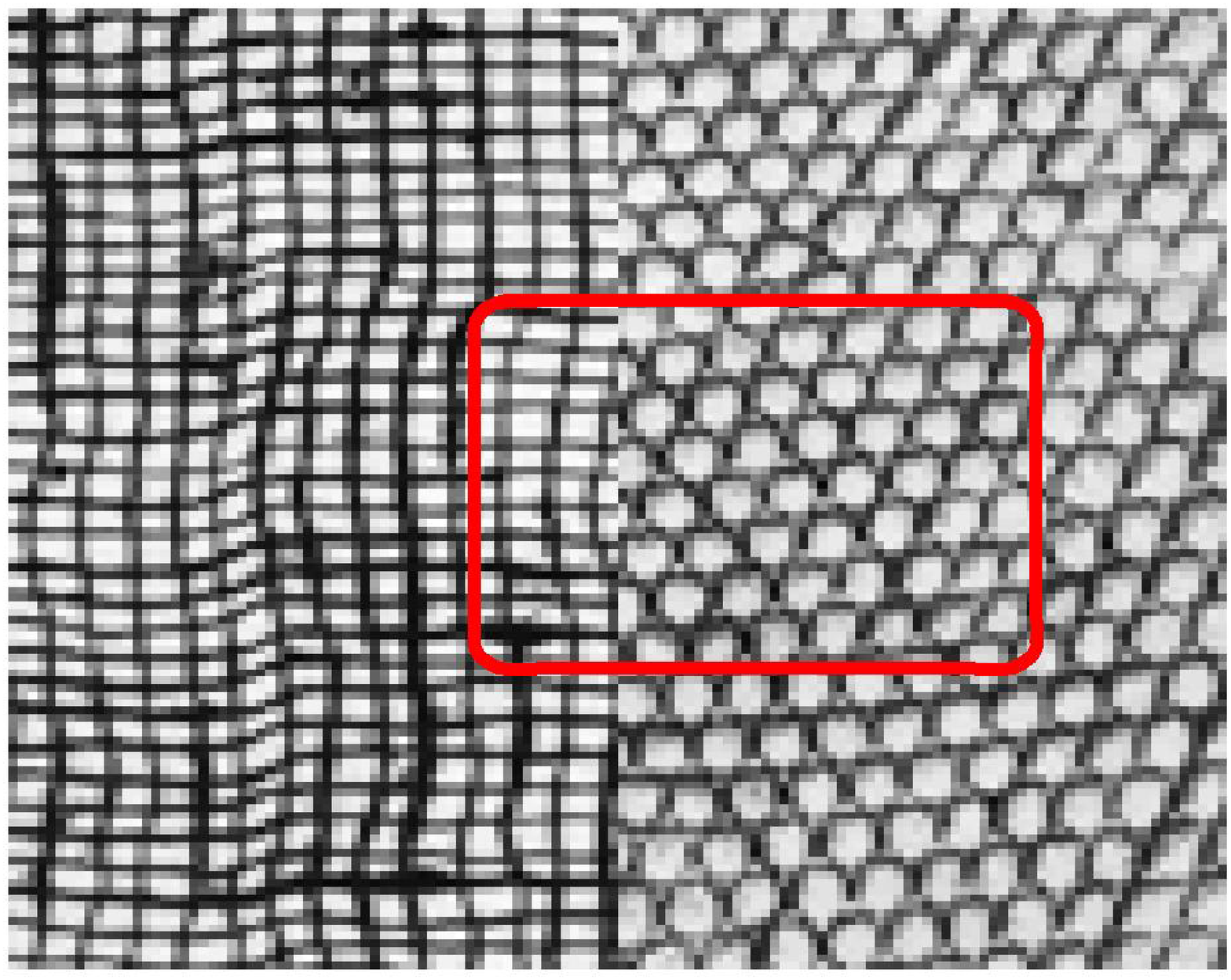}}
  \subfloat{\includegraphics[height=47pt]{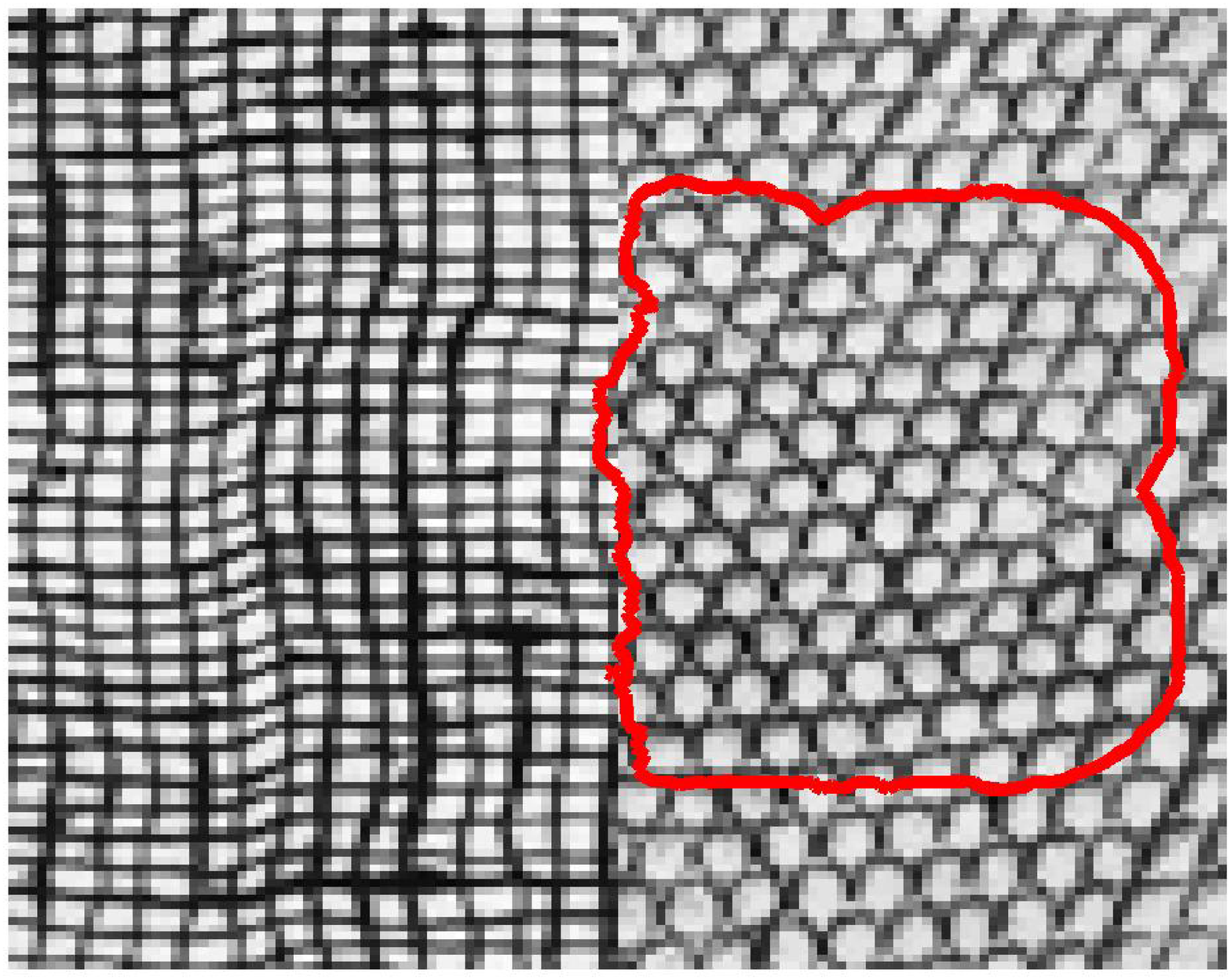}}
  \subfloat{\includegraphics[height=47pt]{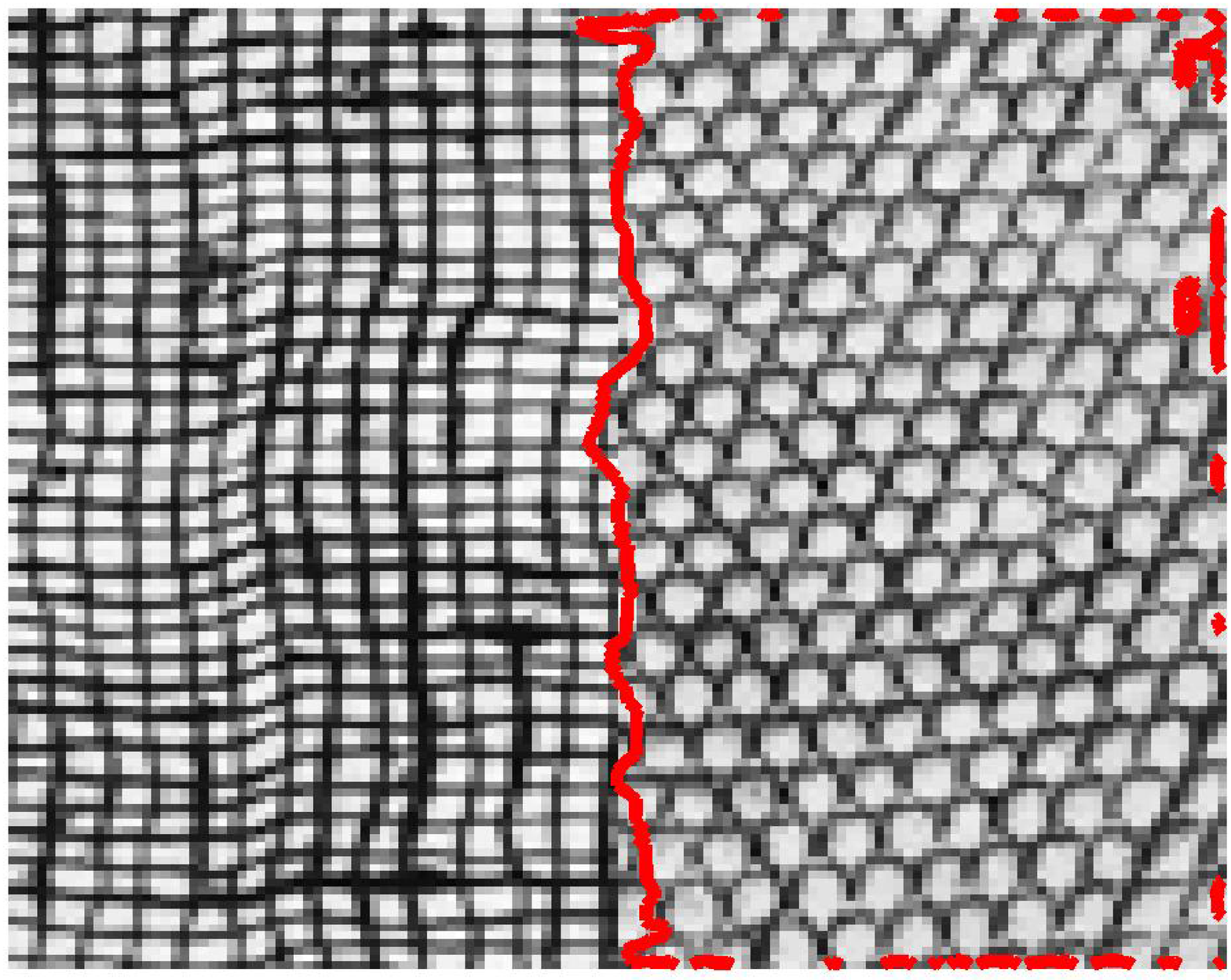}}
  \subfloat{\includegraphics[height=47pt]{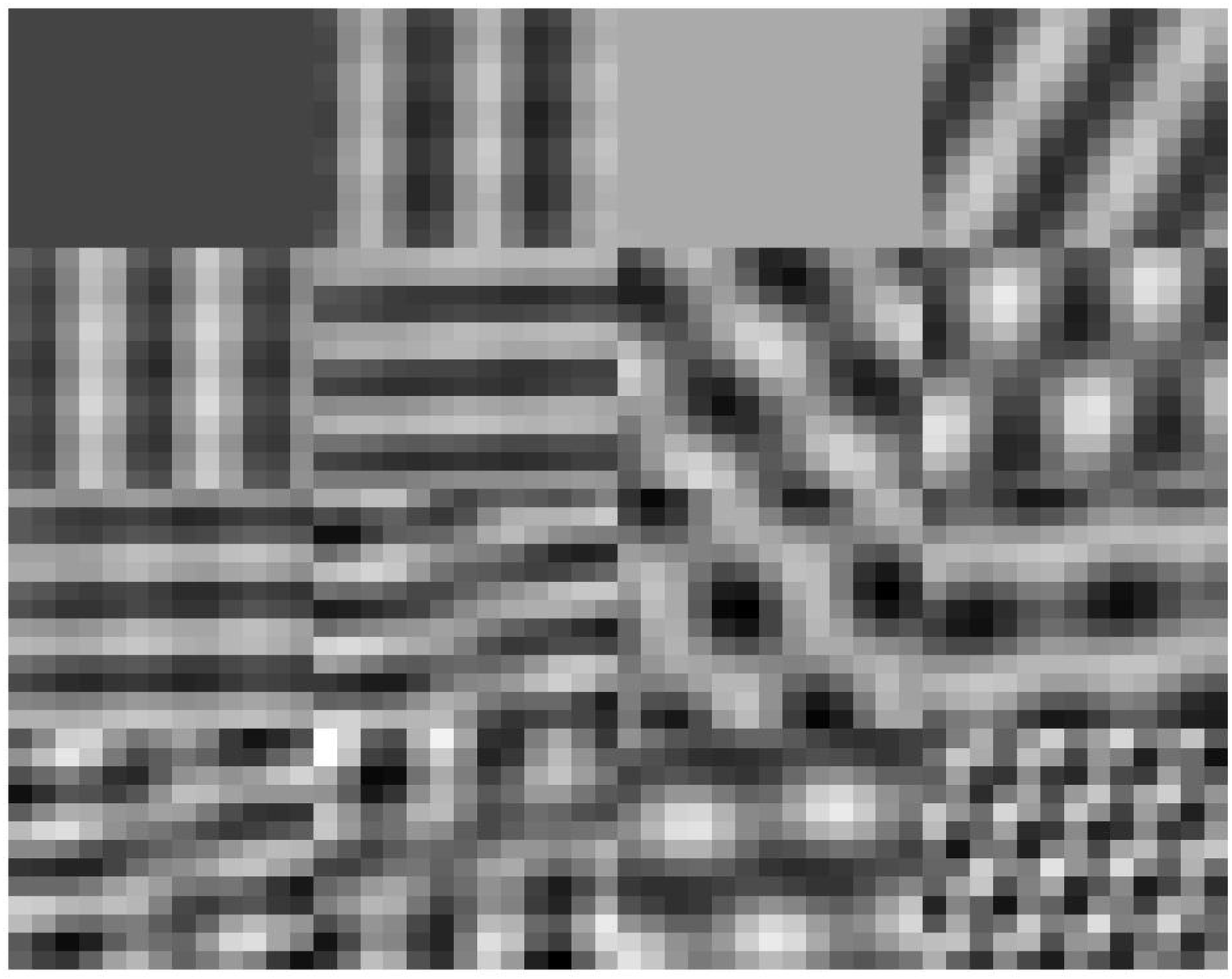}}\\\vspace{-5pt}
  \subfloat{\includegraphics[height=47pt]{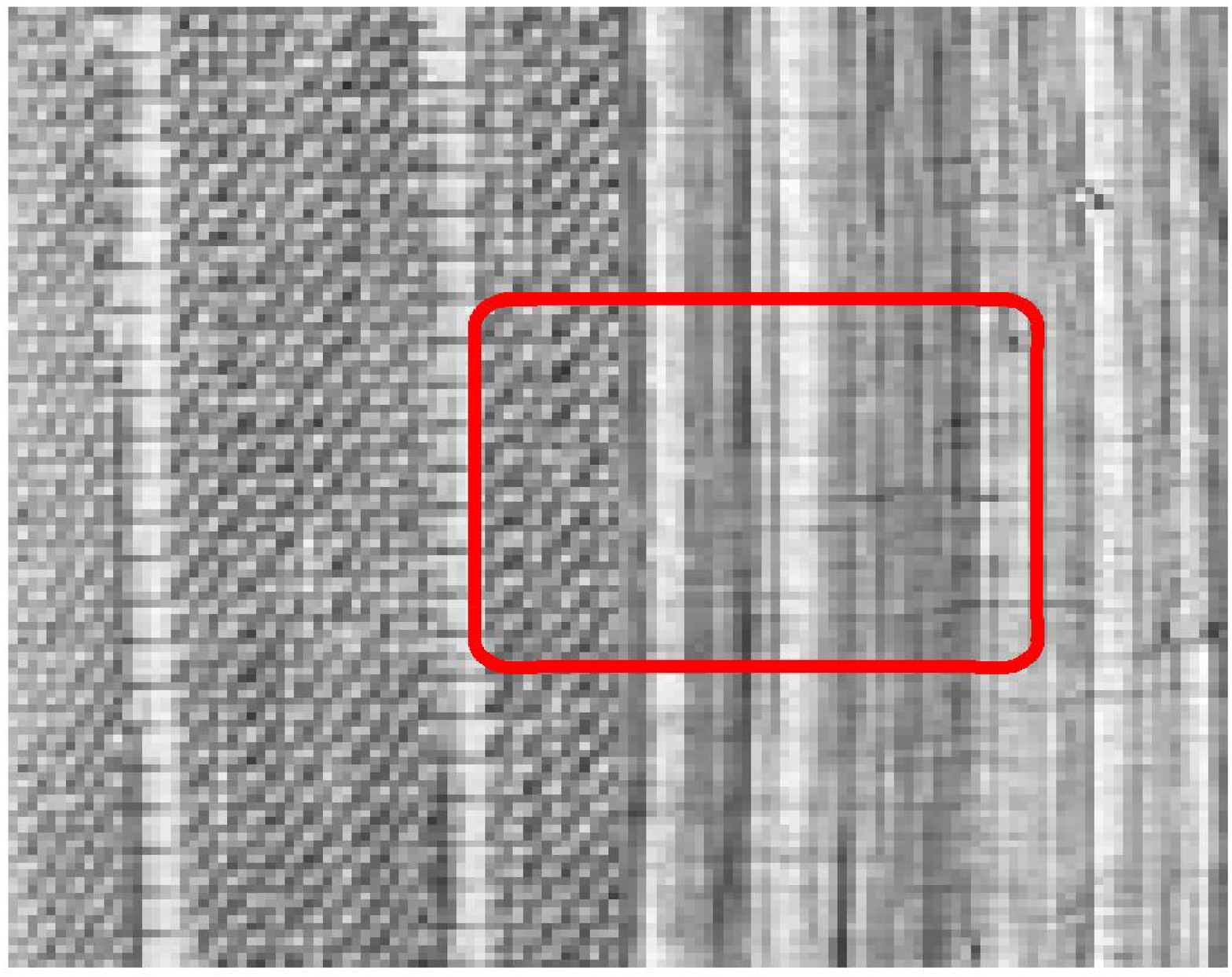}}
  \subfloat{\includegraphics[height=47pt]{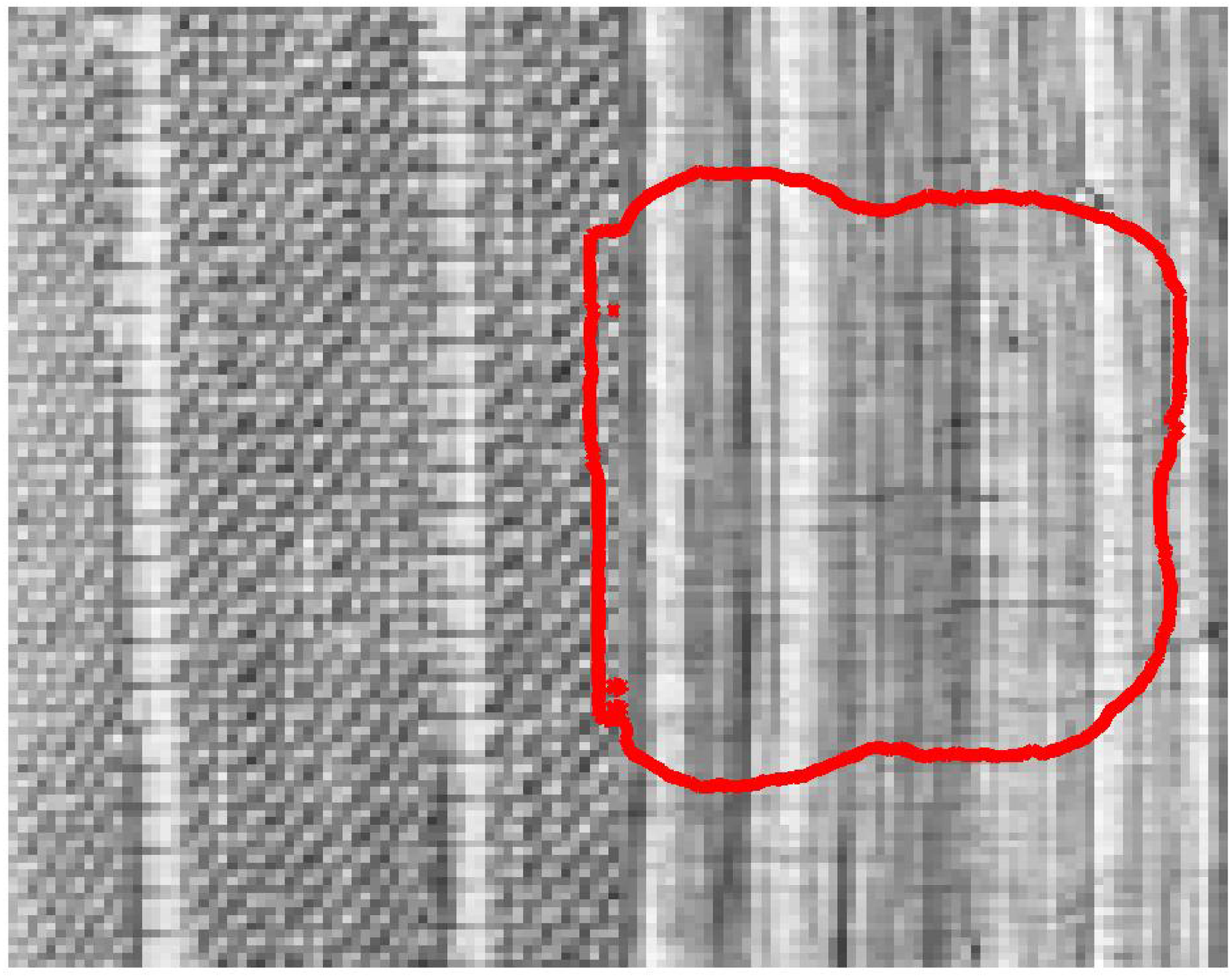}}
  \subfloat{\includegraphics[height=47pt]{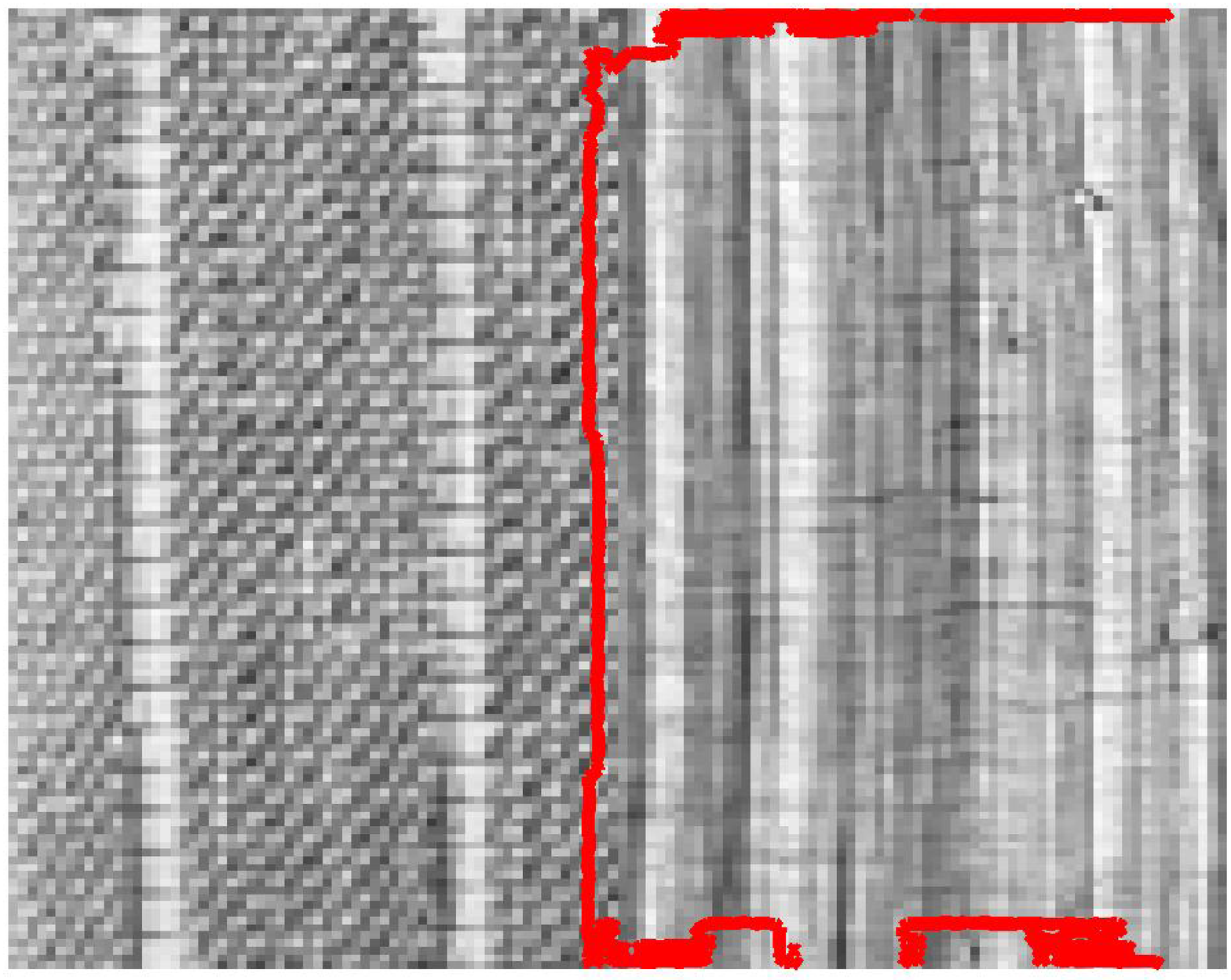}}
  \subfloat{\includegraphics[height=47pt]{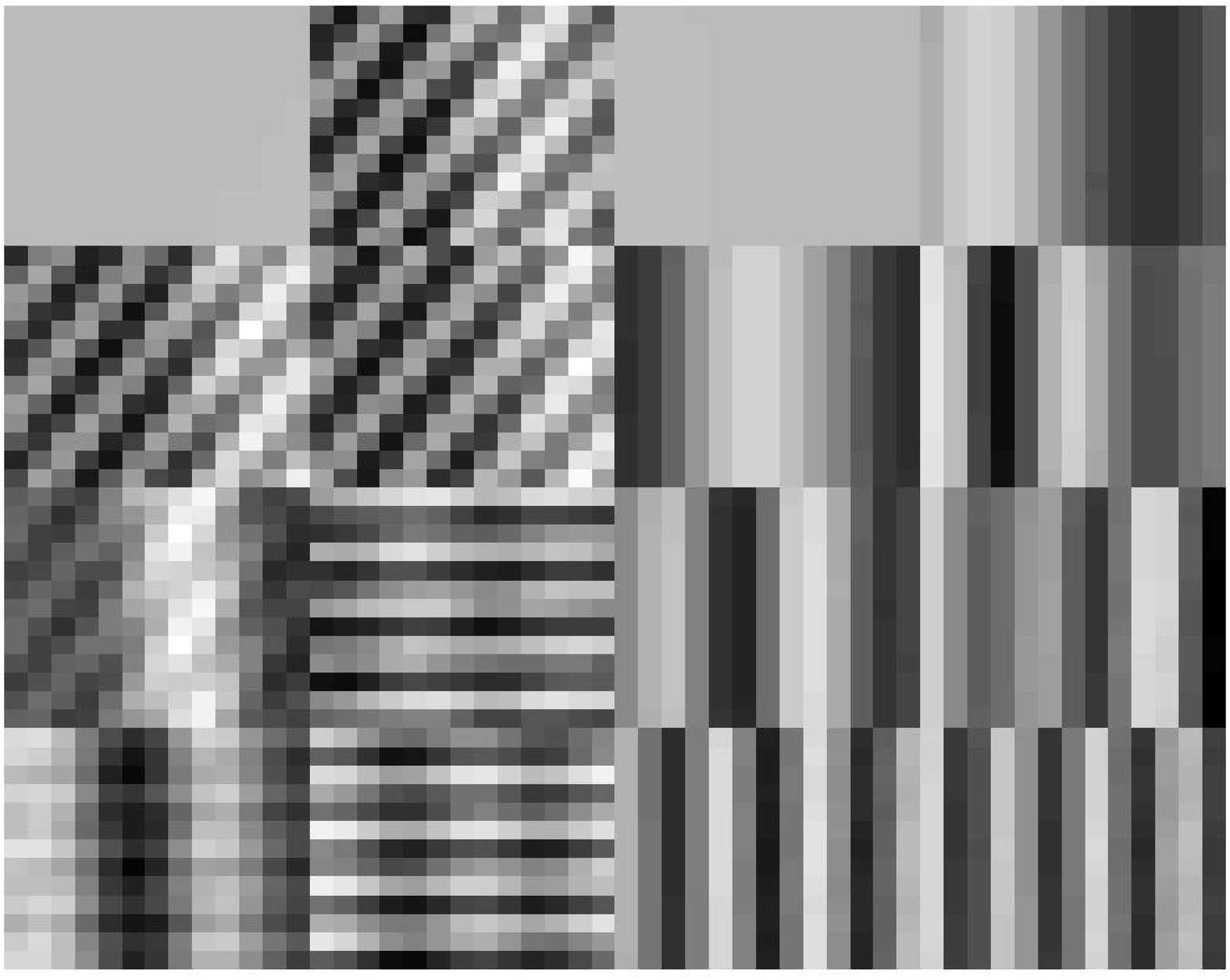}}\\\vspace{-5pt}
  \subfloat{\includegraphics[height=48pt]{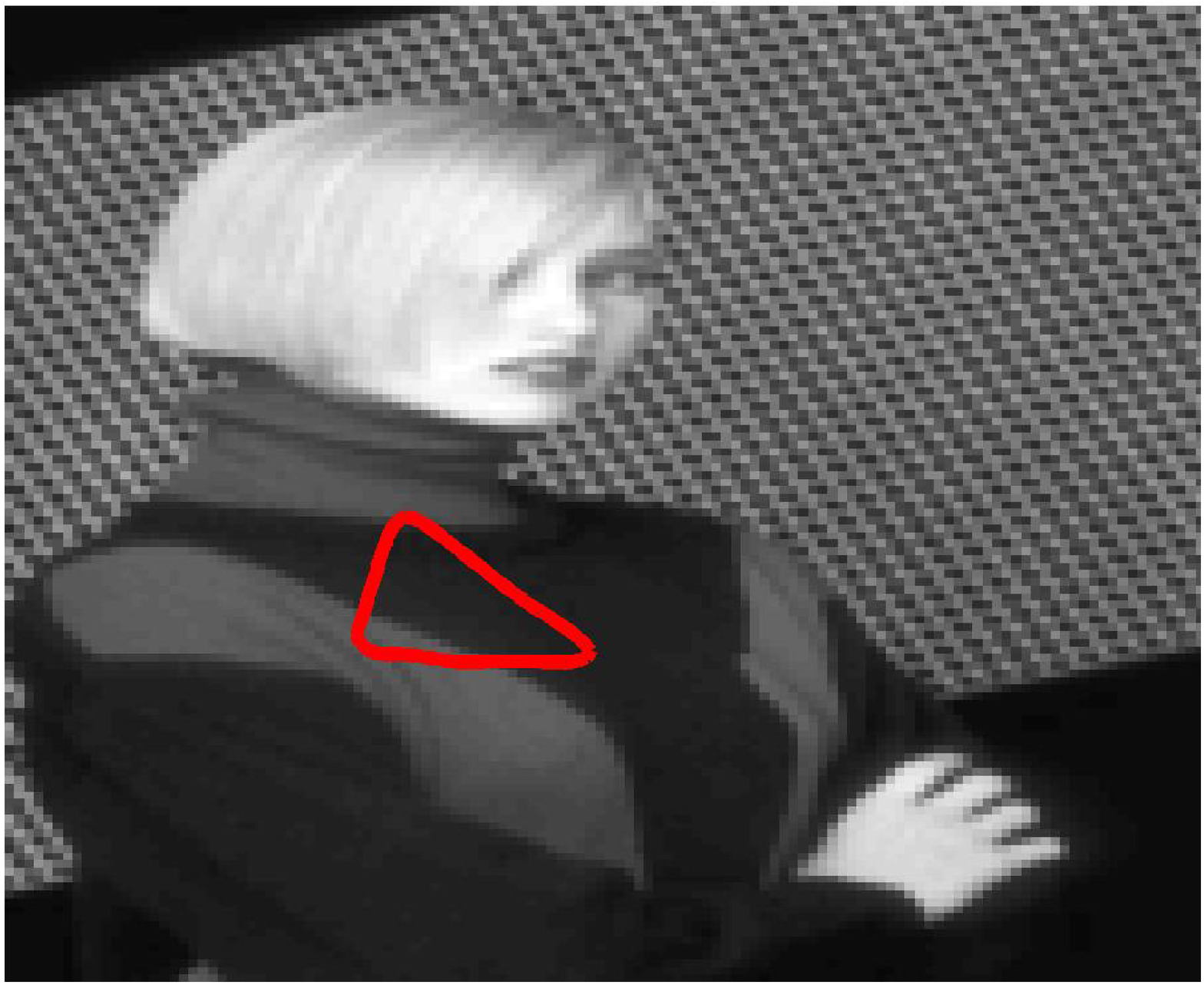}}
  \subfloat{\includegraphics[height=48pt]{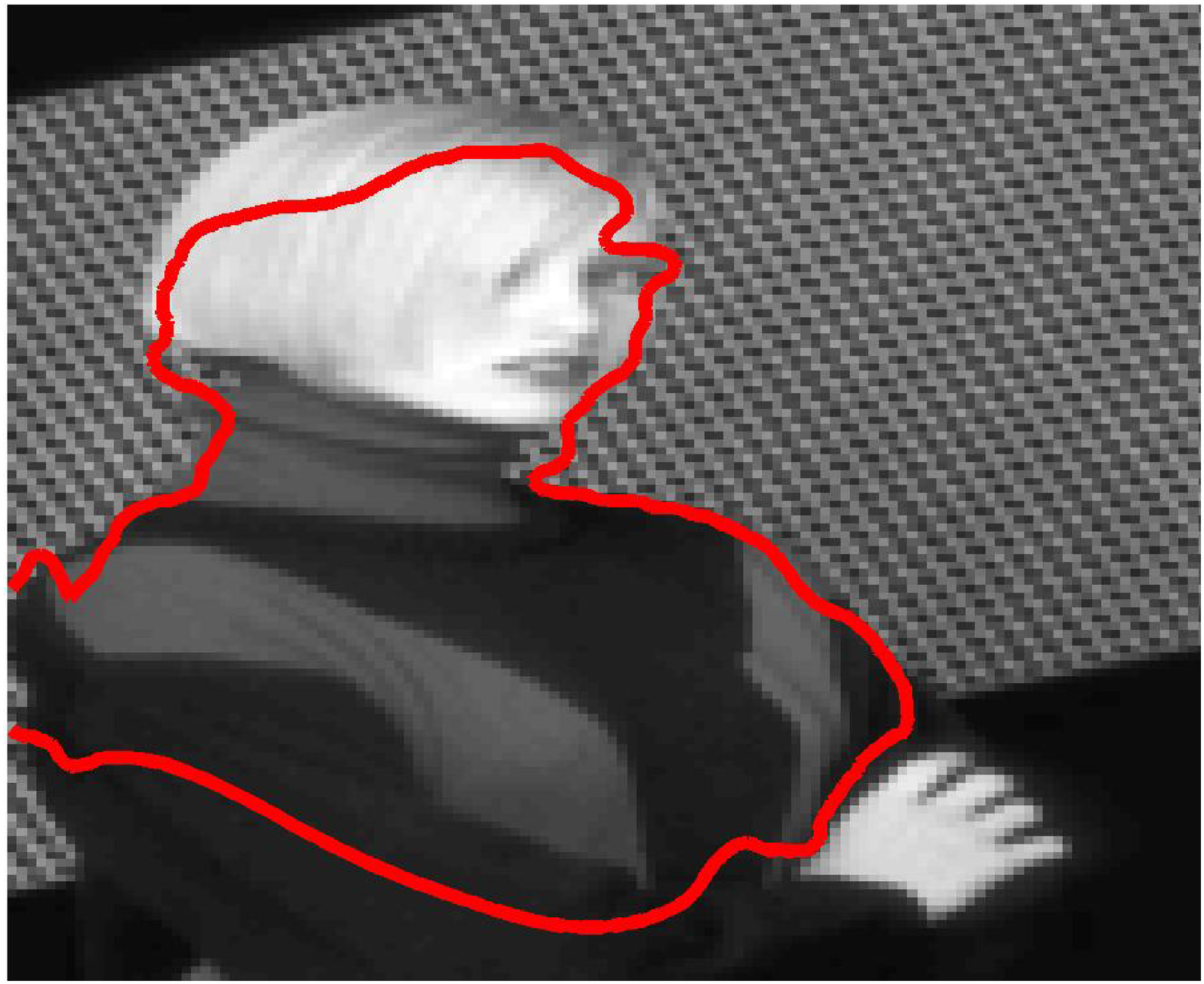}}
  \subfloat{\includegraphics[height=48pt]{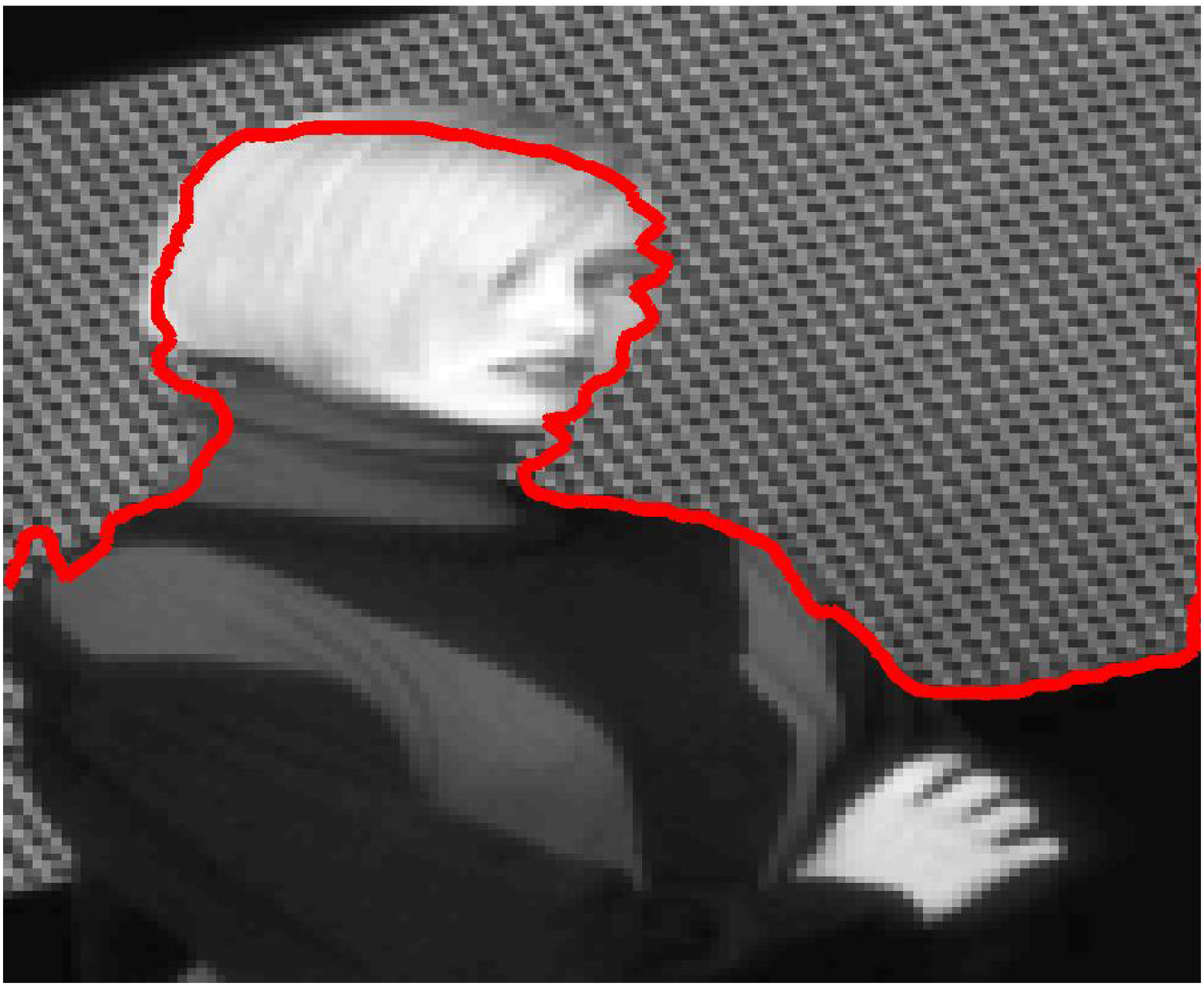}}
  \subfloat{\includegraphics[height=48pt]{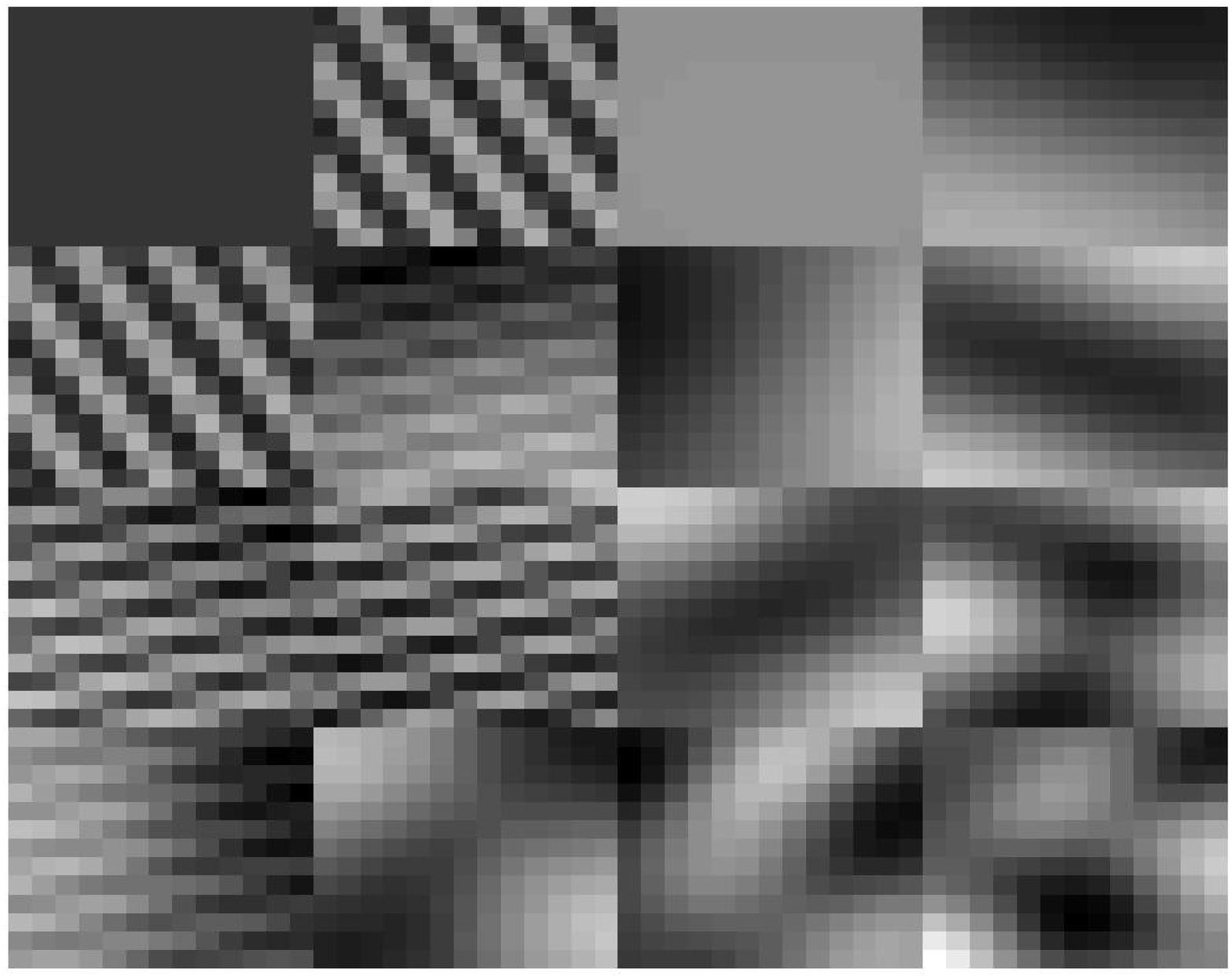}}\\\vspace{-5pt}
\caption{Curve evolution on the Brodatz D104-22 pair (top row) and D85-106 pair (mid row) and a real image taken from Berkeley dataset with the corresponding converged patch bases. The curves are drawn in red (better to view in color). The left/right two columns of the filters correspond to the background/foreground regions. The D22, D106 and the lady are assumed to be the foregrounds. }\label{FIG:show1}
\end{figure}

\begin{figure}
\centering
\subfloat[PS]{\includegraphics[height=48pt]{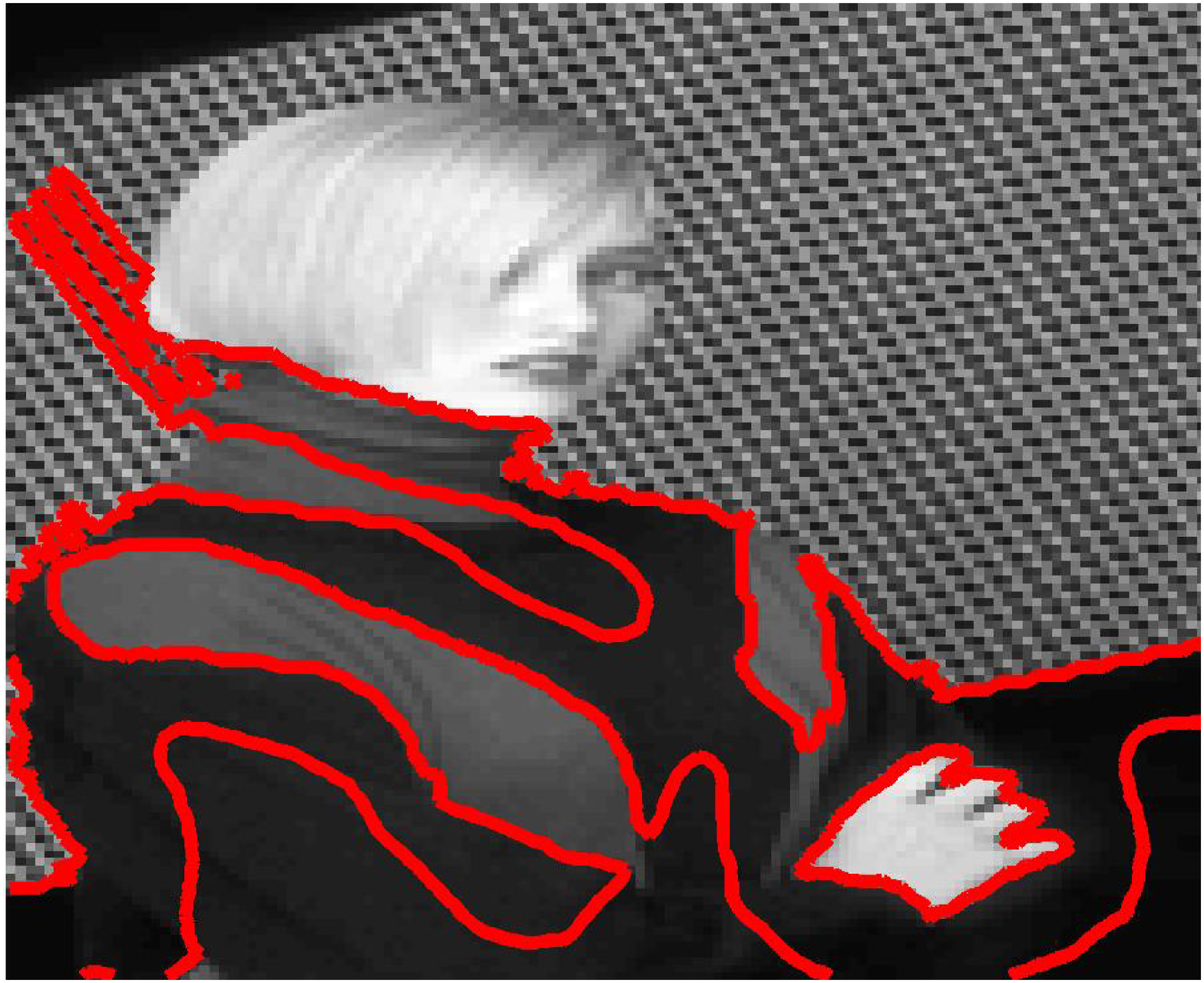}}
\subfloat[LBF]{\includegraphics[height=48pt]{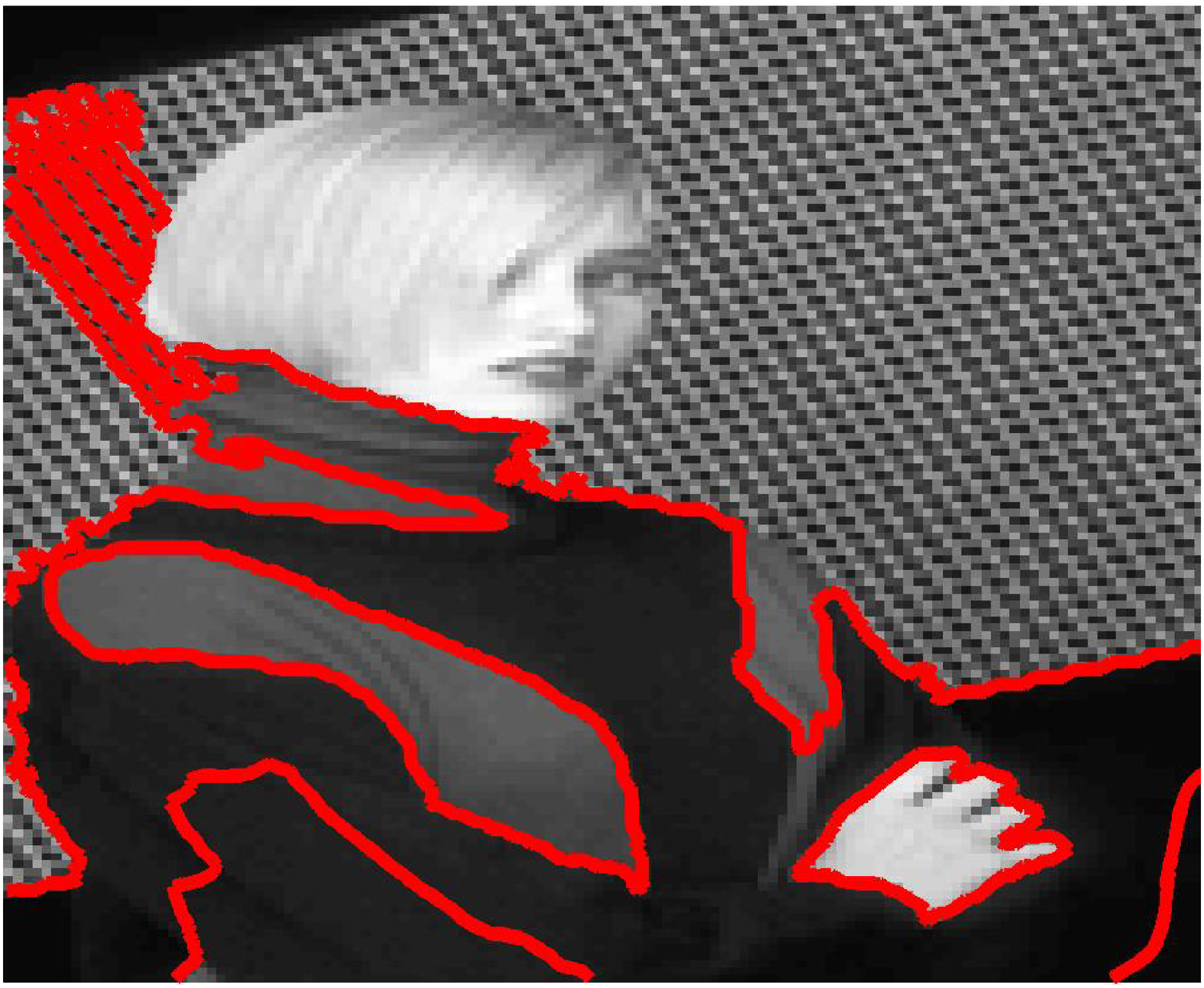}}
\subfloat[Gabor]{\includegraphics[height=48pt]{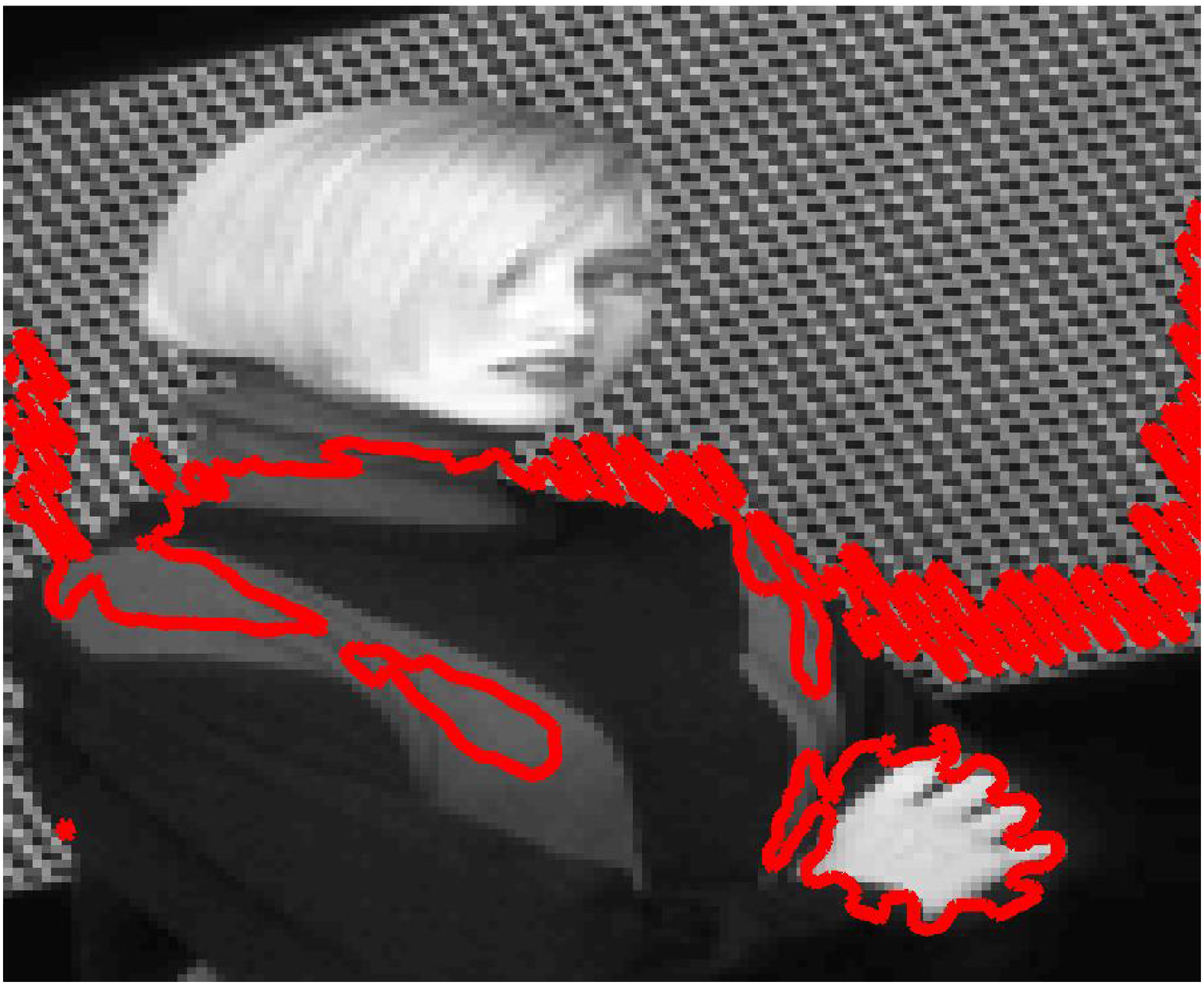}}
\subfloat[HistPC]{\includegraphics[height=48pt]{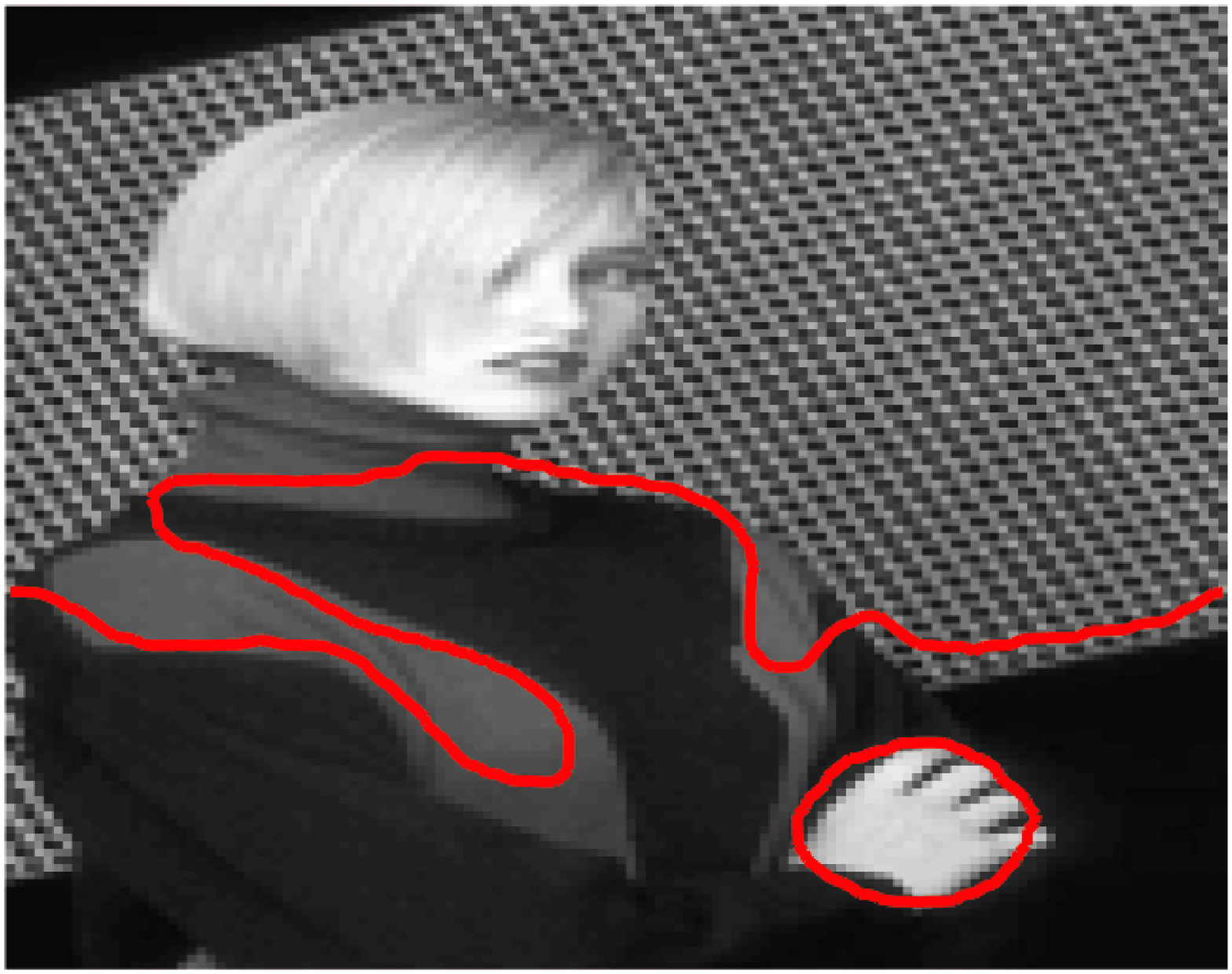}}
\caption{The converged curve evolution results of other methods with the same initialization}\label{FIG:girl_com}
\end{figure}

We adopt the pixel-wise segmentation error rate to measure the quality of the segmentation by using the ground truth. The boxplot in Figure \ref{FIG:Error_rates} shows the segmentation errors for all the methods. We can observe a clear lower error rate by our methods for differentiating the textures that differs only in their structures but not in their intensities. The results by GD and SVD are comparable, but the computational time for the segmentation based on SVD is \textbf{0.145} $\pm$ \textbf{0.048} seconds per iteration for the two datasets, while the computational time for the segmentation based on GD is \textbf{0.137} $\pm$ \textbf{0.006} seconds. This suggests us to use the GD based bases updating scheme for segmentation. The mean error by the local histogram based Chan-Vese active contour for the original textures is small, but the variation of its performance is large. Besides, this local histogram based method is still ineffective for differentiating the different textures having the same mean intensity. We can also observe from the results that Gabor features can deal with textures when the textures differ in their intensities. However, when there is little difference in the intensities of textures, the Gabor features are still powerless. We summarize the quantities in Table \ref{TB:Err_Time}. Besides, we visualize the curve evolutions, as well as the corresponding converged patch bases, for two textured images and a picture from the Berkeley dataset \cite{MartinFTM01} in Figure \ref{FIG:show1}. The picture is composed of different structures, and the piecewise linear patch reconstruction is capable of differentiating the non-smooth structure from the piecewise smooth structure. We may observe that the descriptors are semantic. The results by other models are shown in Fig. \ref{FIG:girl_com}.


\begin{table*}
\caption{Comparison of segmentation error summarized from Figure \ref{FIG:Error_rates}.}\label{TB:Err_Time}
\vspace{-5pt}
\begin{tabular*}{\textwidth}{@{\extracolsep{\fill}}cc|cc|cccc}\hline
&             &\multicolumn{2}{c|}{Our methods}& \multicolumn{4}{c}{Others'}\\\hline
& Data        & SVD   & GD    & LBF   & PS    &  HistPC   & Gabor\\\hline
\multirow{2}{*}{Error(\%)}& Original    & \textbf{12.56$\pm$10.10}  &  \textcolor[rgb]{0.00,0.00,1.00}{\textbf{11.66$\pm$11.15}} &  42.78 $\pm$10.30 & 40.00$\pm$10.52  & 9.96$\pm$15.85 & 33.38$\pm$12.94\\
&No mean    & \textbf{14.76$\pm$10.44}   &  \textcolor[rgb]{0.00,0.00,1.00}{\textbf{13.33$\pm$11.82}} &  45.64 $\pm$12.87 & 43.81$\pm$13.50 & 24.12$\pm$21.56 &   42.68$\pm$8.02 \\\hline
\end{tabular*}
\end{table*}

\subsection{Segmentation by the coupled model with parameter tuning}
\begin{figure}
\centering
  \includegraphics[width=0.45\textwidth]{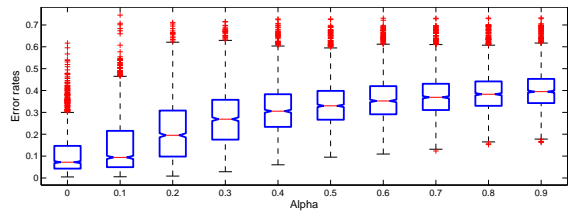} \caption{The box plot of segmentation error against $\alpha$}\label{FIG:Alpha_Err}
\end{figure}

\begin{figure}
\centering
  \includegraphics[width=0.15\textwidth]{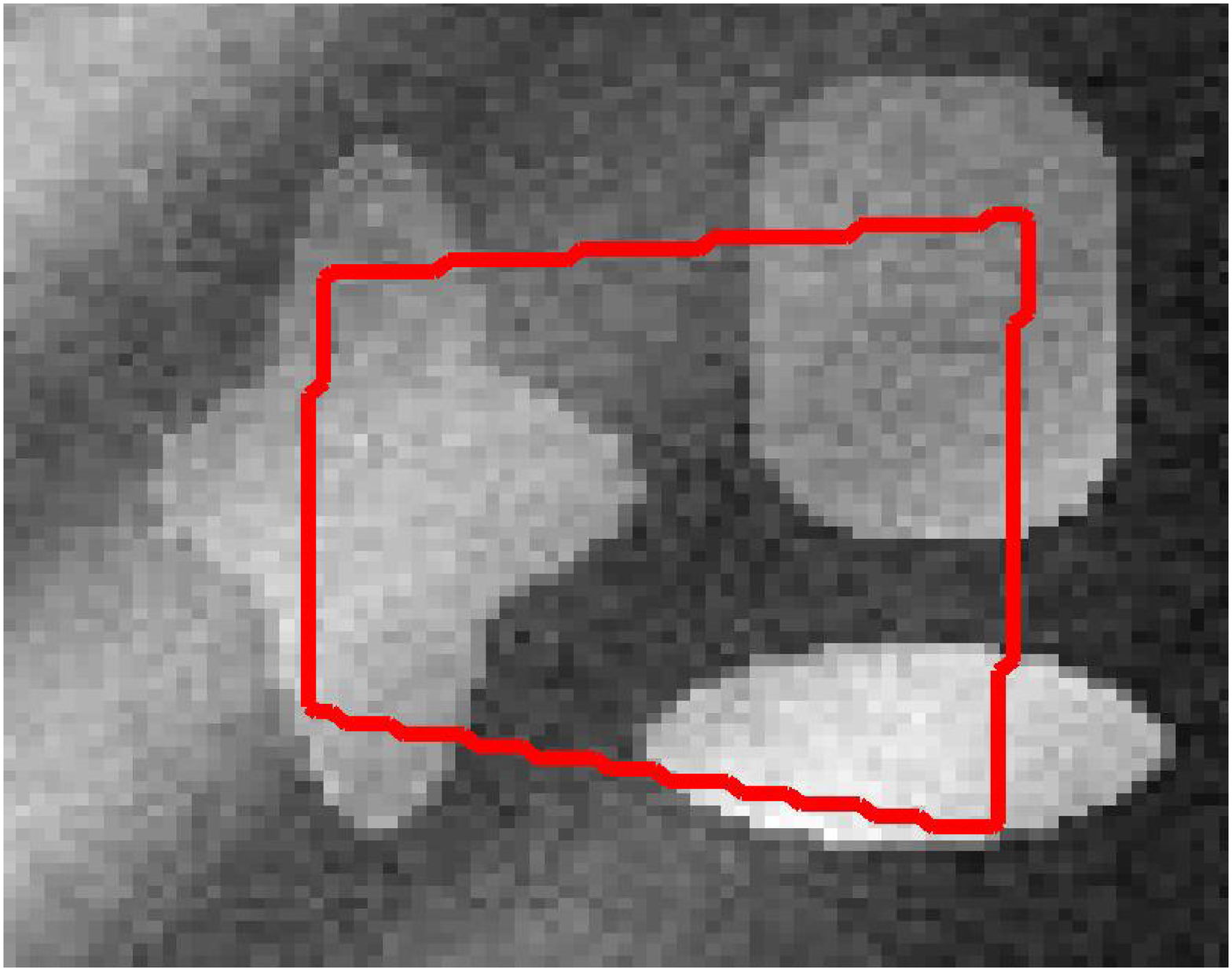}   \includegraphics[width=0.15\textwidth]{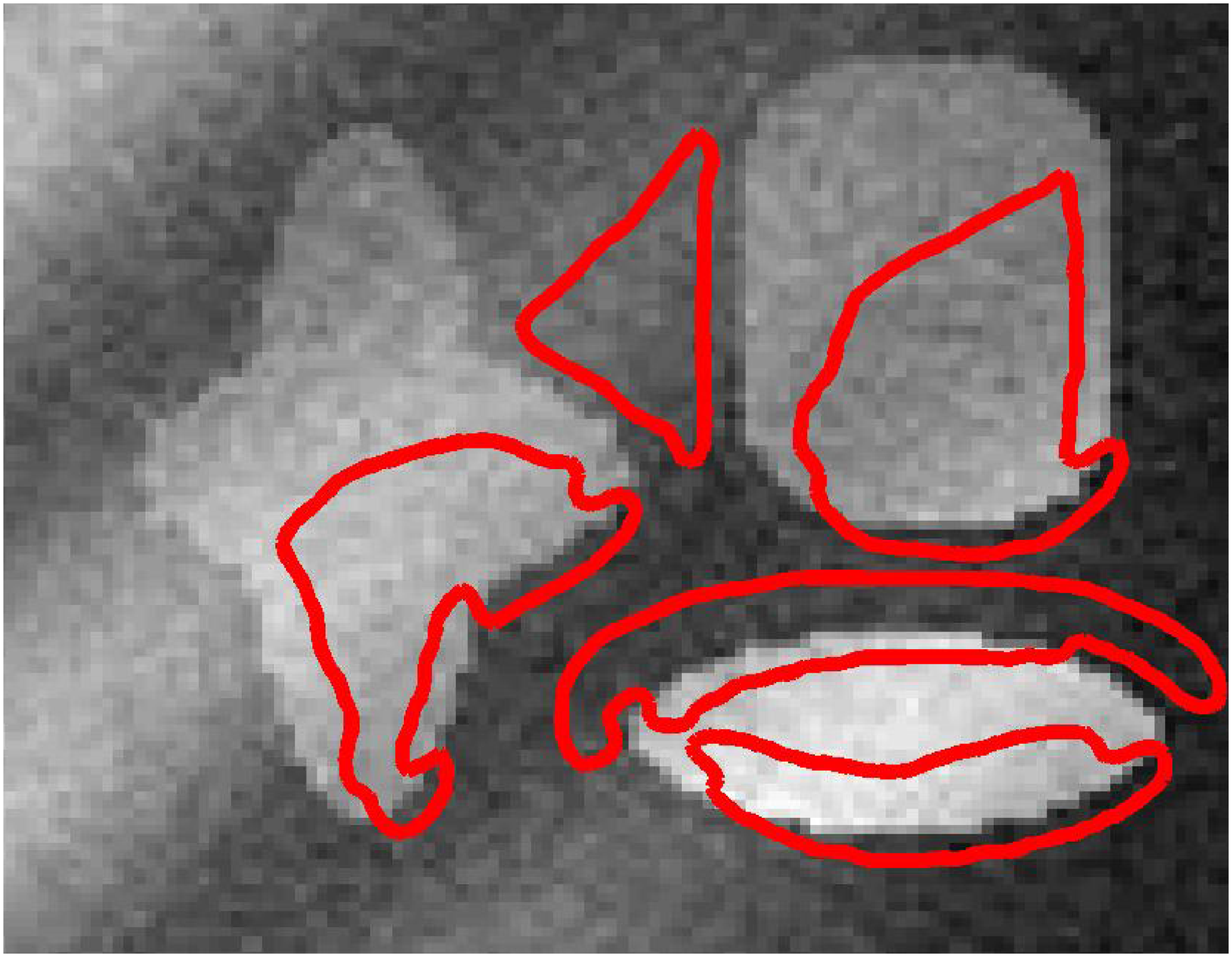} \includegraphics[width=0.15\textwidth]{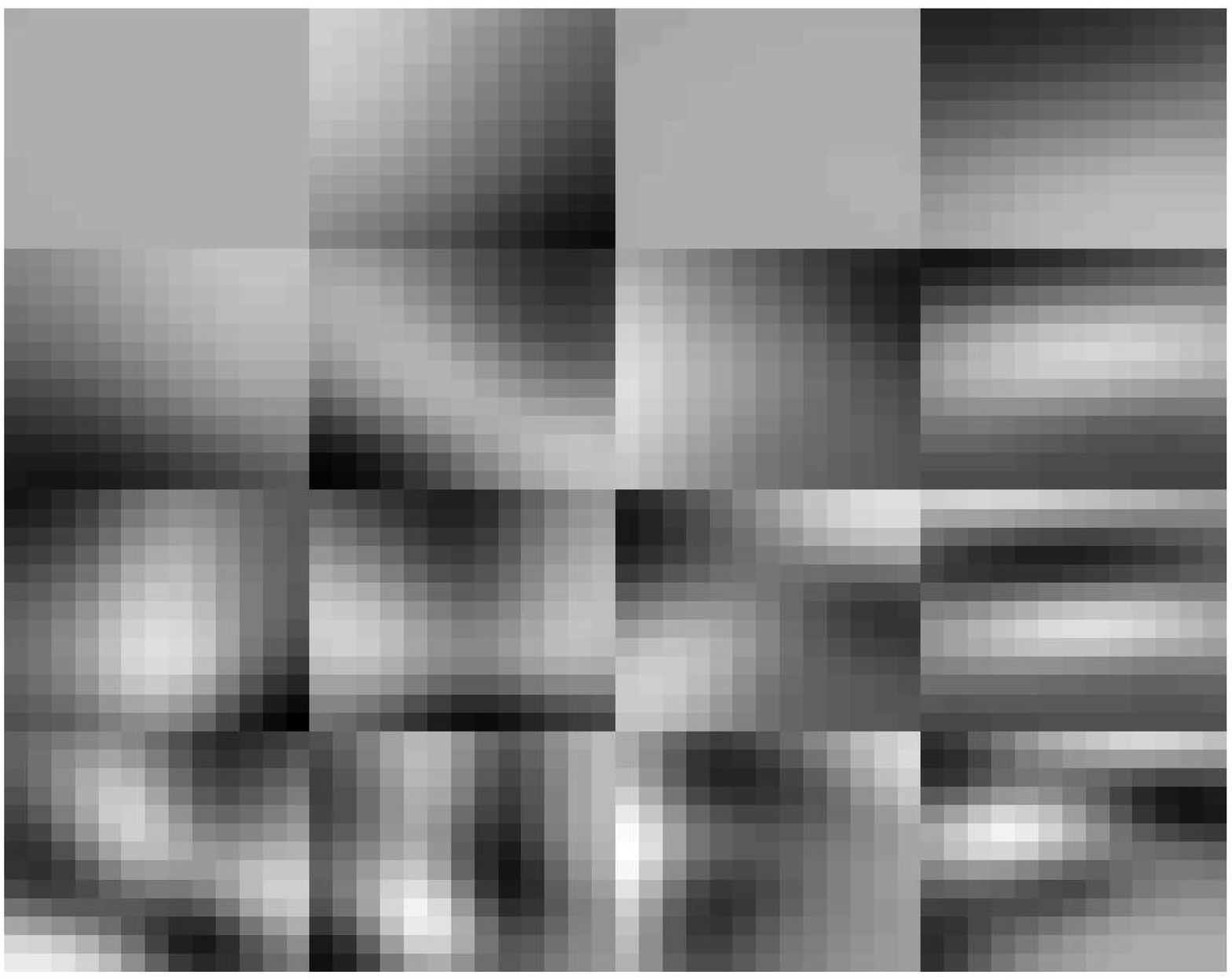}\\
  \includegraphics[width=0.15\textwidth]{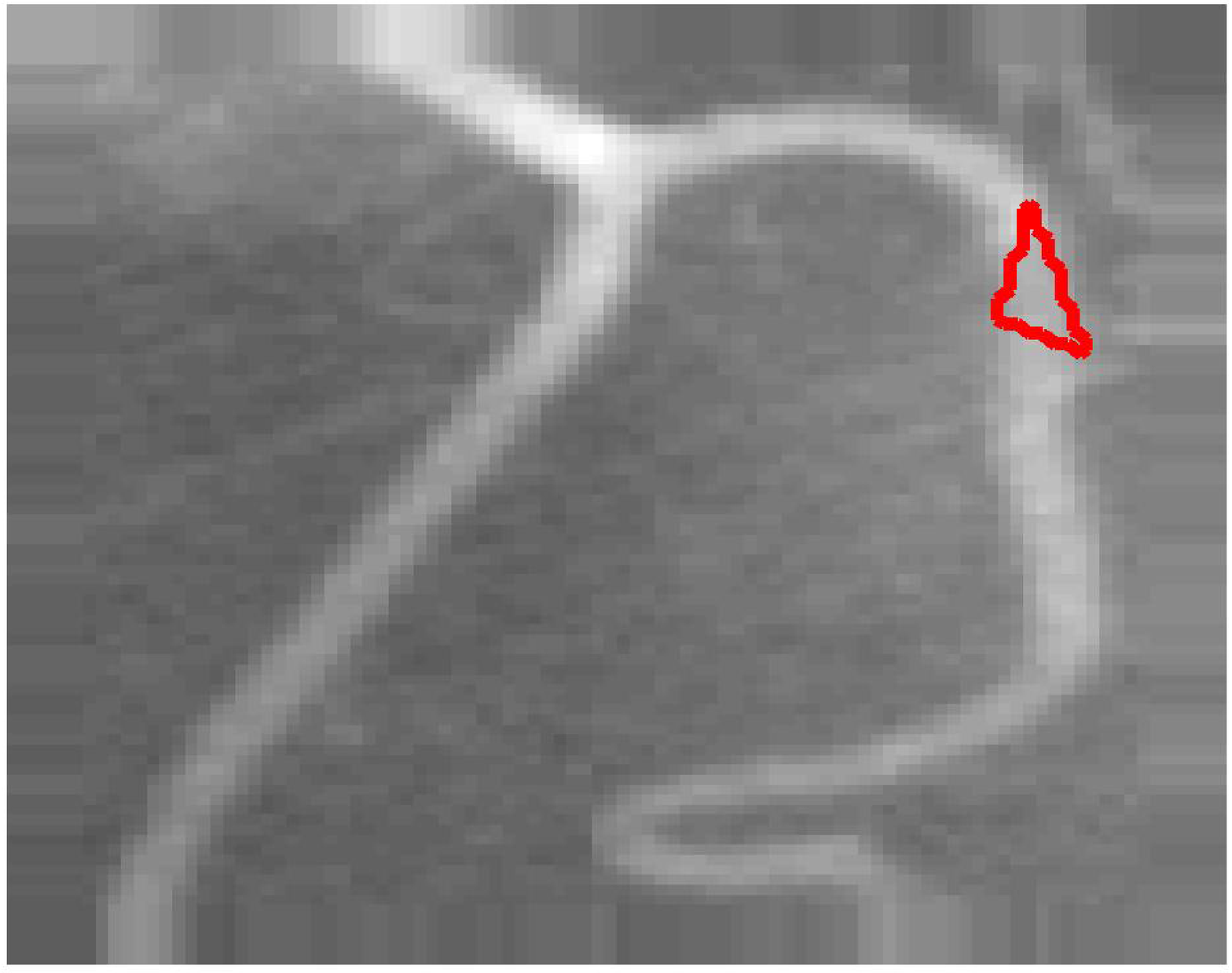} \includegraphics[width=0.15\textwidth]{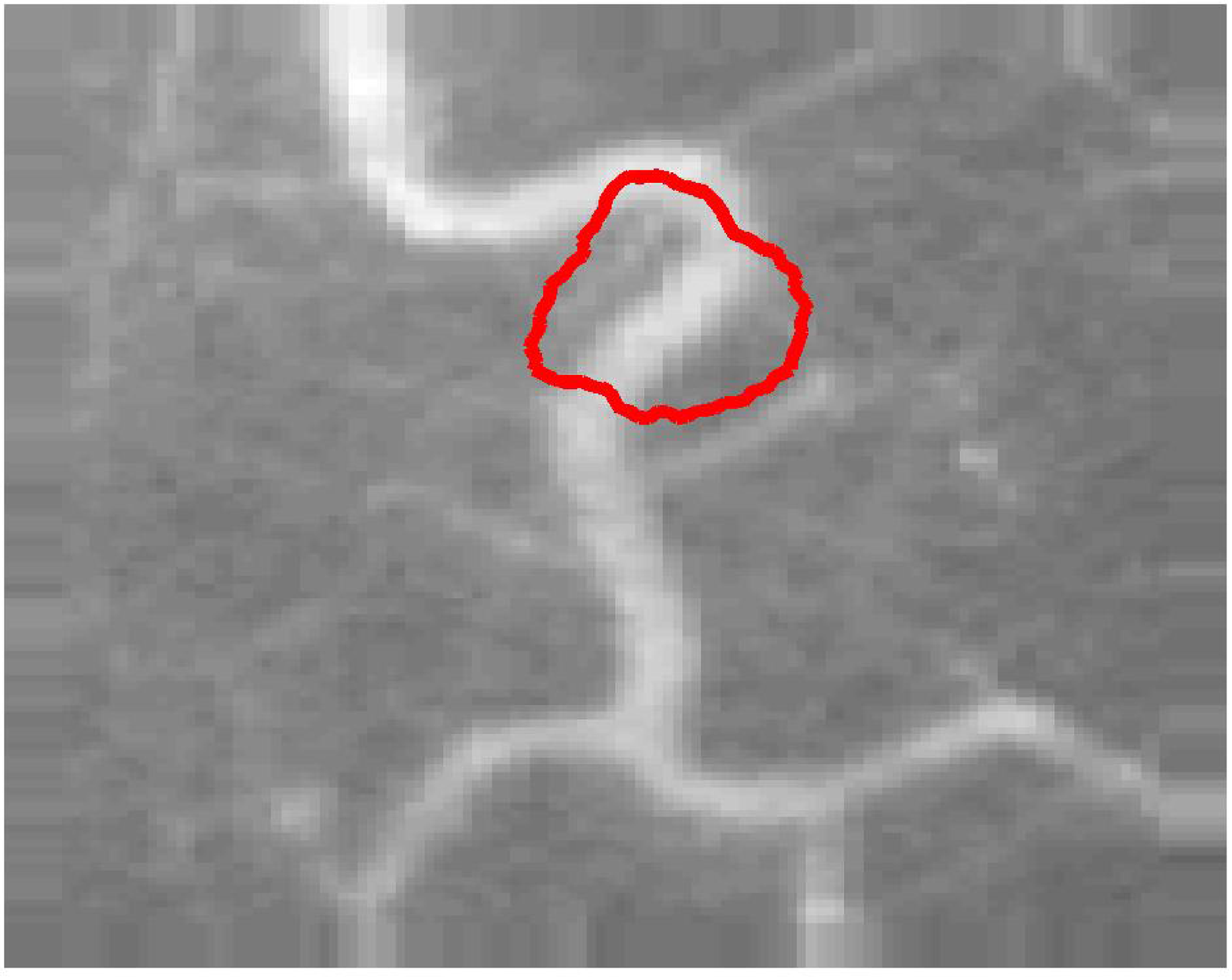}
  \includegraphics[width=0.15\textwidth]{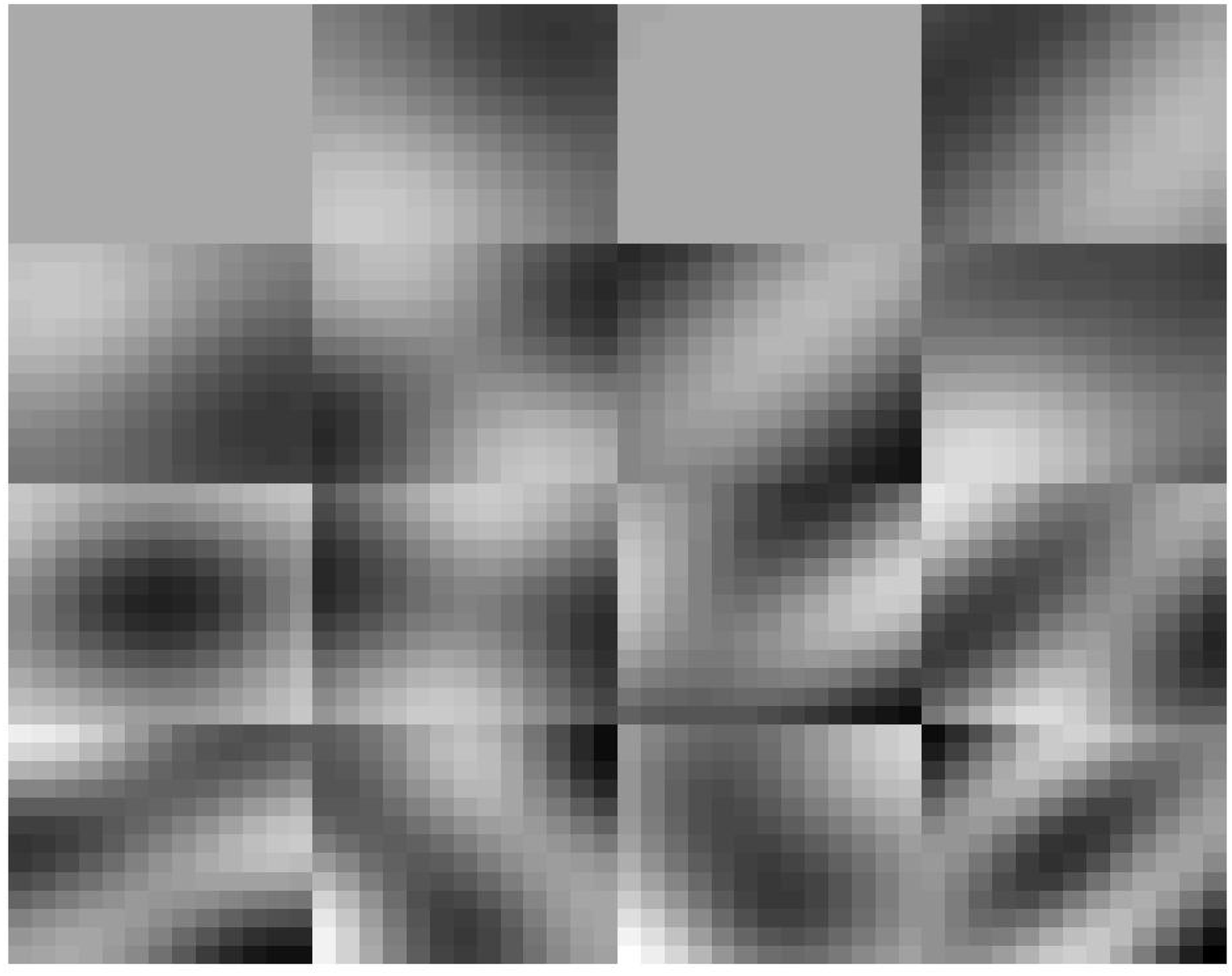}\\  \includegraphics[width=0.15\textwidth]{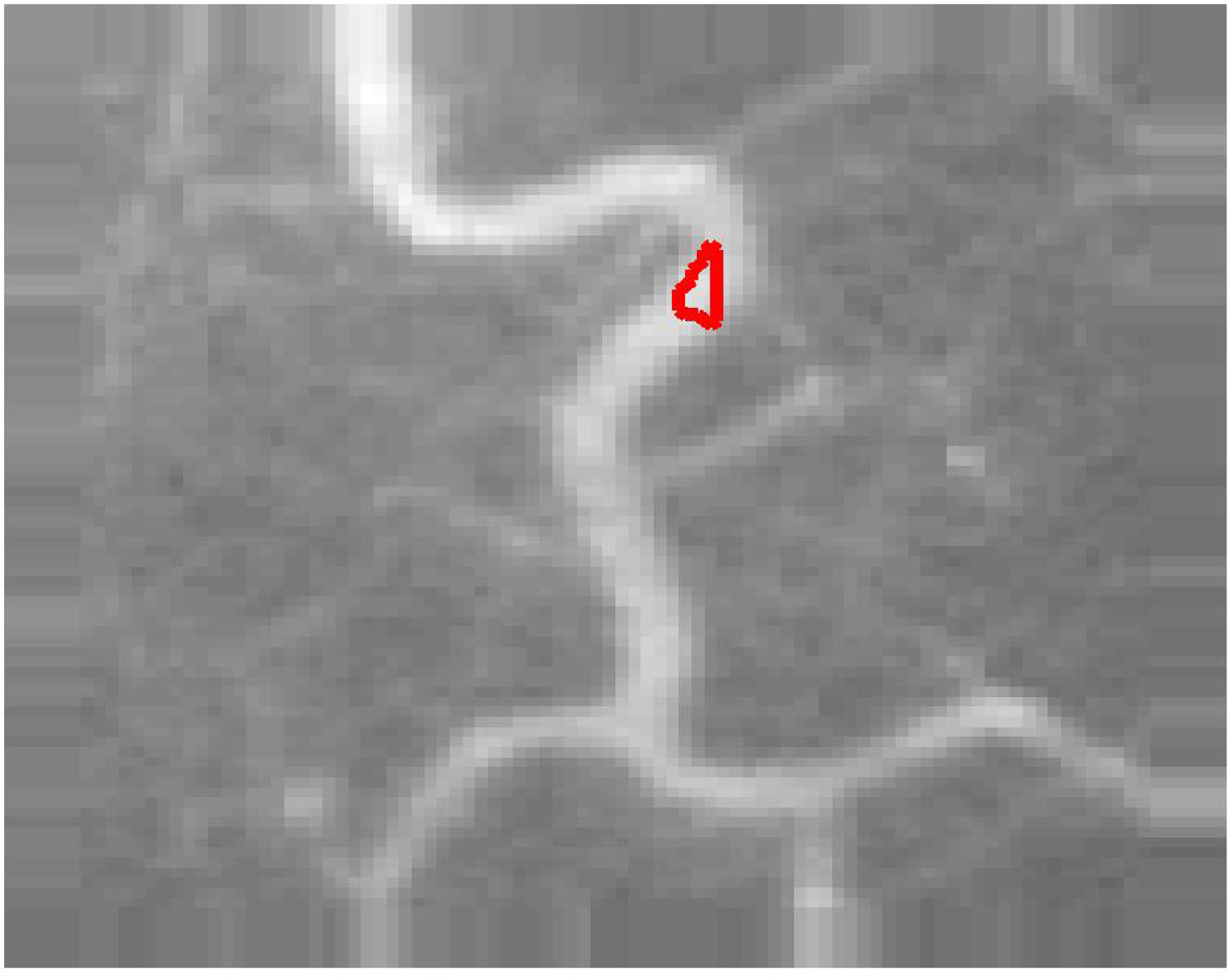} \includegraphics[width=0.15\textwidth]{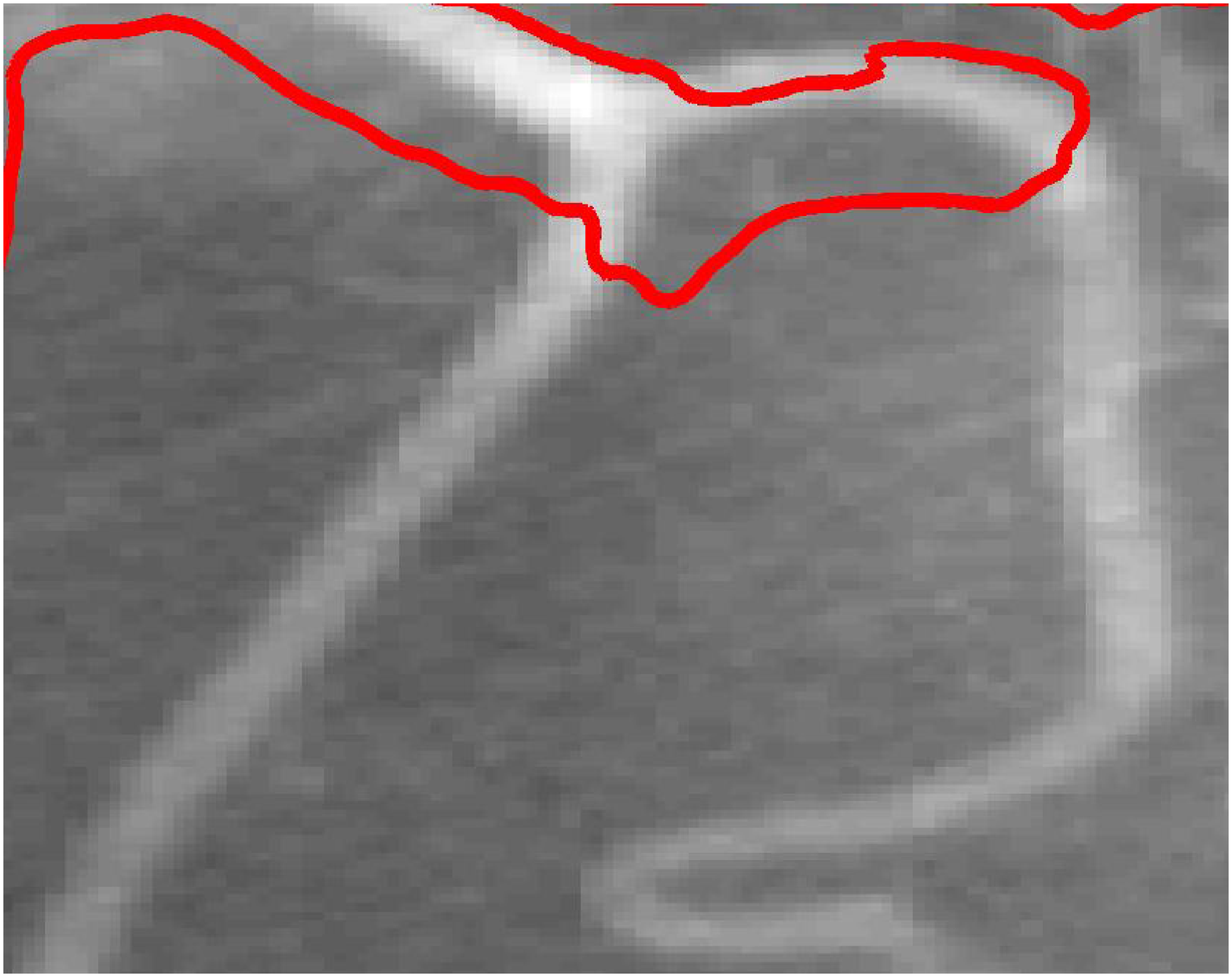}
  \includegraphics[width=0.15\textwidth]{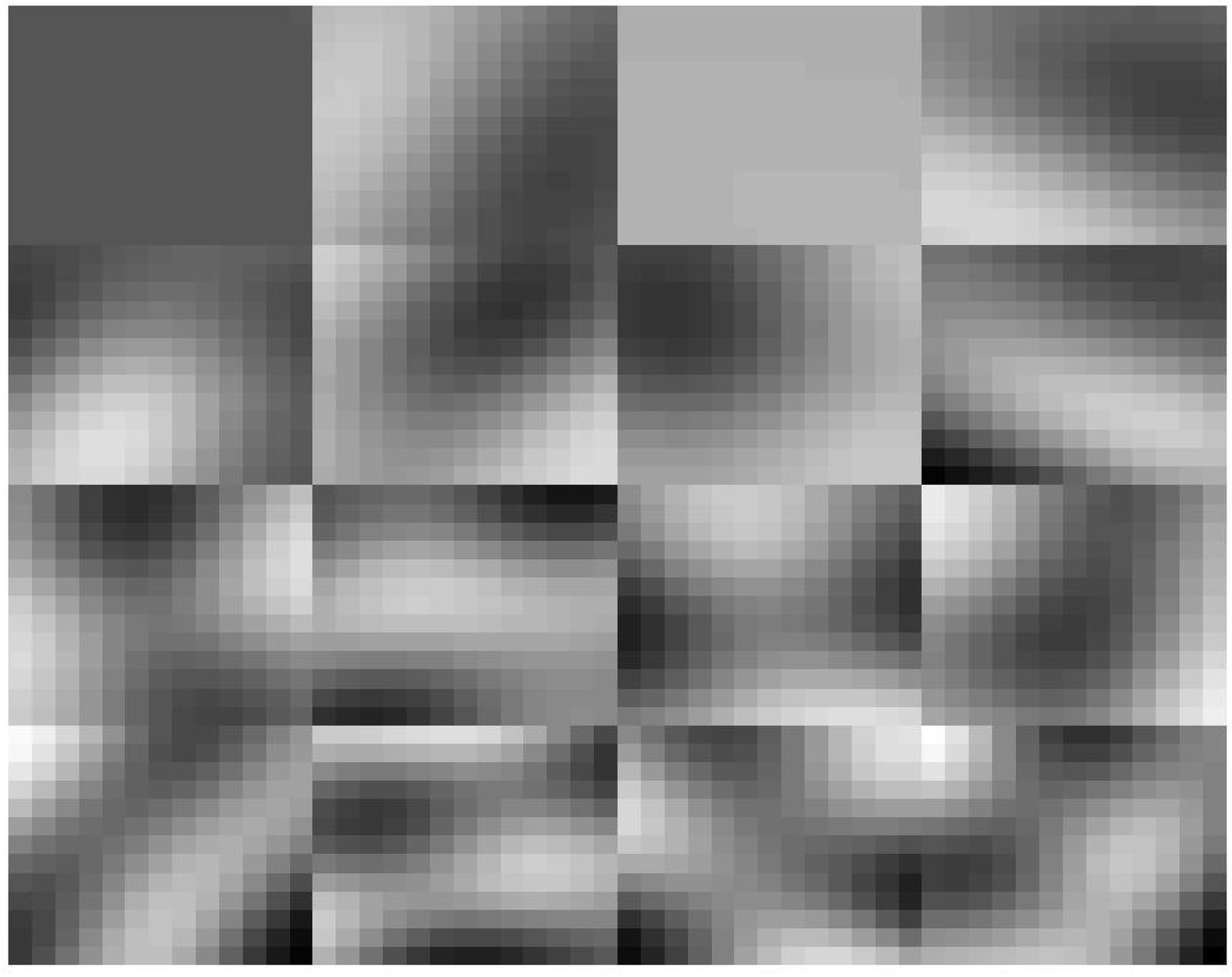}
  \caption{The segmentation of piecewise smooth images by pure linear patch reconstruction. The left column shows the initial contours. The middle column shows the converged curves. The right column shows the optimal bases. In the images of bases, the left two columns are the bases of background region, and the rest correspond to the foreground region (the regions enclosed by the contour curves).}\label{FIG:alpha_0_smooth}
\end{figure}

\begin{figure}
\centering
  \includegraphics[width=0.15\textwidth]{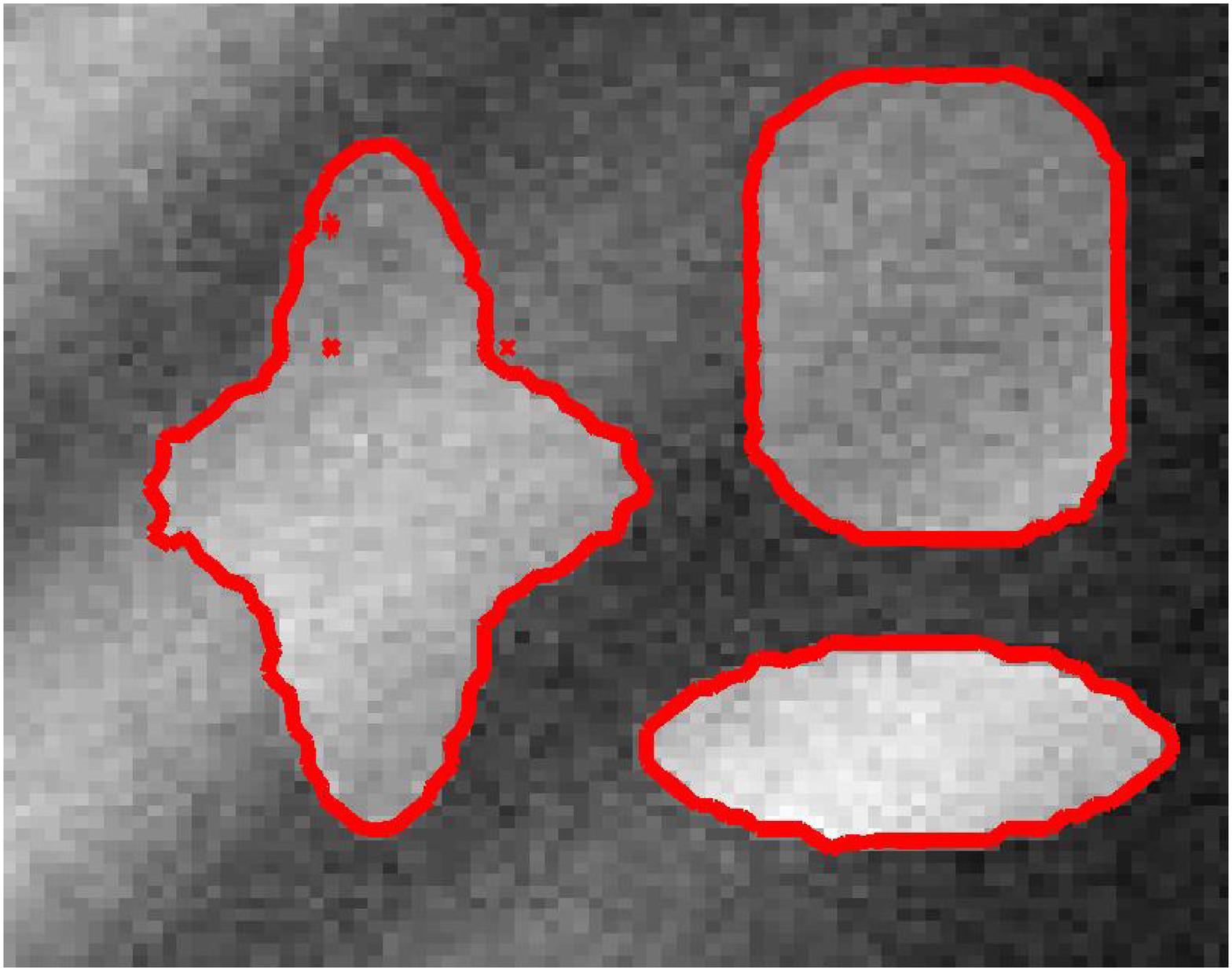}
  \includegraphics[width=0.15\textwidth]{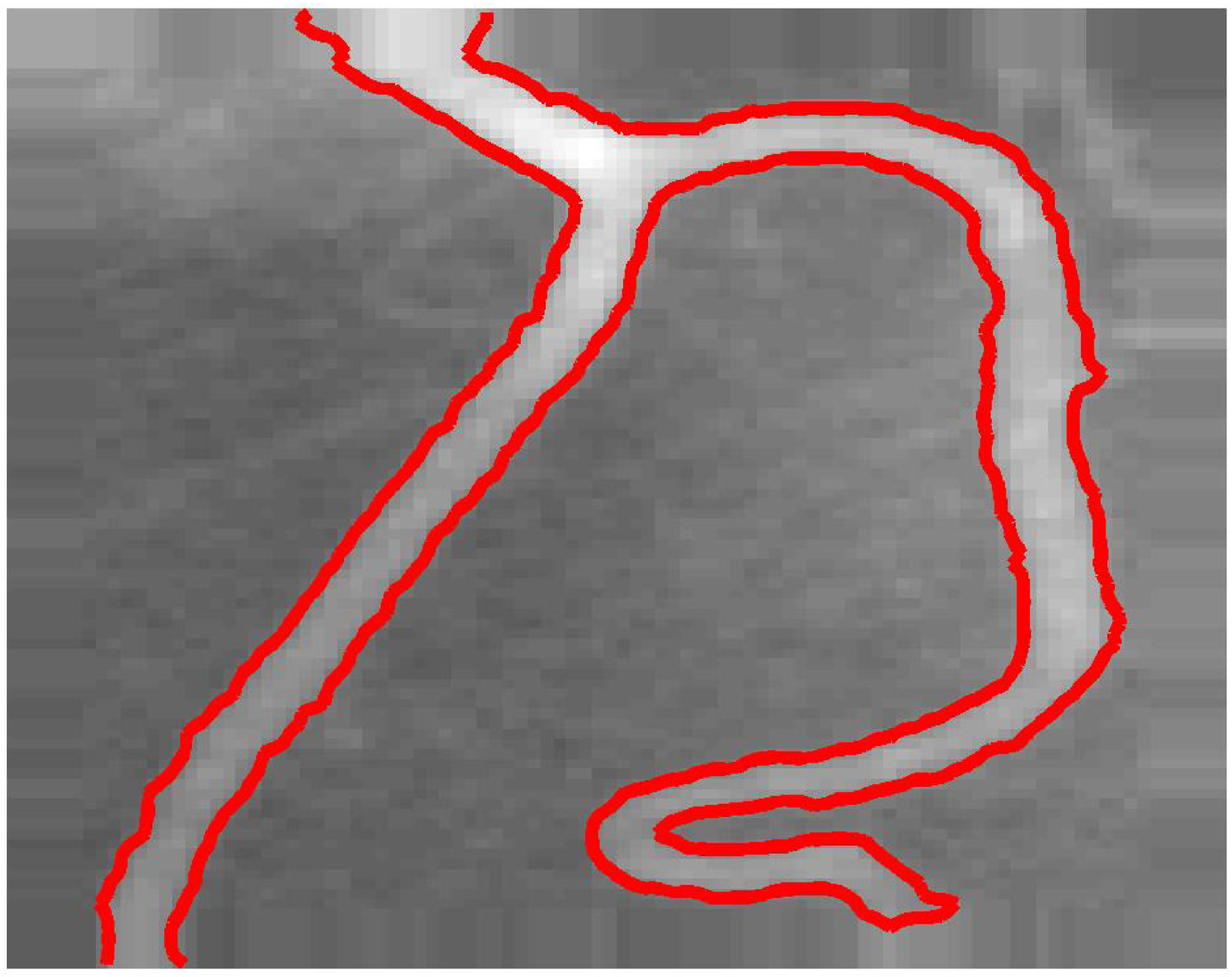}
  \includegraphics[width=0.15\textwidth]{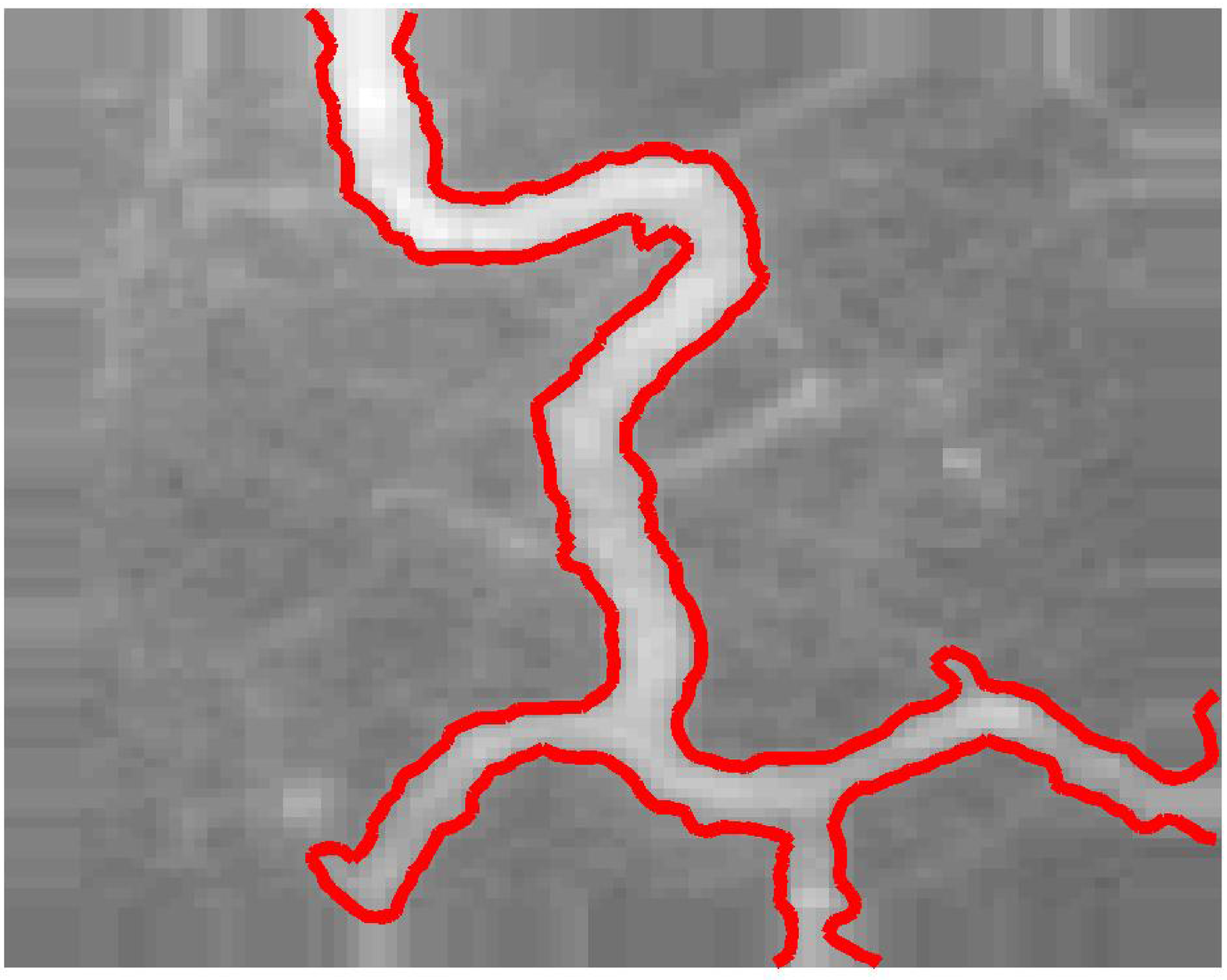}
  \caption{The results of segmenting piecewise smooth images with noise by using the coupled model. The initial curves are same as those in Fig. \ref{FIG:alpha_0_smooth}.}\label{FIG:alpha_0.1_smooth}
\end{figure}

To cope with both smooth and non-smooth contents, we may use the coupled model proposed previously. The quality of the result of the segmentation by the proposed coupled model depends on the value of the parameter $\alpha$. It is obvious that the coupled model tends to be the conventional Mumford-Shah model if $\alpha=1$, and the model tends to be the pure piecewise linear patch reconstruction if $\alpha=0$. Fig. \ref{FIG:alpha_0_smooth} shows the results of applying the pure piecewise linear patch reconstruction to piecewise smooth images (with noise). We can observe that the optimal bases of the regions are similar. Hence, the model can not deal with such images. We hope to find the $\alpha$ such that the coupled model can be used for segmenting piecewise smooth images while its ability for differentiating non-smooth structures is preserved. In our implementation, we run the segmentation with different $\alpha$ values on all the mosaic images composed of the original textures (without subtracting the mean off), obtained in the last subsection. We alter the value of $\alpha$ from $0.1$ to $0.9$ to obtain the error of segmentation shown in Fig. \ref{FIG:Alpha_Err}. We observe the steady performance for $\alpha=0$ and $0.1$, and we observe the clear decay of the performance when changing $\alpha$ from $0.1$ to $0.9$. This means that by selecting $\alpha=0.1$, the performance of the coupled model for differentiating non-smooth image structures is almost as good as that of the pure piecewise linear patch reconstruction. We also apply the model with $\alpha=0.1$ to the piecewise smooth images under noise and the good results are shown in Fig. \ref{FIG:alpha_0.1_smooth}, which indicates that the $\alpha=0.1$ is already sufficient for segmentation of piecewise smooth images. We therefore chose $\alpha=0.1$ in the coupled model to optimally balance the two terms in the coupled model.

\subsection{One-against-all segmentation}
We have so far only considered the presence of two different groups of contents in the image in our formulation, which is known as the two-phase model. Due to the capability of the level set method for handling topology changes, the two-phase model is still capable of partitioning the image into multiple smooth or non-smooth regions of two groups. However, the two-phase model cannot cope with multiple regions of multiple groups. The problem of segmentation of image into multiple different regions can be addressed via the one-against-all strategy. In other words, we may consider a problem of $n$-phase segmentation as $n$ subproblems of two-phase segmentation. In each subproblem, the image is to be partitioned into the target region and the background region which is composed of all the other regions.

\begin{figure*}
\centering
  \includegraphics[width=0.23\textwidth]{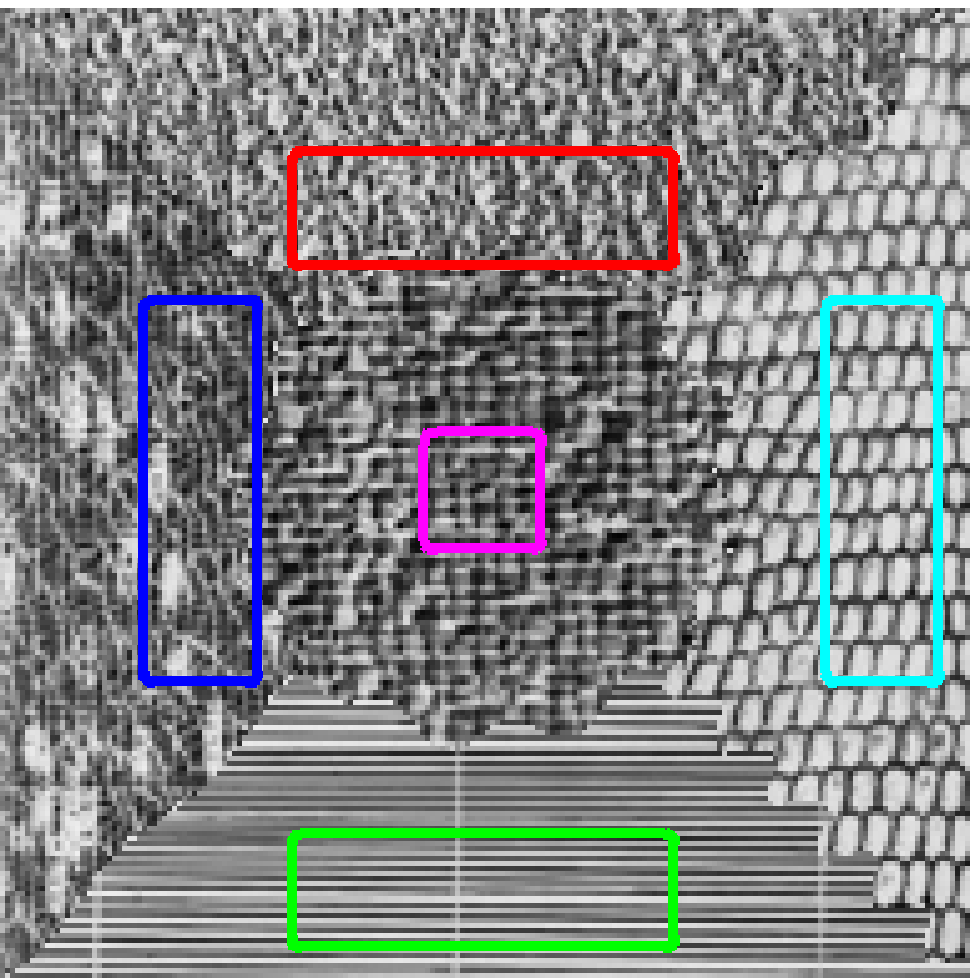}\hspace{-2pt}
  \includegraphics[width=0.23\textwidth]{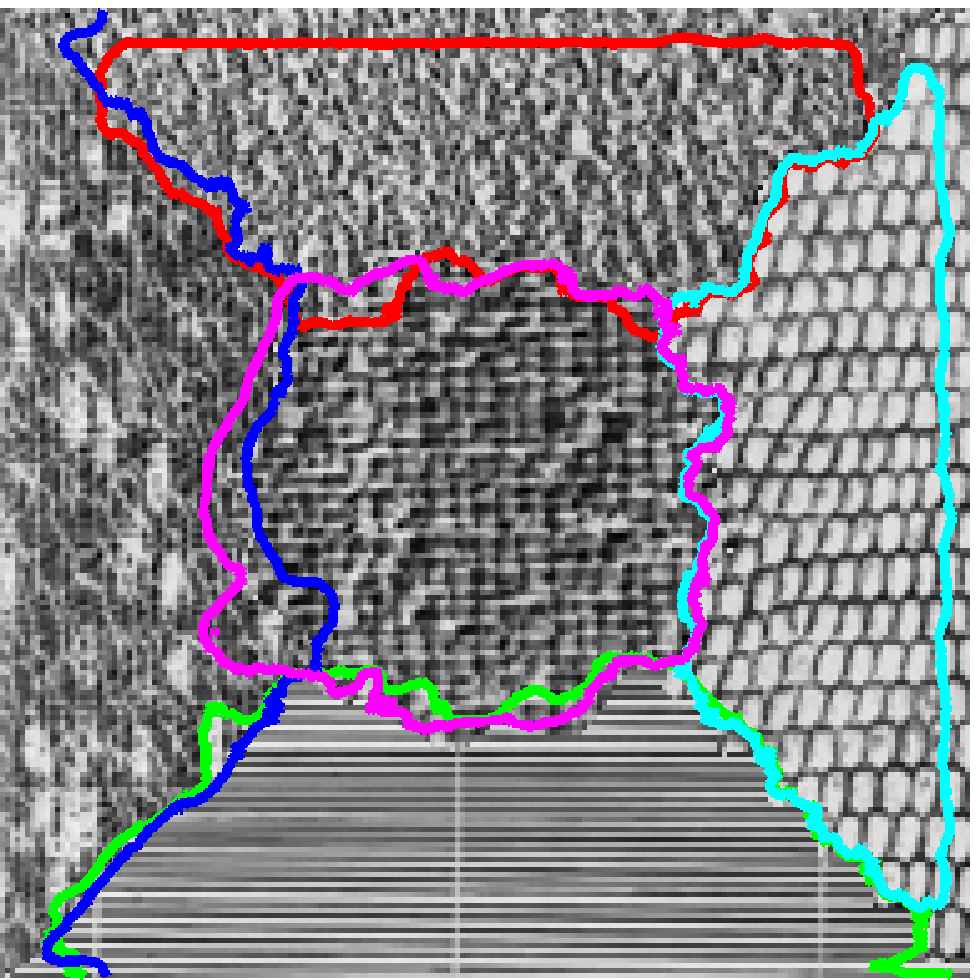}\hspace{-2pt} \includegraphics[width=0.23\textwidth]{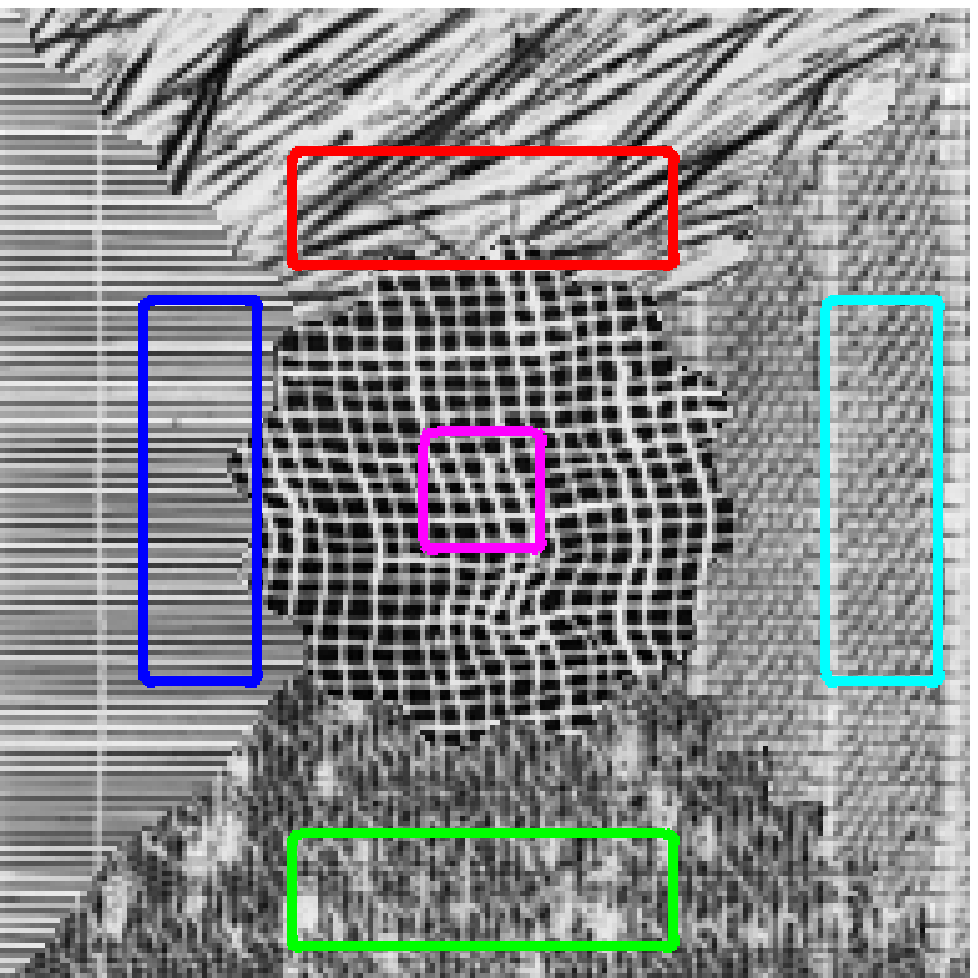}\hspace{-2pt}
  \includegraphics[width=0.23\textwidth]{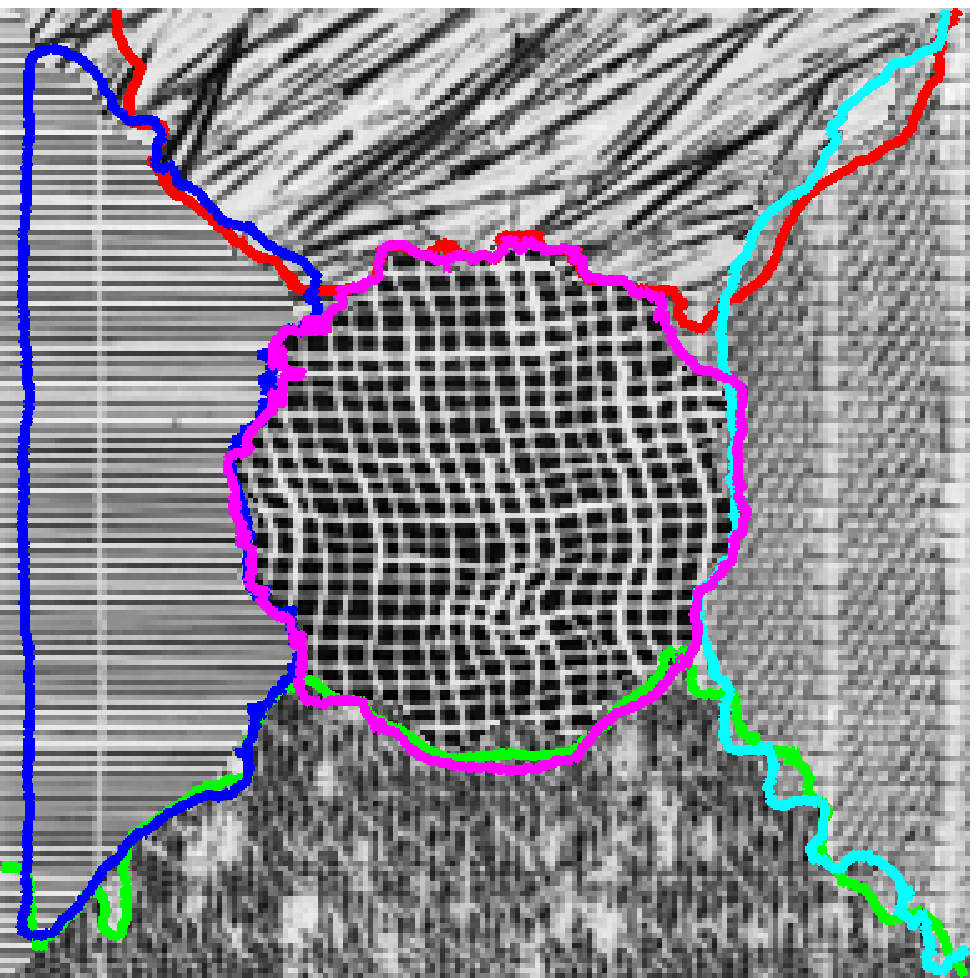}\\
  \vspace{5pt}
  \includegraphics[width=0.45\textwidth]{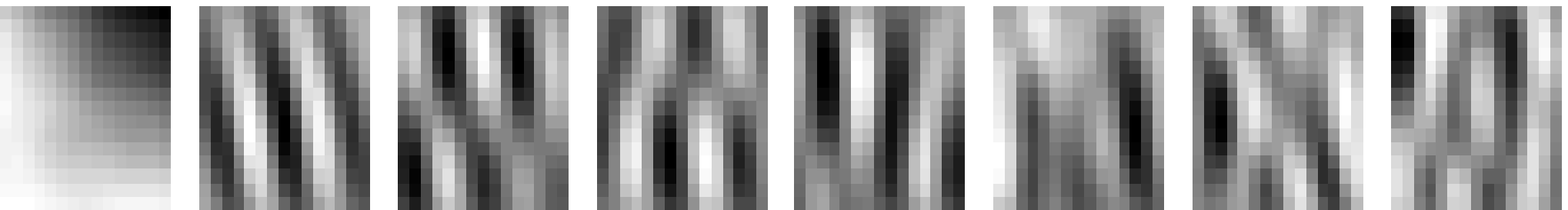}~
  \includegraphics[width=0.45\textwidth]{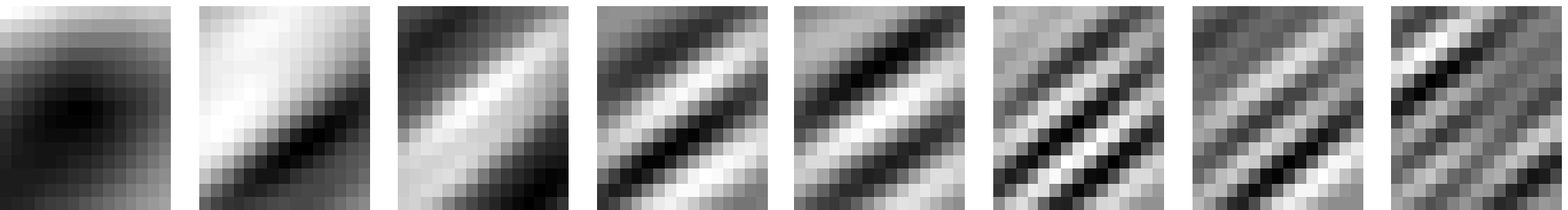}\\ \includegraphics[width=0.45\textwidth]{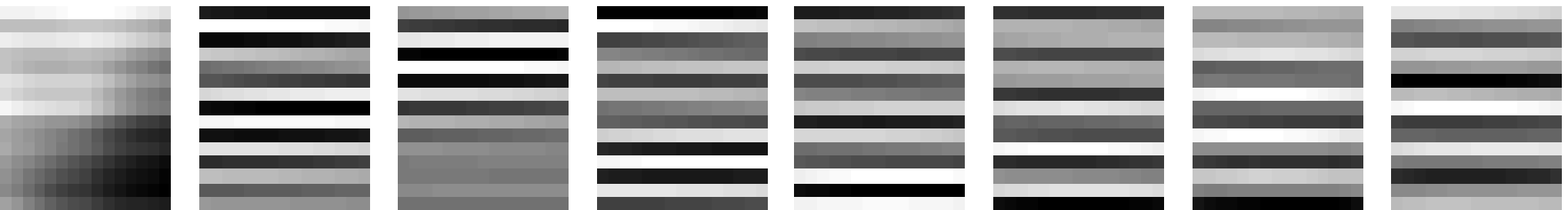}~
  \includegraphics[width=0.45\textwidth]{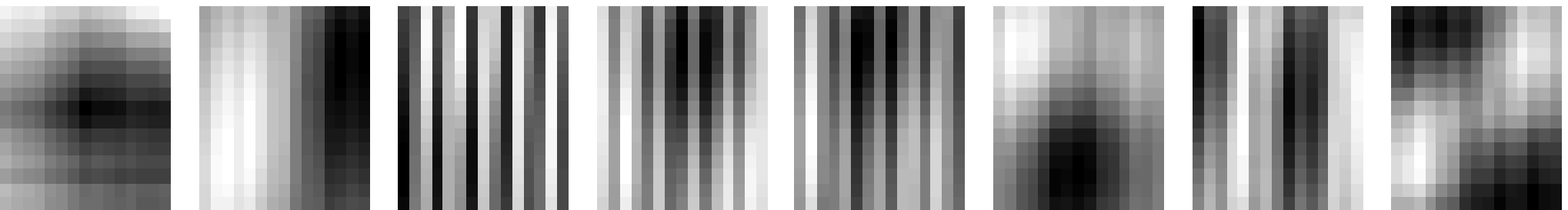}\\ \includegraphics[width=0.45\textwidth]{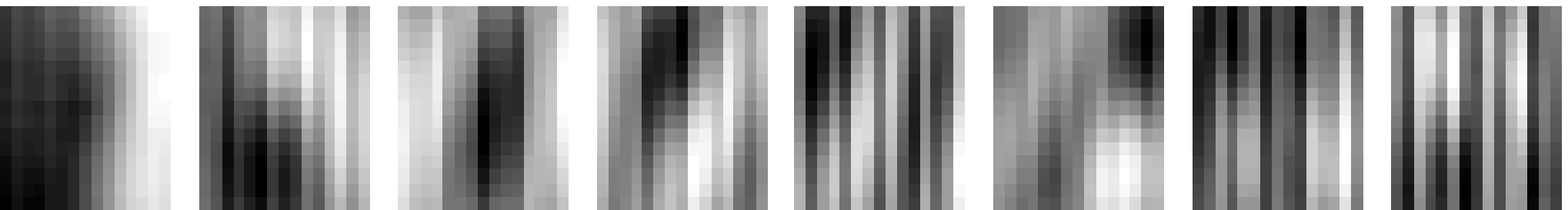}~
  \includegraphics[width=0.45\textwidth]{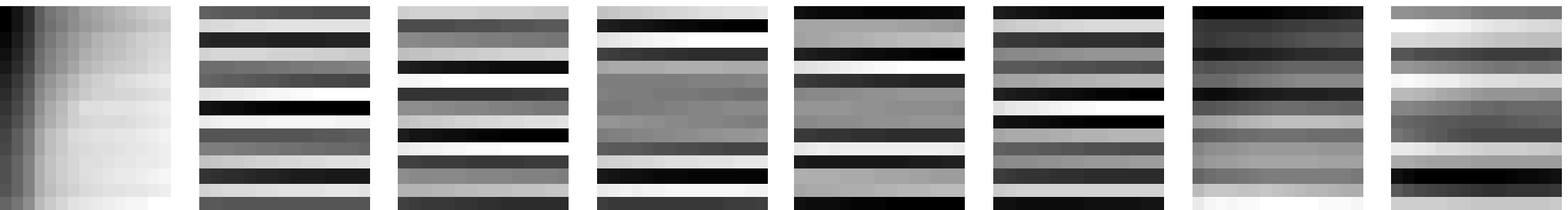}\\  \includegraphics[width=0.45\textwidth]{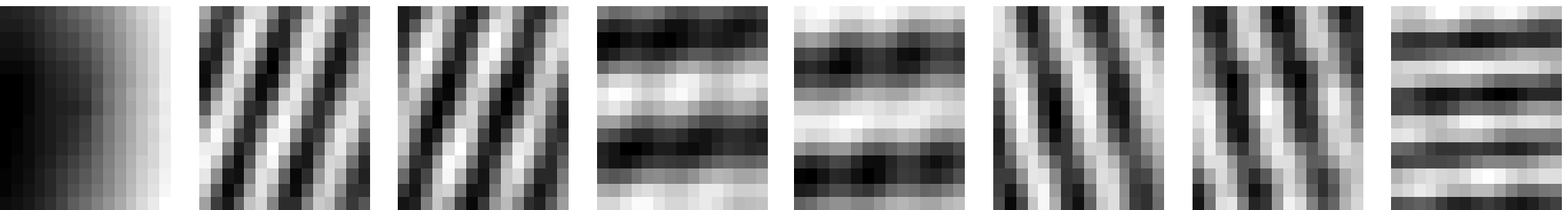}~
  \includegraphics[width=0.45\textwidth]{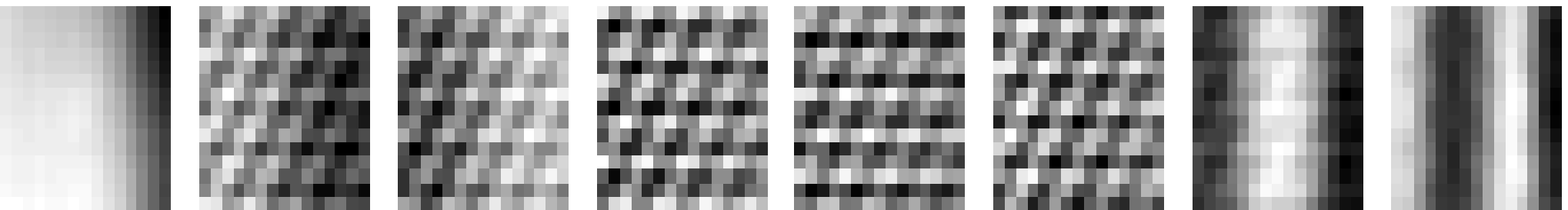}\\\vspace{-10pt} \subfloat[]{\includegraphics[width=0.45\textwidth]{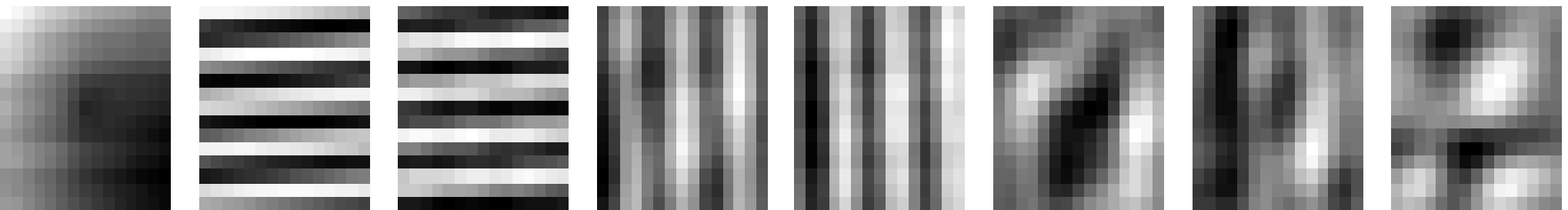}}~
  \subfloat[]{\includegraphics[width=0.45\textwidth]{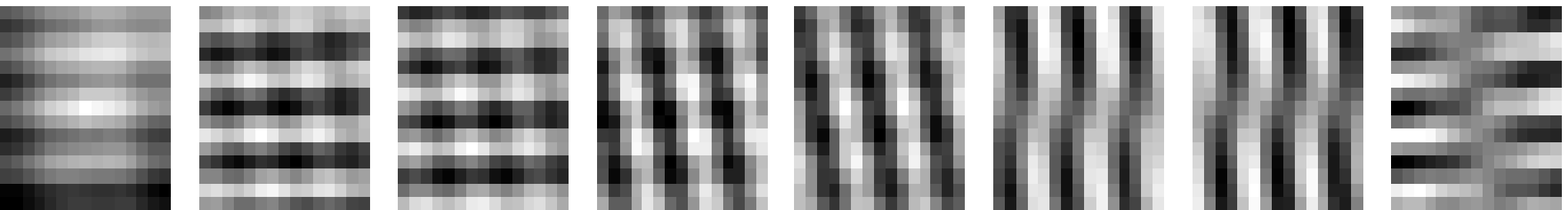}}\\
  \caption{Multi-region segmentation via one-against-all strategy (better to view in color). The top row shows the two input-output pairs of mosaic images composed of multiple texture regions. The input is the initial rectangular contour laying over the image. In each region, a unique color is assigned to the contour.
  The output is the converged curve shown in the same color as in the input. The following rows show the optimal bases corresponding to each region in the sequence of top, bottom, left, right and center.}\label{FIG:one2all_results}
\end{figure*}

To evaluate the one-against-all strategy for coping with multiple different groups of regions, we apply this strategy to the mosaic images containing five different textures. The input and output of the segmentation are shown in Fig. \ref{FIG:one2all_results}. The initial rectangular contours are shown in a unique color assigned to a region in one image. The converged contour curves are shown in the same colors in the other image. The optimal bases corresponding to different regions are also visualized. We can observe that these bases well capture the principal structure of the corresponding regions. The error rates of the segmentation are summarized in Table \ref{TB:one2all_SEG_ERR}.

\begin{table}
\centering
\caption{Error rates (\%) of the segmentation shown in Fig. \ref{FIG:one2all_results} }\label{TB:one2all_SEG_ERR}
  \begin{tabular*}{\columnwidth}{@{\extracolsep{\fill}}c|cccccc}
    \toprule
    & Top  & Bottom  & Left  & Right  & Center  & Total\\ \hline
Fig. \ref{FIG:one2all_results}(a) & 5.16 & 2.19 & 1.94 & 4.3 & 3.2 & 16.8\\
Fig. \ref{FIG:one2all_results}(b) & 2.86 & 2.11 & 3.94 & 2.03 & 1.92 & 12.88 \\
    \bottomrule
  \end{tabular*}
\end{table}

\section{Discussions and conclusion}\label{SEC:Con}
\subsection{Discussions on the limitations}
The size of the patches and the number of bases are predetermined, which is empirical. Both of the two properties affect the segmentation. For example, when increasing the patch size, the number of bases needed for small reconstruction error would increase due to the enlarged dimensionality of the patch. Therefore, to achieve a small error of segmentation the number of bases needed has to be large. Larger patch size and more bases would also provide a richer description of the non-smooth structure, which may help segmentation in complex situations. With a small patch size, which means the reconstruction error can be small with few bases, the discrimination might not be clear, since the same set of bases may give similar reconstruction errors to different group of patches in such case.

This work did not really target at natural image segmentation. Specifically, this work explores the mathematical model for addressing an important aspect of the natural image segmentation, i.e. that of coping with non-smooth structures in the segmentation of images which are 2D signals. For natural image segmentation, more sophisticated framework has to be adopted to mimic human vision.

\subsection{Conclusion}
We propose a unified energy minimization model of non-smooth image segmentation without requiring any separate process of feature extraction. The model is the energy minimization of the error of piecewise linear patch reconstruction. The segmentation error rate of the proposed model is proven to be bounded by the patch reconstruction error. The gradient descent method for solving the linear patch reconstruction is proven to be globally optimal under mild conditions on the initialization. The experiments validate our theoretical claim and show the clear supreme performance of our methods over other relevant methods. The linear patch reconstruction in our approach can be viewed as an unsupervised dictionary learning process. Segmentation with features depends largely on the prior knowledge, whilst our approach is mostly data-driven. Our approach is useful when prior knowledge is inapplicable.

\renewcommand{\theequation}{A-\arabic{equation}}
\setcounter{equation}{0}  
\renewcommand*\thesection{\Alph{section}}

\section*{Appendix}


\begin{proof}[Proof of Proposition \ref{PROP:Opt_H}]

Since $H_l\in\{0,1\}$ and $\sum_{l=1}^2 H_l = 1$, we have  $\sum_{l=1}^2 \mathcal{E}_{l}(x,y) H_l$ $= \big\{\mathcal{E}_1(x,y),\mathcal{E}_2(x,y)\big\}$. Thus, for any position $[x,y]^T\in\Omega$ we can choose $\mathcal{E}_{l'}=\min\{\mathcal{E}_1(x,y),\mathcal{E}_2(x,y)\}$, which is equivalent to $\{H_l\}^*= \argmin\limits_{\{H_l\}} \left\{\sum\limits_{l=1}^2 \mathcal{E}_{l}(x,y) H_l\right\}$ at $[x,y]^T$. As a result, $\sum\limits_{l=1}^2 \mathcal{E}_{l}(x,y) H_l^*\leq\sum\limits_{l=1}^2 \mathcal{E}_{l}(x,y) H_l$ for any $H_l$. Thus, $\int_\Omega\sum_{l=1}^2 \mathcal{E}_l(x,y) H_l^* dxdy\leq\int_\Omega \sum_{l=1}^2 \mathcal{E}_l(x,y) H_ldxdy$, which completes our proof.
\end{proof}
Note that the proof relies on the Axiom of Choice for the continuous domain. The choosing is feasible if we approximate the integral by discretization.

\begin{proof}[Proof of Proposition \ref{Prop:Seg_Patch_err}]
To prove this proposition, we require the following lemma.
\begin{lemma}\label{PROP:SEG_err_1}
Given the same condition in Proposition \ref{Prop:Seg_Patch_err} then the following holds.
\begin{equation}\label{BD:Prob_SEG_ERR1}
\begin{split}
\varepsilon_{seg}&\leq {\int_\Omega \mathcal{E}_1(x,y) H dxdy\over|\Omega_1|\mathcal{E}_2(x',y')}\\
&+{\int_\Omega \mathcal{E}_2(x,y) (1-H) dxdy\over|\Omega_2|\mathcal{E}_1(x',y')}
\end{split}
\end{equation}
\end{lemma}
The proof of this lemma is due to Markov's inequality. This bound connects the error of segmentation to the error of reconstruction.

Additionally, we require the following fundamental model of functions in the literature of signal analysis, such as wavelets \cite{Meyer92wavelets} and especially compressive sensing \cite{Donoho06CS} \cite{CandesTao06CS}.
\begin{definition}[Universal energy bound]
Given a (discrete) function $f\in\mathbb{R}^N$, and any subset of fixed system of sorted orthogonal bases $\{{v}_i|i=1,2,...,K\leq N\}$ such that $\|\langle {f},{v}_1\rangle\|_{l_2}\geq\|\langle {f},{v}_2\rangle\|_{l_2}\geq...\geq\|\langle {f},{v}_K\rangle\|_{l_2}$, then the following bound holds for every $0<n\leq N$,
\begin{equation}\label{BD:Power_Decay}
\|\langle f, {v}_n\rangle\|_{l_2} \leq R n^{-q},
\end{equation}
where $R = \|\langle f, {v}_1\rangle\|_{l_2}$, and $0\leq q<\infty$.
\end{definition}
The worst case is when $\|\langle {f},{v}_1\rangle\|_{l_2}=\|\langle {f},{v}_2\rangle\|_{l_2}=...=\|\langle {f},{v}_K\rangle\|_{l_2}$, i.e. $q=0$.

From Definition of universal energy bound, the linear patch reconstruction error at every pixel could be bounded by using (\ref{BD:Power_Decay}) as follows.
\begin{equation}\label{BD:patch_error}
\begin{split}
\|\square\|\mathcal{E}_l^{\square} &=  \big\|\mathbf{p} - \sum_{k=1}^{K}\big\langle\mathbf{p},\mathbf{v}_k\big\rangle_{\square}\mathbf{v}_k\big\|^2_{\square}\\
&=\|\mathbf{p}\|^2_{\square}-\sum_{k=1}^{K}\big\langle\mathbf{p},\mathbf{v}_k\big\rangle_{\square}^2\\
&\geq\|\mathbf{p}\|^2_{\square}-\sum_{k=1}^{K}R_l^2 k^{-2q_l}\\
&\geq\|\mathbf{p}\|^2_{\square}-R_l^2 K^{2-2q_l}
\end{split}
\end{equation}
where $l=1,2$. Substituting (\ref{BD:patch_error}) into the denominator of (\ref{BD:Prob_SEG_ERR1}), and if the denominator is positive, we complete our proof.
\end{proof}

\begin{theorem}\label{THM:NonCONVEX}
The optimization problem defined in Eq. (\ref{EQ:Rec_Error_K}) is nonconvex.
\end{theorem}
\begin{proof}[Proof of Theorem \ref{THM:NonCONVEX}]
First, we shall prove that the objective functional is nonconvex. Taking functional derivatives of the energy twice, we have the following.
\begin{equation}
\begin{split}
&{D^2\over D{\mathbf{v}_i^l}^2}E_p\Big(\{\mathbf{v}_i\}\Big)\\
&=-\int\limits_\Omega  I(x+u', y+v')I(x+u, y+v) H_l dxdy\\
&=-\int\limits_\Omega  I(x+u', y+v')H_l I(x+u, y+v)H_l  dxdy\\
&=-\int\limits_\Omega  I_{Hl}(x+u',y+v')I_{Hl}(x+u,y+v)  dxdy\\
&=-\Lambda_{Hl}(u',v',u,v)
\end{split}
\end{equation}
If this functional is convex, $-\Lambda_{Hl}(u',v',u,v)$ will be positive semi-definite. However, we can show that the $-\Lambda_{Hl}(u',v',u,v)$ is actually negative semi-definite as follows.

\begin{equation}
\begin{split}
&-\int\limits_\Omega h(u',v') \Lambda_{Hl}(u',v',u,v)h(u,v) dudv du'dv'\\
&=-\int\limits_\Omega \left|\int\limits_{\square} I(x+u, y+v)h(u,v)dudv\right|^2H_l dxdy\\
&\leq0
\end{split}
\end{equation}
Hence, the objective functional is concave. Further, we consider the constraints in terms of orthonormality. Let $\{\mathbf{x}_i\}$ and $\{\mathbf{y}_i\}$ both satisfy the constraints. We wonder whether the convex combination also satisfy the constraints. For example, let $\mathbf{w}_1 = \alpha\mathbf{x}_1+(1-\alpha)\mathbf{y}_1$, where $\alpha>0$. Then we can verify that $\|\mathbf{w}_1\|\neq 1$, unless $\mathbf{x}_1=\mathbf{y}_1$, which is generally not true. Therefore, the constraints are also nonconvex. Either of the above two facts can complete our proof.\end{proof}

\begin{theorem}\label{THM:MANYSOL}
In the optimization problem defined in Eq. (\ref{EQ:Rec_Error_K}),
if $\{\mathbf{v}_k,k=1,2,...,K\}$ is the global optimal solution, then the transformation of $\{\mathbf{v}_k,k=1,2,...,K\}$ via an orthogonal matrix $\mathcal{R}^{K\times K}$ is also the global optimal solution.
\end{theorem}
\begin{proof}[Proof of Theorem \ref{THM:MANYSOL}]
First, we define a linearly transformed patch basis $\{\mathbf{w}_k\}$ as follows.
\begin{equation}
\begin{split}
\mathbf{w}_1(u,v) &= \big[\mathbf{v}_1^l(u,v),\mathbf{v}_2^l(u,v),...,\mathbf{v}_{K}^l(u,v)\big]^T\mathbf{a}_1\\
\mathbf{w}_2(u,v) &= \big[\mathbf{v}_1^l(u,v),\mathbf{v}_2^l(u,v),...,\mathbf{v}_{K}^l(u,v)\big]^T\mathbf{a}_2\\
&\vdots\\
\mathbf{w}_{K}(u,v) &= \big[\mathbf{v}_1^l(u,v),\mathbf{v}_2^l(u,v),...,\mathbf{v}_{K}^l(u,v)\big]^T\mathbf{a}_{K}
\end{split}
\end{equation}
where $\mathbf{A}=[\mathbf{a}_1,\mathbf{a}_2,...,\mathbf{a}_{K}]^T$ is an $K\times K$ mixing matrix. We may simply write the following.
\begin{equation}
\begin{split}
\mathbf{w}_1 = \mathbf{V}\mathbf{a}_1,
\mathbf{w}_2 = \mathbf{V}\mathbf{a}_2,\dots,\mathbf{w}_{K} = \mathbf{V}\mathbf{a}_K
\end{split}
\end{equation}
where $\mathbf{V}=\big[\mathbf{v}_1^l(u,v),\mathbf{v}_2^l(u,v),...,\mathbf{v}_{K}^l(u,v)\big]^T$.

Hence, the energy in terms of $\mathbf{w}_k(u,v)$ could be rewritten as follows.
\begin{equation}
\begin{split}
&U\Big(\{\mathbf{w}_k\}\Big)= \int_\Omega \sum\limits_{k=1}^{K} \Big\langle \mathbf{p},\mathbf{w}_k\Big\rangle_{\square}^2 H_l dxdy\\
&=\int_\Omega \sum\limits_{k=1}^{K} \Big\langle \mathbf{p},\mathbf{V}\mathbf{a}_k\Big\rangle^2_{\square} H_l dxdy\\
&=\int_\Omega \left\| \mathbf{A} \mathbf{V}^T\mathbf{p}\right\|^2 H_l dxdy\\
&=\int_\Omega \left(\mathbf{p}^T\mathbf{V}\big[\mathbf{A}^T\mathbf{A}\big]\mathbf{V}^T\mathbf{p}\right) H_l dxdy
\end{split}
\end{equation}
If the mixing matrix $\mathbf{A}$ is orthogonal, we obtain the following.
\begin{equation}
\begin{split}
&U\Big(\{\mathbf{w}_i\}\Big) = \int_\Omega \left(\mathbf{p}^T\mathbf{V}\mathbf{V}^T\mathbf{p}\right) H_l dxdy= U\Big(\{\mathbf{v}_i^l\}\Big)
\end{split}
\end{equation}
Therefore, if $\{\mathbf{v}_i^l\}$ is the global optimal solution, the $\{\mathbf{w}_i\}$ is also the global optimal solution, which completes the proof.
\end{proof}

To prove Theorem \ref{THM:EigenProc} we require some lemmas.
\begin{lemma}\label{LM:SND}
The integral operator $\big[\Lambda_{Hl}\big]$ is symmetric positive semi-definite.
\end{lemma}
The symmetry is straightforward. The proof of positive semi-definiteness is as in the proof of Theorem \ref{THM:NonCONVEX}.

%
Hence according to Mercer's theorem of eigen-decomposition of symmetric nonnegative definite bounded integral operator, we can write $\Lambda_{Hl}(u,v,u',v')$ in the form of infinite series of eigenfunctions as follows.

\begin{theorem}[Mercer's representation]
Assuming $\|[\Lambda_{Hl}(u,v,u',v')]\|<\infty$, then
\begin{equation}
\Lambda_{Hl}(u,v,u',v') = \sum_{i=1}^{\infty} \lambda_i \mathbf{e}_i(u,v)\mathbf{e}_i(u',v')
\end{equation}
where $\lambda_1\geq\lambda_2\geq\ldots>0$ are the eigenvalues, $\{\mathbf{e}_i(u,v),i=1,2,...\}$ is the set of eigenfunctions, and the eigenfunctions form a system of orthogonal basis.
\end{theorem}
A proof of the above may be found in \cite{AshBook_InformationTheory}. Now we are in a position to prove the Theorem \ref{THM:EigenProc}.
\begin{proof}[Proof of Theorem \ref{THM:EigenProc}:]
First, we write the gradient descent differential equation for the \textsc{STEP} 1 as follows.
\begin{equation}
\begin{split}
&{\partial \mathbf{v}_1^l(u,v,t)\over\partial t}\\ &=\int \Lambda_{Hl}(u,v,u',v')\mathbf{v}_1^l(u',v',t)du'dv'
\end{split}
\end{equation}
where $\mathbf{v}_1^l(u,v,0) = v_0$. The corresponding update equation for the $n-$th iteration is the following.
\begin{equation}
\begin{split}
\mathbf{v}_1^{l,n+1}(u,v) &=\mathbf{v}_1^{l,n}(u,v) \\
&~~+ \Delta_t \int \Lambda_{Hl}\mathbf{v}_1^{l,n}(u',v')du'dv'
\end{split}
\end{equation}
Suppose $\mathbf{v}_0=\sum_{i=1}^\infty \alpha_i\mathbf{e}_i(u,v)$, the update equation for the first iteration could be rewritten as follows.
\begin{equation}
\begin{split}
\mathbf{v}_1^{l,1}(u,v) &= \sum_{i=1}^\infty \alpha_i\mathbf{e}_i(u,v)\\
&~~+ \Delta_t \int \Lambda_{Hl} \sum_{i=1}^\infty \alpha_i\mathbf{e}_i(u,v)du'dv'\\
&=\sum_{i=1}^\infty \alpha_i\mathbf{e}_i(u,v) + \Delta_t \sum_{i=1}^\infty \lambda_i\alpha_i\mathbf{e}_i(u,v)\\
&=\sum_{i=1}^\infty (1+\Delta_t\lambda_i)\alpha_i\mathbf{e}_i(u,v)
\end{split}
\end{equation}
where we applied Mercer's representation. Hence, the update equation for the $n+1-$th iteration is the following.
\begin{equation}
\begin{split}
\mathbf{v}_1^{l,n}(u,v) &= \sum_{i=1}^\infty (1+\Delta_t\lambda_i)^n \alpha_i\mathbf{e}_i(u,v)\\
                        &= (1+\Delta_t\lambda_1)^n \Bigg[\alpha_1\mathbf{e}_1(u,v)\\
                        &~~~~+\sum_{i=2}^\infty \underbrace{\left({1+\Delta_t\lambda_i\over 1+\Delta_t\lambda_1}\right)^n}_{\approx0}\alpha_i\mathbf{e}_i(u,v)\Bigg]\\
&\approx (1+\Delta_t\lambda_1)^n \alpha_1\mathbf{e}_1(u,v)
\end{split}
\end{equation}
The normalization constraint gives us the desired result.
\begin{equation}
\lim_{n\rightarrow\infty }\mathbf{v}_1^{l,n}(u,v)={\mathbf{v}_1^{l,n}(u,v)\over \|\mathbf{v}_1^{l}(u,v)\|}= \pm\mathbf{e}_1(u,v)
\end{equation}
where we may omit the $\pm$ sign. The proofs for the subsequent steps is hence straightforward, where we only need to replace $\mathbf{v}_0=\sum_{i=1}^\infty \alpha_i\mathbf{e}_i(u,v)$ with $\mathbf{v}_0=\sum_{i=j}^\infty \alpha_i\mathbf{e}_i(u,v), j=2,3,...,K$. This setting is due to the Gram-Schmidt process in the constraints. The energy $U(\{\mathbf{v}_i^{l}\})$ is therefore the sum of eigenvalues due to Mercer's representation.
\end{proof}

\begin{proof}[Proof of Corollary \ref{COR:LinCvg}]
According to the proof of Theorem \ref{THM:EigenProc}, the $n-$th iteration for STEP $k$ is the following.
\begin{equation}
\begin{split}
\mathbf{v}_k^{l,n}(u,v) &= (1+\Delta_t\lambda_k)^n \Bigg[\alpha_k\mathbf{e}_k(u,v)\\
&~~~~+\sum_{h=k+1}^N \left({1+\Delta_t\lambda_h\over 1+\Delta_t\lambda_k}\right)^n\alpha_h\mathbf{e}_h(u,v)\Bigg]\\
\end{split}
\end{equation}
Let $\widetilde{\mathbf{v}}_k^{l,n}(u,v)=\mathbf{v}_k^{l,n}(u,v)/(1+\Delta_t\lambda_k)^n$ , $\beta_h = \left\| {1+\Delta_t\lambda_h\over 1+\Delta_t\lambda_k}\right\|$. Note that the eigenvalues are ordered. Hence, $\beta_h<1$. From these we deduce the following.
\begin{equation}
\begin{split}
&{\|\widetilde{\mathbf{v}}_k^{l,n+1}(u,v)-\alpha_k\mathbf{e}_k(u,v)\|\over\|\widetilde{\mathbf{v}}_k^{l,n}(u,v)-\alpha_k\mathbf{e}_k(u,v)\|} = {\sum_{h=k+1}^N\beta_h^{n+1}|\alpha_h|\over\sum_{h=k+1}^N\beta_h^n|\alpha_h|}\\
&= 1-{\sum_{h=k+1}^N\beta_h^{n}|\alpha_h|(1-\beta_h)\over\sum_{h=k+1}^N\beta_h^n|\alpha_h|}\\
&\leq1-\sum_{h=k+1}^N{\beta_h^{n}|\alpha_h|(1-\beta_h)\over (N-k-1) \beta_*^n|\alpha_*|}\\
&\leq1-{1\over {N-k-1}}\sum_{h=k+1}^N(1-\beta_h)\\
&={1\over {N-k-1}}\sum_{h=k+1}^N\beta_h<1
\end{split}
\end{equation}
where $\beta_*^n|\alpha_*| = \max\{\beta_h^n|\alpha_h|,h=k+1,i+2,...,N\}$.
The above completes the proof.
\end{proof}

\begin{proof}[Proof of Theorem \ref{THM:BOUND}:]
The energy of Eq. (\ref{EQ:Proj_K}) in terms of $\{\mathbf{w}_i,i=1,2,...,k\}$ is the following.
\begin{equation}\label{EQ:Opt}
\begin{split}
&U(\{\mathbf{w}_k\}) = \int\limits_{\Omega} \sum_{k=1}^{K}\langle \mathbf{p},\mathbf{w}_k\rangle_{\square}^2 dxdy\\
&= \int\limits_{\{u,v,u',v'\}} \sum_{k=1}^{K}\mathbf{w}_k(u,v)\mathbf{w}_k(u',v')\Lambda_{Hl} dudv du'dv'\\
&= \sum_{k=1}^{K}\sum_{h=1}^N \alpha_{kh}^2\lambda_{h}
\end{split}
\end{equation}
where $\mathbf{w}_k=\sum_{h=1}^N \alpha_{kh}\mathbf{e}_h$. Note that $\sum_{h=1}^N \alpha_{kh}^2=1$. In words, the summation over $h$ is a convex combination of $\lambda_h$. Note that the $\lambda_h$ are ordered from large values to small values for $h=1$ to $N$. Hence, the following holds.
\begin{equation}\label{EQ:BD_beta}
\begin{split}
\sum_{h=1}^N \alpha_{kh}^2\lambda_{h}&= \sum_{h=1}^K \alpha_{kh}^2\lambda_{h}+\sum_{h'=K+1}^N \alpha_{kh'}^2\lambda_{h'}\\
&=\sum_{h=1}^K \alpha_{kh}^2\lambda_{h}+\sum_{h'=K+1}^N \alpha_{kh'}^2\left(\sum_{h=1}^{K}\beta_h\right)\lambda_{h'}\\
&\hbox{`where,}\sum_{h=1}^{K}\beta_h=1, \hbox{and, } \beta_h>0\\
&\leq\sum_{h=1}^K \alpha_{kh}^2\lambda_{h}+\sum_{h'=K+1}^N \alpha_{kh'}^2\left(\sum_{h=1}^{K}\beta_h\lambda_{h}\right)
\end{split}
\end{equation}
The last inequality is due to that the convex combination of bigger values is bigger than smaller values. Now we consider a special case of $\beta_h$ as follows.
\begin{equation}
\beta_h= {1-\sum_{k=1}^K\alpha^2_{kh}\over \sum_{k=1}^k\left(1-\sum_{h=1}^K\alpha^2_{kh}\right)}
\end{equation}

This $\beta_h$ satisfies the following.
\begin{equation}
\sum_{h=1}^K\beta_h= {\sum_{h=1}^K\left(1-\sum_{k=1}^K\alpha^2_{kh}\right)\over \sum_{k=1}^K\left(1-\sum_{h=1}^K\alpha^2_{kh}\right)}=1
\end{equation}

Besides, regarding the inequality in (\ref{EQ:BD_beta}) we have the following.
\begin{equation}
\begin{split}
&\sum_{k=1}^K \left( \alpha_{kh}^2+\beta_h\sum_{h'=K+1}^N \alpha_{kh'}^2\right)\\
&= \sum_{k=1}^K \left[\alpha_{kh}^2+\beta_h\left(1-\sum_{h'=K}^N \alpha_{kh'}^2\right)\right]\\
&= \sum_{k=1}^K \left[\alpha_{kh}^2+{1-\sum_{k=1}^K\alpha^2_{kh}\over \sum_{k=1}^K\left(1-\sum_{h=1}^K\alpha^2_{kh}\right)}\left(1-\sum_{h'=1}^K \alpha_{kh'}^2\right)\right]\\
&= \sum_{k=1}^K \left[\alpha_{kh}^2+{1-\sum_{h'=1}^K \alpha_{kh'}^2\over \sum_{k=1}^K\left(1-\sum_{h=1}^K\alpha^2_{kh}\right)}\left(1-\sum_{k=1}^K\alpha^2_{kh}\right)\right]\\
&=\sum_{k=1}^K \alpha_{kh}^2+\left(1-\sum_{k=1}^K\alpha^2_{kh}\right)=1
\end{split}
\end{equation}
Substituting the above into (\ref{EQ:BD_beta}), we have the following.
\begin{equation}
\begin{split}
U(\{\mathbf{w}_k\})&\leq \sum_{h=1}^{K}\lambda_{h}\sum_{k=1}^K \left( \alpha_{kh}^2+\beta_h\sum_{h'=K+1}^N \alpha_{kh'}^2\right)\\
&=\sum_{h=1}^{K} \lambda_{h} = U(\{\mathbf{e}_k\})
\end{split}
\end{equation}
which completes our proof.
\end{proof}
The global optimality of PCA for the reconstruction of zero-mean vectors has been reported in \cite{Baldi89PCAOpt}. Our proof of global optimality of the eigenpatches for reconstruction of arbitrary patches is quite different from theirs, and our procedure of the proof is simpler.

{\scriptsize
\bibliographystyle{IEEEtran}
\bibliography{LevelSetActiveContours,MRFseg,Texture,SignalProcessing}
}

\ifCLASSOPTIONcaptionsoff
  \newpage
\fi

\end{document}